\documentclass[journal]{IEEEtran}

\ifCLASSINFOpdf
\else
   \usepackage[dvips]{graphicx}
\fi
\usepackage{url}

\usepackage{amsmath,amssymb,amsfonts}
\usepackage{mathrsfs}

\usepackage{graphicx}
\graphicspath{{./figures/}{./EXPfigures/}}

\usepackage{multirow}
\usepackage{multicol}
\usepackage{booktabs}
\usepackage{textcomp}

\usepackage{algorithm}  
\usepackage{algorithmicx}
\usepackage{algpseudocode}  

\begin{document}

\title{Image Steganography based on Iteratively Adversarial Samples of A Synchronized-directions Sub-image}

\author{Xinghong Qin, Shunquan Tan, \IEEEmembership{Senior Member, IEEE}, Bin Li, \IEEEmembership{Senior Member, IEEE}, Weixuan Tang and Jiwu Huang, \IEEEmembership{Fellow, IEEE}
\thanks{This paragraph of the first footnote will contain the date on which you submitted your paper for review. It will also contain support information, including sponsor and financial support acknowledgment. For example, ``This work was supported in part by the U.S. Department of Commerce under Grant BS123456.'' }
\thanks{Xinghong Qin, Bin Li and JiWu Huang are with Guangdong Key Laboratory of Intelligent Information Processing and Shenzhen Key Laboratory of Media Security, College of Information Engineering, Shenzhen University, Shenzhen 518060, China. (e-mail: 2156130112@email.szu.edu.cn, libin@szu.edu.cn and huangjw@szu.edu.cn).}
\thanks{Shunquan Tan is with College of Computer and Software Engineering, Shenzhen University, Shenzhen 518060, China.(e-mail: tansq@szu.edu.cn)}
\thanks{Weixuan Tang is with Institute of Artificial Intelligence and Blockchain, Guangzhou University, Guangdong 510006, China.(e-mail: tweix@gzhu.edu.cn)}
\thanks{Bin Li is also with Peng Cheng Laboratory, Shenzhen 518024, China.}}

\markboth{Journal of \LaTeX\ Class Files, Vol. 14, No. 8, August 2015}
{Shell \MakeLowercase{\textit{et al.}}: Bare Demo of IEEEtran.cls for IEEE Journals}
\maketitle

\begin{abstract}
Nowadays a steganography has to face challenges of both feature based staganalysis and convolutional neural network (CNN) based steganalysis. In this paper, we present a novel steganography scheme denoted as {ITE-SYN} (based on \textit{ITE}ratively adversarial perturbations onto a \textit{SYN}chronized-directions sub-image), by which security data is embedded with synchronizing modification directions to enhance security and then iteratively increased perturbations are added onto a sub-image to reduce loss with cover class label of the target CNN classifier. Firstly an exist steganographic function is employed to compute initial costs. Then the cover image is decomposed into some non-overlapped sub-images. After each sub-image is embedded, costs will be adjusted following clustering modification directions profile. And then the next sub-image will be embedded with adjusted costs until all secret data has been embedded. If the target CNN classifier does not discriminate the stego image as a cover image, based on adjusted costs, we change costs with adversarial manners according to signs of gradients back-propagated from the CNN classifier. And then a sub-image is chosen to be re-embedded with changed costs. Adversarial intensity will be iteratively increased until the adversarial stego image can fool the target CNN classifier. Experiments demonstrate that the proposed method effectively enhances security to counter both conventional feature-based classifiers and CNN classifiers, even other non-target CNN classifiers.

\end{abstract}


\IEEEpeerreviewmaketitle

\section{Introduction}
\label{sec:intro}
\IEEEPARstart{S}{teganography} and steganalysis are a pair of antagonistic players, which compete with each other and consequently benefit and  evolve from the competition\cite{Li2011}. Image steganography is the scientific art of concealing secret information within digital images and trying to escape detected by steganalysis. On the opposite, steganalysis devotes to reveal the presence of secret information through detecting whether there are traces left by data hiding. Recently most image steganographic schemes are designed under distortion minimization framework\cite{Fridrich2007}. Under the framework, the embedding costs of components of a image are estimated to guide that modifications are distributed to the most complex or textured areas to minimize expected distortion. A sequence of heuristic algorithms such as 
S-UNIWARD\cite{Holub2014a} and HILL\cite{Li2014} etc. directly compute costs. Based on multivariate Gaussian model, a sequence of algorithms such as MVG\cite{Fridrich2013}, 
MiPOD\cite{MiPOD2016} and MGR\cite{Qin_MGR} etc. model components of a image as a sequence of independent Gaussian variables and try to minimize the Kullback-Leibler divergence (KLD) between the cover and the stego objects with the corresponding Fisher information (FI) to compute embedding probabilities. And then a practical efficient coding method such as STCs (Syndrome-Trellis Codes)\cite{Filler2011} is employed to execute embedding operations. Furthermore, some synchronizing modification directions (SMD) profiles\cite{Denemark_synch2015} are proposed to improve steganographic security, such as CMD\cite{Li_CMD2015}, SYNCH\cite{Denemark_synch2015} and ASYM\cite{Hu2018_ASYM} etc.

Oppositely, steganalysis commonly uses machine learning models to learn differences of distributions from cover images to stego images. With ensemble classifiers (ECs)\cite{Kodovsky2012}, the state of the art steganalystic methods adopt high-order statistics with much high dimensional features such as SRM\cite{Fridrich2012} and TLBP\cite{Li2017_TLBP} etc. to achieve competent discriminative accuracy. Furthermore, some selection-channel-aware version features are proposed to improve steganalystic performance, such as maxSRM\cite{Denemark2014Selection} and PDASS\cite{Tan2017_PDASS} etc. In recent years, deep convolutional neural networks (CNN) are the mainstream in machine learning domain. Since Tan et al. proposed a CNN steganalysis by using a stacked convolutional auto-encoder in the pre-training procedure\cite{TanNet2014}, there have been many CNN steganalystic models\cite{Xu2016_DL,Xu2017_CNN,Ye2017,Boroumand2019_SRNet} 
proposed to achieve better and better detecting performance even overperform ECs with conventional features. 

For countering CNN steganalystic models detecting, there have been some steganographic schemes proposed. Being trained on generative adversarial network (GAN), Tang et al. proposed ASDL-GAN\cite{Tang2017_ADV} which performs similar to S-UNIWARD, and Yang et al. proposed other one more efficient UT-GAN\cite{Yang2018} based on U-Net. Inspired by adversary samples\cite{Szegedy2013_Adverarial,Cubuk2017_ADV}, to fool the target CNN steganalysis, Zhang et al. proposed ADS\cite{Zhang2018_ADS} which iteratively adds noise to a cover image to enhance the cover image and then embed secret information to the enhanced cover image, 
and Tang et al. proposed {ADV-EMB}\cite{Tang2019_ADS} which randomly split a cover image to two parts and trying to added minimum amount of adversarial perturbation to the second part according to signs of gradients to increase loss of the stego image, 
and Zha et al. proposed ECN few-pixel-attack\cite{Zha2019_ADV} to directly attack the stego image with modification of $\pm 4$ times values, which is based on ECN (extraction conservation noise) and iteratively modify the population number of pixels of the stego image and select candidates with the maximum loss as the output adversarial sample. For above schemes, {ADV-EMB} achieves good performance.
Furthermore, with min-max strategy, Bernard et al. proposed Min-Max function to enhance steganographic performance of ADV-EMB, by which they dynamically selected the most difficult stego images associated with the best classifiers at each iteration\cite{Bernard2019_MM}. 
Obviously these schemes pay concentration on countering CNN steganalystic models. 
Simultaneously, Chen et al. proposed SPP\cite{Chen2019_SPP} which via stego post-processing to reduce image residual difference from the cover and the stego to enhance steganographic capability to counter feature-based classifiers such as ECs with SRM or maxSRMd2. It only pays attention to resist on detection of conventional classifiers.

However, nowadays a steganographic scheme has to face challenges of both conventional feature-based steganalystic models and CNN steganalystic classifiers. In this paper, incorporating iteratively increasing adversarial perturbations and synchronizing modification directions profile, we propose a novel steganographic scheme for spatial images. Employing an exist cost function, such as S-UNIWARD\cite{Holub2014a}, HILL\cite{Li2014}, MiPOD\cite{MiPOD2016} and MGR\cite{Qin_MGR} etc., initial costs of a cover image are computed. We decompose the image into some non-overlapped sub-images, and sequentially embed corresponding segment of messages into sub-images. Before each sub-image is embedded, with SMD profile, costs are adjusted according to modifications to prompt synchronizing embedding directions. If the target CNN classifier discriminates the stego image correctly, iteratively increase the adversarial intensity to change adjusted costs with adversarial manner, and select one sub-image to be re-embedded until the stego image can fool the CNN classifier. We name the proposed scheme \textit{ITE-SYN} because it is based on \textit{ITE}ratively adversarial perturbations onto a \textit{SYN}chronized-directions sub-image. Compare for the current state of the art schemes, our method has some characterizers such as following.
\begin{itemize}
  \item [1)] Synchronizing modification directions profile applied, stego images achieve high capability to counter steganalystic classifiers. Incorporating adversarial perturbation, stego images achieve high performance to fool the target CNN steganalystic models. 
  \item [2)] The cover image is decomposed into some non-overlapped sub-images and adversarial perturbation is only applied onto one sub-image, which limits the amount of adversarial elements small and costs little computational time for each re-embedding operation. The division guarantees the SMD profile can be applied. And, adversarial costs are changed based on adjusted costs, which guarantees stego images keeping secure level of SMD.
  \item [3)] Iteratively increase adversarial intensity to improve adversarial capability, which prompts to limit adversarial perturbations minimal and introduce diversity of stego images.
\end{itemize}

Extensive experiments illustrate that the proposed method significantly improve steganographic performance to counter both conventional classifiers and CNN models.

The rest of this paper is organized as follows. Some basic concepts or related works are simply introduced in section \ref{sec:pre}. Section \ref{sec:method} describes the proposed method in detail. Extensive experiments are exhibited in section \ref{sec:exper}. Section \ref{sec:conclusion} is conclusion.

\section{Related works}
\label{sec:pre}

\subsection{Notations}
\label{sec:pre:ssec:notation}

In this paper, capital letters in bold are used to present vectors or matrices, and the corresponding low-case letters are used to present elements of vectors or matrices. A pair of number in parentheses $(i,j)$ is used to denote the location index of the $i$-th row and the $j$-th column. Caligraphic letters are used to present sets. The flourish letters are used to present functions or classifiers. Superscripts are used to indicate location indexes, and subscripts are used to indicate the image sets and steganographic methods used. Specially, we use $\mathbf{C}$, $\mathbf{S}$ and $\mathbf{Z}$ to denote a cover image, its corresponding stego image and adversarial stego image respectively, $\xi$, $\rho$ and $\pi$ to denote initial cost, adjusted cost and embedding probability respectively. 

\subsection{Distortion minimization framework}
\label{sec:pre:ssec:distortion}

Under the distortion minimization framework\cite{Fridrich2007}, the steganographic embedding can be formulated as the following optimization problem.
\begin{equation}\label{eq:md}
  \min_{\pi } \sum_{(i,j)=(1,1)}^{(\mathit{M},\mathit{N})}\sum_{\Delta c\in \mathcal{X}}\pi_{\Delta c}^{(i,j)}\rho_{\Delta c}^{(i,j)},  \quad \text{s.t. } h({\pi}) = L,
\end{equation}
where $h=-\sum_{\Delta c\in \mathcal{X}}\pi_{\Delta c}^{(i,j)}\log \pi_{\Delta c}^{(i,j)}$ is entropy function, in which $h$ is in bit unit and the logarithmic base is 2. And $\pi$ is embedding probabilities which are Gibbs distributions\cite{Filler2010} as 
\begin{equation}\label{eq:pi}
  \pi_{\Delta c}^{(i,j)}=\frac{e^{-\lambda\Delta c \rho_{\Delta c}^{(i,j)}}}{\sum_{t \in \mathcal{X}}e^{-\lambda t\rho_t^{(i,j)}}}.
\end{equation}
where $\lambda$ is determined by the payload constraint. Under the ternary modification model, there are $\Delta c^{(i,j)}\in \mathcal{X}=\{-1, 0, 1\}$. We use symbols + and - to denote +1 and -1 operations respectively  when not ambiguous.

\subsection{Steganalystic model}
\label{sec:pre:ssec:steganalys}

The essence of a steganalysis is a binary classifier.
\begin{equation}\label{eq:fun_classifier}
    \begin{aligned}
      \mathscr{F}(\mathbf{X})=\begin{cases}
      0 & \text{ if } \Phi (\mathbf{X})<  0.5 \\ 
      1 & \text{ if } \Phi (\mathbf{X})\geq 0.5,
      \end{cases}    
    \end{aligned}
\end{equation}
where $\Phi (\mathbf{X}) \in [0,1]$ is probability respect to the input $\mathbf{X}$ to indicate $\mathbf{X}$ being a cover ($\mathscr{F}=1$) or a stego ($\mathscr{F}=0$). For assess a steganalysis, there are two import indexes called the false-alarm probability $P_{FA}$ and missed-detection probability $P_{MD}$.
\begin{equation}\label{eq:fun_pfa}
  P_{FA}=Pr\{\mathscr{F}(\mathbf{X})|_{\mathbf{X}\in \mathcal{C}}=0\},
\end{equation}
\begin{equation}\label{eq:fun_pmd}
  P_{MD}=Pr\{\mathscr{F}(\mathbf{X})|_{\mathbf{X}\in \mathcal{S}}=1\},
\end{equation}
where $\mathcal{C}$ and $\mathcal{S}$ are the cover set and the stego set respectively.
The total performance is the probability of detection error computed from $P_{FA}$ and $P_{MD}$ and expressed in \eqref{eq:pe}.

The purpose of adversarial samples is increasing $P_{MD}$.

\subsection{Clustering modification directions}
\label{sec:pre:ssec:cmd}

Following the clustering rule (CR)\cite{Li2014}, which states that it is better to make embedding modifications clustered rather than scattered, clustering modification directions (CMD)\cite{Li_CMD2015} profile takes mutual impacts of embedding modifications of neighbor pixels into consideration to enhance steganographic security. In terms of CMD, a cover image is decomposed into several non-overlapped sub-images and each sub-image is embedded in sequence, in which costs are re-computed and adjusted according to embedding modifications before each sub-image is embedded. 

Given a cover image $\mathbf{C}$ with scale $W \times H$, where $W$ and $H$ are width and height respectively, and security data $\mathbf{M}$ with length $\mathit{L}$, employing an exist steganographic scheme, initial costs $\mathbf{\xi}$ are computed, and then the image is decomposed to $W_s \times H_s$ non-overlapped sub-images.

Each sub-image is composed of 
\begin{equation}\label{eq:sub_image}
    \begin{aligned}
        \mathbf{C}^{(a,b)}=\{c^{(i,j)}|  \, i=\mathit{k}_a W_s + a, \, j=\mathit{k}_b H_s +b \},
    \end{aligned}
\end{equation}
where $\mathit{k}_a\in \{0,1,\cdots,\lfloor \frac{W}{W_s} \rfloor-1\}$, $\mathit{k}_b\in \{0,1,\cdots,  \lfloor \frac{H}{H_s} \rfloor-1\}$. 

Sub-images are embedded in foreword zig-zag sequence as $\mathbf{C}^{(1,1)}\Rightarrow \mathbf{C}^{1,2} \Rightarrow \dots \mathbf{C}^{(1,W_s)}\Rightarrow \mathbf{C}^{(2,W_s)} \Rightarrow \mathbf{C}^{(2,W_s-1)} \Rightarrow \dots$. Before each sub-images is embedded, costs are adjusted as
\begin{equation}\label{eq:cmd_cost}
    \begin{aligned}
        \begin{cases}
        {\rho}_{+}^{(i,j)}=\xi_{+}^{(i,j)}/\beta   & \text{if \,\,} \sum_{\Delta c^{(r,s)}\in \mathcal{N}^{(i,j)}}{\Delta c^{(r,s)}}>0,\\
        {\rho}_{-}^{(i,j)}=\xi_{-}^{(i,j)}/\beta   & \text{if \,\,} \sum_{\Delta c^{(r,s)}\in \mathcal{N}^{(i,j)}}{\Delta c^{(r,s)}}<0,\\
        {\rho}_{\pm}^{(i,j)}= \xi_{\pm}^{(i,j)}  & \text{otherwise}, \\
        \end{cases}
    \end{aligned}
\end{equation}
where the subscripts $\pm$ denote embedding modification directions. $\Delta \mathbf{C}=\mathbf{S}-\mathbf{C}$ are embedding modifications, and $\mathcal{N}^{(i,j)}=\{\Delta c^{(r,s)}|r \in \{i-1, i+1\}, s\in\{j-1, j+1\}\}$ is the neighbor modification set of the $(i, j)^{th}$ location. And, CMD factor $\beta>1$ pursues the next sub-image is modified with synchronized directions.

\subsection{Iterative least-likely class method}
\label{sec:pre:ssec:isg}

Fast gradient sign method (FGSM)\cite{Goodfellow2014_ADS} is an fast tool of adversarial examples, which is expressed as
\begin{equation}\label{eq:fgms}
  \mathbf{X}_{adv}=\mathbf{X}+\epsilon\cdot sign(\bigtriangledown _\mathbf{X}\mathscr{L}(\mathbf{X},\mathbf{y}))
\end{equation}
where $\mathbf{X}$ and $\mathbf{X}_{adv}$ are the clean image and an adversarial sample fooling the target CNN classifier respectively, and $\mathscr{L}$ is the loss function, and $\mathbf{y}$ is the correct class label. $\epsilon \in [0,255]$ is constrained by pixels limitation. In practice, it is surprising universality of adversarial error at small $\epsilon$ to result in the CNN model classifying the adversarial sample to a wrong class. The least-likely attack form of FGMS is expressed as
\begin{equation}\label{eq:fgms_target}
  \mathbf{X}_{adv}=\mathbf{X}-\epsilon\cdot sign(\bigtriangledown _\mathbf{X}\mathscr{L}(\mathbf{X},\mathbf{y}_{ll}))
\end{equation}
where $\mathbf{y}_{ll}$ is the least-likely class (or target class) label. The purpose of the least-likely attack is pushing the CNN model to classify the adversarial sample to the target class $\mathbf{y}_{ll}$. 

Kurakin et al. introduce a simple enhancement of FGMS\cite{KurakinGB16}, where the gradient sign adversarial operations are executed iteratively and the result is clipped.
\begin{equation}\label{eq:igsm}
  \begin{aligned}
    \mathbf{X}_{adv}^0 & =\mathbf{X}, \\
    \mathbf{X}_{adv}^N & =Clip_{\mathbf{X},\epsilon}\{\mathbf{X}_{adv}^{N-1}-a\cdot sign(\bigtriangledown _\mathbf{X}\mathscr{L}(\mathbf{X},\mathbf{y}_{ll}))\} \\
  \end{aligned}
\end{equation}
where $a<\epsilon$ is a step. It produces superior results to FGSM. The enhanced method is called iterative least-likely class method (ILLCM).

\section{Method}
\label{sec:method}

In this section, we propose the steganographic scheme {ITE-SYN}, which is based on iteratively increasing adversarial perturbations onto a synchronized modification directions sub-image. First, we will draw the base framework of the proposed scheme. And then, we will describe the two core operations of the proposed scheme in detail.

\subsection{Base framework}
\label{sec:method:framework}

In the proposed scheme, the process of creating a stego image includes two phases, embedding data with synchronizing modification directions and iteratively adding adversarial perturbations onto a sub-image. In the first phase, we produce the stego image following simplified version of clustering modification directions profile, which is introduced in section \ref{sec:method:ssec:block}. For implementing the scheme, an image are decomposed into $2\times 2=4$ non-overlapped sub-images, and then each sub-image is embedded in sequence. In the second phase, we re-embed one sub-image with iteratively adversarial manner to fool the target CNN steganalystic classifier.  

The rest of this section is divided into three subsections, each of which devotes to a specific design of {ITE-SYN}.

\subsubsection{Clustering modification directions}
\label{sec:method:ssec:framework:cmd}

It is indicated that CNN is similar to conventional feature-based models in structure\cite{TanNet2014}. As well as it has been proved that SMD profile can enhance steganographic security\cite{Denemark_synch2015,Li_CMD2015}. Therefore, it is predictable that SMD profile can enhance steganographic performance to counter CNN classifiers. However, as far as we know, there still has been no any published work taking SMD profile into design of steganographic scheme to counter CNN classifiers. 

Clustering modification directions (CMD)\cite{Li_CMD2015} method is an effective, practicable and simple implementation of SMD profile, which core idea is that decomposing the image into some non-overlapped sub-images and embedding each sub-image in sequence. Before each sub-image is embedded, initial embedding costs are re-computed and then embedding costs are adjusted with expression \eqref{eq:cmd_cost} to prompt embedding modifications are clustered.

In this paper, we use a simplified version of CMD, by which the initial costs are computed once before embedding and randomly select the start sub-image. The detail processes are described in section \ref{sec:method:ssec:block}. There are following reasons for the simplification.
\begin{itemize}
  \item By using the state of the art cost schemes designed under minimizing distortion framework, embedding modifications are mainly distributed in texture or complex areas of the image, and amplitudes of modifications are slight (e.g $\pm 1$). Whereas the change rate is small even through under high payload rates. Therefore, re-computing initial costs introduce limited contribution for improving steganographic performance but more computational time. 
  \item Subsequently embedding costs are also adjusted to produce adversarial noise.
  \item Randomly select the start sub-image to introduce more diversity, which compensates for the deficiency caused by simplification.
\end{itemize}

\subsubsection{Iteratively adversarial steganographic perturbations}
\label{sec:method:ssec:framework:illcm}

Many machine learning classifiers are vulnerable to adversarial samples\cite{Szegedy2013_Adverarial}. It motivates us to design steganographic algorithm to attack CNN steganalysis.

For steganography, secret data is embedded into the cover to produce the stego. Following both adversarial samples and steganographics, a steganographier can only change the stego to attack the CNN classifier. Therefore, we can only produce adversarial samples to improve $P_{MD}$.
However, constrained by extraction of embedding data, we cannot directly apply adversarial methods, such as \eqref{eq:fgms}, \eqref{eq:fgms_target}, \eqref{eq:igsm}, etc. 

The essence of steganographics for digital images is hiding the secret information and secret communication by embedding the secret information to the cover image to produce the stego image and then transmitting the stego image. The practice of non-additive distortion\cite{Filler2010,Denemark_synch2015,Li_CMD2015,Hu2018_ASYM} has told us that we can adjust distortions as needed to make embedded changes occurred in the desired distribution to improve steganographic security. Consequently, we can modify embedding costs and re-embed the secret information to prompt that the new embedding modifications are equivalent to the original embedding modifications plus the adversarial noise, such as 
\begin{equation}\label{eq:adv_noise}
  \begin{aligned}
    \Delta \mathbf{C}'&=\mathbf{Z}-\mathbf{C} =\mathbf{Z}-\mathbf{S}+\mathbf{S}-\mathbf{C} \\
    &=\mathbf{\eta}+\Delta \mathbf{C},
  \end{aligned}
\end{equation}
where $\mathbf{\eta}=\mathbf{Z}-\mathbf{S}$ are adversarial noise.
We follow the idea of \eqref{eq:fgms_target} for the purpose of prompting the CNN classifier to discriminate the stego image as the cover class. For a correctly classified stego image, we increase costs with the same modification directions as signs of gradients back-propagated from the CNN classifier, whereas decrease costs with the opposite modification directions of signs of gradients, to push more modifications distributed to negative directions of gradients to realize function of adversarial noise.
For prior results, we also adopt the idea of \eqref{eq:igsm} to iteratively increase adversarial intensity.
For preserving properties of the stego image, we only apply adversarial perturbations onto a sub-image.
The detail processes are described in section \ref{sec:method:ssec:adversarial}.

\subsubsection{Impact of corporation}
\label{sec:method:ssec:framework:corporation}

Both SMD profile and adversarial perturbations are realized by adjusting embedding costs as needed to boost embedding modifications occurred as desired distribution. By SMD profile, embedding costs are adjusted according to embedding changes of neighborhood to synchronize modifications to reduce distance between steganalystic features. And by adversarial perturbations, embedding costs are adjusted according to gradients back-propagated by the target CNN classifier to urge new embedding modifications to reduce the loss to prompt the target CNN classifier discriminate the adversarial stego image as the cover class. Structures of both feature-based ensemble classifiers and CNN classifiers are similar. The difference is that the features of the former are computed by a stationary algorithm instead that the features of the later are learned from training samples. Therefore, the corporation of SMD and adversarial perturbations can introduce extra gain. It is main signification of the proposed method.

For the proposed scheme, we execute SMD operations firstly. Even through we use a simplified version of CMD method, for the convenience of discussion, it is assumed that the SMD operations are optimal. We use iteratively increasing adversarial intensity to produce the adversarial stego image in the next phase. Assuming that the step of the adversarial intensity is small enough and number of iterations is not limited, theoretically, there is always an optimal iteration to maximize the extra gain, at which adversarial stego images can not only furtherest fool the target CNN classifier, but also most strongly resist on detection of other steganalystic models. Unfortunately, as far as we know, there is no relevant knowledge and method to theoretically derive the optimal step and iteration number. 
It is predicable that a bigger celling of the adversarial intensity is beneficial to fool the target CNN classifier. However, negative effects of a bigger adversarial celling are more computational time and possible to counteract benefit of SMD profile.
Consequently, we need trade off impacts to select factors.

\subsection{Synchronized-directions embedding}
\label{sec:method:ssec:block}

Clustering modification directions (CMD)\cite{Li_CMD2015} is an effective and practicable SMD profile. We simplify CMD to implement synchronizing modification directions in the first phase of our method.
Given a cover image $\mathbf{C}$ with scale $\mathit{W} \times \mathit{H}$ and security data $\mathbf{M}$ with length $\mathit{L}$, employing an exist steganographic scheme, we initialize its costs $\mathbf{\xi}$, and decompose it into some non-overlapped sub-images, and then embed secret data into each sub-image with synchronizing modification directions profile by sequence. The detail algorithm is described in Algorithm \ref{alg:syn_emb}.

\begin{algorithm}[tbp]  
  \caption{ \textbf{Embedding secret data with synchronizing modification directions.}\newline
  $\mathbf{C}$ is the cover image, $\mathbf{S}$ is the stego image, $\mathbf{M}$ is the secret data.\newline
  $\mathbf{\xi}$ is the initial costs, $\mathbf{\rho}$ is the adjusted costs. \newline
  $W_s$ and $H_s$ are quantities of horizontal and vertical decomposition respectively.\newline
  $\beta$ is the CMD factor.}  
  \label{alg:syn_emb}  
  \begin{algorithmic}[1]  
    \Require 
      $\mathbf{C}$, $\mathbf{M}$, $W_s$, $H_s$, $\beta$
    \Ensure 
      $\mathbf{S}$
    \State Compute $\mathbf{\xi}$
    \State $\mathbf{S} \gets \mathbf{C}$\\
          $\mathbf{\rho} \gets \mathbf{\xi}$
    \State Decompose $\mathbf{C}$ into $W_s\times H_s$ non-overlapped sub-images following \eqref{eq:sub_image}\\
           Decompose $\mathbf{\rho}$ into  $W_s\times H_s$ corresponding non-overlapped sub-sets\\
           Averagely split $\mathbf{M}$ into $W_s\times H_s$ corresponding segments
    \State Randomly select a cover sub-image $\mathbf{C}^{(i,j)}$
    \While {Sub-segment data pending for being embedded}
    \State $\mathbf{S}^{(i,j)} \gets$  Embed $\mathbf{M}^{(i,j)}$ into $\mathbf{C}^{(i,j)}$ with $\mathbf{\rho}^{(i,j)}$
    \State Update $\mathbf{S}^{(i,j)}$ into $\mathbf{S}$
    \State Adjust costs $\mathbf{\rho}$ with $\Delta \mathbf{C}=\mathbf{S}-\mathbf{C}$ following \eqref{eq:cmd_cost}
    \State $(i,j) \gets $ Index of the next cover sub-image
    \EndWhile
    \State 
    \Return $\mathbf{S}$  
  \end{algorithmic}
\end{algorithm}  

There have been many effective cost methods proposed. We employ a given method to compute initial costs $\mathbf{\xi}$ of the cover image $\mathbf{C}$.  In this paper, we use four most recent steganographic schemes, including two heuristic methods S-UNIWARD\cite{Holub2014a} and HILL\cite{Li2014}, and two model-based methods MIPOD\cite{MiPOD2016} and MGR\cite{Qin_MGR}, to illustrate our method.
 
In this paper, we select $\mathit{W}_s=\mathit{H}_s=2$. Therefore, there are four sub-images. We use zig-zag sequence to embed sub-images. For diversity, we randomly select the start sub-image instead of the first sub-image $\mathbf{C}^{(1,1)}$. Thus, the embedding sequence is the closed loop as $\mathbf{C}^{(1,1)}\Rightarrow \mathbf{C}^{(1,2)}\Rightarrow \mathbf{C}^{(2,2)} \Rightarrow \mathbf{C}^{(2,1)} \Rightarrow$.

After the stego image $\mathbf{S}$ has been updated, we adjust costs according to embedding changes $\Delta\mathbf{C}=\mathbf{S}-\mathbf{C}$.  
We compute the initial costs $\mathbf{\xi}$ once instead re-compute $\mathbf{\xi}$ from $\mathbf{S}$ before each sub-image is embedded.
In \cite{Li_CMD2015}, the optimal parameter $\beta=9$ for $512\times 512$ gray-scale images based on experiments. In this paper, we use $256\times 256$ gray-scale images and arbitrarily set CMD factor $\beta=10$.

In practice, embedding codes such as STCs(syndrome trellis codes)\cite{Filler2011} are used to execute steganographic embedding. In experiments, the optimal embedding simulator can be used as well. In this paper, we implement the embedding simulator as 
\begin{equation}\label{eq:simulator}
	\Delta c^{(i,j)}=\begin{cases}
	1 & \text{ if } \mathit{p}^{(i,j)}<\pi_{+}^{(i,j)} \\ 
	-1 & \text{ if } \mathit{p}^{(i,j)}>1-\pi_{-}^{(i,j)} \\ 
	0 & \text{ otherwise } ,
	\end{cases}
\end{equation}
where probability matrix $\mathbf{P}$ is the same scale as the image and is generated from a uniform distribution on interval [0, 1].

\subsection{Adversarial embedding a sub-image}
\label{sec:method:ssec:adversarial}

Many machine learning classifiers are vulnerable to adversarial samples. A frequently small fraction of adversarial perturbation can lead to targeted misclassification or to result in a specific output classification\cite{Cubuk2017_ADV}.
However, there are some defects when directly modify a stego image to produce an adversarial stego image because it is  constrained by extracting embedded data\cite{Zha2019_ADV}. Our idea is that inducing changes of embedding modifications to be occurred as distribution of adversarial noise. Consequently, abiding by laws of steganographic embedding, we adjust embedding costs to implement adversarial perturbations. For limiting a small fraction of adversarial perturbation to preserve properties of stego images, we only apply adversarial perturbations onto a sub-image.

Iterative least-likely class method (ILLCM)\cite{KurakinGB16} produces superior results to FGSM\cite{Goodfellow2014_ADS}. ILLCM is expressed by \eqref{eq:fgms_target}-\eqref{eq:igsm}. For an image $\mathbf{X}$, the adversarial sample $\mathbf{X}_{adv}$ is produced by adding adversarial noise with negative directions of gradients back-propagated from the loss of the target CNN classifying $\mathbf{X}$ to the least-likely class $\mathbf{y}_{ll}$, which is helpful to reduce the loss to prompt the target CNN classifier to discriminate $\mathbf{X}_{adv}$ to the $\mathbf{y}_{ll}$ class. And, it can achieve a superior results by iteratively increasing the adversarial intensity with a small step.
For steganogaphy for images, there are $\mathbf{X}=\mathbf{S}$, $\mathbf{X}_{adv}=\mathbf{Z}$ and $\mathbf{y}_{ll}=\mathbf{y}_c$ respectively. We iteratively adjust costs with a small step $\Delta \gamma$ according to signs of gradients of the stego image $\mathbf{S}$ with cover label $\mathbf{y}_c$ classified by the target CNN model. For prompting more embedding modifications with opposite directions of gradients, embedding costs with same modification directions as signs of gradients will be increased, whereas embedding costs with opposite modification directions to signs of gradients will be decreased. Expressed as the following.
\begin{equation}\label{eq:adv_cost_p}
  \begin{aligned}
    \rho _{adv+}^{(i,j)}=\begin{cases}
    \rho_+^{(i,j)}(1+k\Delta\gamma) & \text{ if } \bigtriangledown \mathscr{L}^{(i,j)}(\mathbf{S},\mathbf{y}_c)>0, \\ 
    \rho_+^{(i,j)}/(1+k\Delta\gamma) & \text{ if } \bigtriangledown \mathscr{L}^{(i,j)}(\mathbf{S},\mathbf{y}_c)<0, \\ 
    \rho_+^{(i,j)} & \text{ otherwise },
    \end{cases}
  \end{aligned}
\end{equation}
\begin{equation}\label{eq:adv_cost_m}
  \begin{aligned}
    \rho _{adv-}^{(i,j)}=\begin{cases}
    \rho_-^{(i,j)}(1+k\Delta\gamma) & \text{ if } \bigtriangledown \mathscr{L}^{(i,j)}(\mathbf{S},\mathbf{y}_c)<0, \\ 
    \rho_-^{(i,j)}/(1+k\Delta\gamma) & \text{ if } \bigtriangledown \mathscr{L}^{(i,j)}(\mathbf{S},\mathbf{y}_c)>0, \\ 
    \rho_-^{(i,j)} & \text{ otherwise },
    \end{cases}  
  \end{aligned}
\end{equation}
where $\mathscr{L}$ is the loss function of the target CNN classifier, and $\mathbf{y}_c$ is the cover class label, and $\gamma=k\Delta \gamma\leq \gamma_{max}$ is adversarial intensity factor, and $k\in \mathbb{N}$ is the number of iteration, and $\Delta \gamma>0$ and $\gamma_{max}<\infty$ are the step and the celling of the adversarial intensity respectively.  
Constrained by extracting embedded data, increasing iteration makes only a bigger adversarial intensity instead of clipping each iterative image, which is different from ILLCM. For a stego image $\mathbf{S}$, the gradients $\bigtriangledown \mathscr{L}(\mathbf{S},\mathbf{y}_c)$ are only computed once, which reduces computational complexity.

Then, we only re-embed one sub-image by using the adversarial embedding costs $\mathbf{\rho}_{adv}$ to create adversarial noise.
For a given trained CNN steganalystic classifier $\mathscr{F}$ and a stego image $\mathbf{S}$, if the classifier discriminates the stego image to the correct class, we select a sub-image and re-embed the corresponding segment secret data following up adversarial perturbation process as Algorithm \ref{alg:adv_emb}.

\begin{algorithm}[tbp]  
  \caption{ \textbf{Adversarial Embedding a sub-image to fool the target {CNN} classifier.}\newline
  This algorithm follows Algorithm \ref{alg:syn_emb}.\newline 
  $\mathscr{F}$ is the target CNN classifier, $\mathbf{y}_c$ is the class label of cover images. \newline
  $\gamma_{max}$ is the celling of the adversarial intensity factor, $\Delta \gamma$ is the step interval.\newline 
  $\mathbf{S}$ and $\mathbf{\rho}$ are the stego image and the adjusted costs produced from Algorithm \ref{alg:syn_emb} respectively.\newline
  $\mathbf{\rho}_{adv}$ is the adversarial costs, $\mathbf{Z}$ is the adversarial stego image. \newline
  Other variables are refer to Algorithm \ref{alg:syn_emb}.}  
  \label{alg:adv_emb}  
  \begin{algorithmic}[1]  
    \Require 
      $\mathscr{F}$, $\mathbf{C}$, $\mathbf{M}$, $\mathbf{S}$, $\mathbf{\rho}$, $\Delta \gamma$, $\gamma_{max}$
    \Ensure 
      $\mathbf{Z}$
    \State $\mathbf{Z} \gets \mathbf{S}$
    \If {$\mathscr{F(\mathbf{Z})}=1$} 
    \State \Return $\mathbf{Z}$
    \EndIf 
    \State Compute $\bigtriangledown \mathscr{L}(\mathbf{S},\mathbf{y}_c)$
    \State Randomly select a cover sub-image $\mathbf{C}^{(i,j)}$
    \While {$\mathbf{C}^{(i,j)}$ has not been tried re-embedded}
      \State $k\gets 1,\quad \gamma \gets k\Delta \gamma$
        \While {$\gamma < \gamma_{max}$}
          \State Compute $\mathbf{\rho}_{adv}$ following \eqref{eq:adv_cost_p} and \eqref{eq:adv_cost_m}
          \State $\mathbf{Z}^{(i,j)} \gets$  Embed $\mathbf{M}^{(i,j)}$ into $\mathbf{C}^{(i,j)}$ with $\mathbf{\rho}_{adv}^{(i,j)}$
          \State Update $\mathbf{Z}^{(i,j)}$ into $\mathbf{Z}$
          \If {$\mathscr{F(\mathbf{Z})}=1$} 
          \State 
          \Return $\mathbf{Z}$
          \EndIf 
        \State $k\gets k+1, \quad \gamma=k\Delta \gamma$
        \EndWhile
      \State $(i,j) \gets $ Index of the next cover sub-image
      \State $\mathbf{Z} \gets \mathbf{S}$
    \EndWhile
    \State 
    \Return $\mathbf{Z}$  
  \end{algorithmic}
\end{algorithm}  

For diversity, we randomly select the start sub-image to perform adversarial operations. Once the adversarial stego image $\mathbf{Z}$ can fool the target classifier leading result in a cover class, output $\mathbf{Z}$ and end the adversarial process. Otherwise we select next sub-image to go through the same procedures. If fooling the classifier failed finally, the adversarial stego image is resumed as the stego image $\mathbf{S}$.

Since adversarial costs are based on costs adjusted according to CMD profile, adversarial stego images can not only effectively fool the target CNN classifier, but also keep high steganographic security to counter other classifiers, even through the classifiers are re-trained by using them. This opinion is proved by experiments described in sections \ref{sec:exper:ssec:gamma} and \ref{sec:exper:ssec:comparison} too. 

In this paper, we set the step interval $\Delta \gamma=0.1$ and select the celling of the adversarial intensity factor $\gamma_{max}=10$ also refer to CMD factor $\beta=10$.
For a single sub-image, the maximum amount of iterations is 100. Therefore, on the worst case, the maximum amount of iterations is 400. However, only one sub-image is re-embedded in each iteration, which limits the total expense of computation. This opinion is proved by experiments described in \ref{sec:exper:ssec:time} too.

\section{Experiments}
\label{sec:exper}

In this section, we conducted experiments to evaluate the proposed ITE-SYN method. 
\begin{itemize}
  \item In practice, stego images are produced by STCs tool. For evaluation, stego images are always created by the optimal embedding tool. We compared for performances of stego images produced by STCs and the embedding tool respectively. It will be reported in section \ref{sec:exper:ssec:setup}.
  \item There are two factors $\beta$ and $\gamma_{max}$ in the proposed algorithm. For convenience of discussion, we only investigated impacts of different $\gamma_{max}$. It will be reported in section \ref{sec:exper:ssec:gamma}.
  \item We compared performances of the proposed method for the state of the art steganographic methods. Performances include capabilities of both deceiving original steganalystic classifiers and countering detection of adversarial trained classifiers. It will be reported in section \ref{sec:exper:ssec:comparison}.
  \item In section \ref{sec:exper:ssec:time}, we evaluated computational time of the proposed method.
\end{itemize}

\subsection{Setup}
\label{sec:exper:ssec:setup}

\begin{enumerate}
  \item[1)] Image sets: We use the following two cover image sets.
    \begin{itemize}
      \item BOSS256. It contains 20000 gray-scale images from union of BOSSBase v1.01\cite{Bas2011} and BOWS2\cite{bows_2}, each of which contains 10000 gray-scale images with original size $512\times 512$, by resizing each image to scale $256\times 256$ by using \textit{imrezise} function with default setting in Matlab. For CNN classifiers, randomly choose 5000 and 1000 images from BOSSBase for testing and validation respectively, and rest images for training. 
      \item ALASKA256. It contains 47000 gray-scale images with size $256\times 256$ randomly chosen from ALASKA\cite{alaska2019}, which contains 50000 images in RAW format taken from 21 different cameras. We have downloaded only 49231 RAW images successfully. We followed procedures in \cite{alaska2019} to convert RAW format images to 16-bit TIF format true color images by using RawTherapee v5.7. And then we used Matlab with default parameters if no specification to produce 8-bit $256 \times 256$ gray-scale images. We proportionally resized the smaller side of each image to 256 pixels by using \textit{imresize} function, and then centering cropped it to size $256 \times 256$ by using \textit{imcrop} function. Finally, we converted images to gray-scale format by using \textit{rgb2gray} function. We randomly selected 47000 images. For CNN classifiers, we split samples to three sets containing 2000 images, 10000 images and 35000 images for validation, testing and training respectively.
    \end{itemize}
  \item[2)] Cost functions: we use four current state of the art cost schemes, including two heuristic methods S-UNIWARD\cite{Holub2014a} and HILL\cite{Li2014} and two model-based methods MiPOD\cite{MiPOD2016} and MGR\cite{Qin_MGR}, to compute initial costs.
  \item[3)] Classifiers: we use six current state-of-the-art classifiers, including three CNN classifiers and three conventional classifiers.
    \begin{itemize}
      \item XuNet\cite{Xu2016_DL} and YeNet\cite{Ye2017}: these two CNN steganalystic networks were used as the target classifiers. For each classifier of XuNet, training was executed $120k$ iterations, and the model with the best steganalystic performance from the last $20k$ iterations of validation results was selected. As under payload rate 0.4 bpp, we directly trained YeNet under payload rate 0.2 bpp instead of adapting curriculum learning. For each classifier of YeNet, training was executed $700k$ iterations, and the model with the best steganalystic performance from the last $100k$ iterations of validation results was selected.
      \item SRNet\cite{Boroumand2019_SRNet}: this CNN steganalystic network was used as non-target CNN classifier. 
      \item SRM\cite{Fridrich2012}, maxSRMd2\cite{Denemark2014Selection} and PDASS\cite{Tan2017_PDASS}: ensemble classifiers (ECs)\cite{Kodovsky2012} with these three feature sets were used to evaluate performance of resisting on conventional classifiers. 
      \item Unless specified, other settings of CNN classifiers were as same as what were described in the corresponding papers.
    \end{itemize}
  \item[4)] Compared methods: the current state of the art steganography by using adversarial samples ADV-EMB\cite{Tang2019_ADS} and its applied Min-Max distortion function\cite{Bernard2019_MM} method were employed. 
    \begin{itemize}
      \item As described in the corresponding paper, we executed nine rounds of Min-Max processes.
    \end{itemize}
  \item[5)] For STCs, we selected the scale size $H=10$ of the parity-check matrix $\mathbb{H}$.
  \item[6)] Evaluation performance. The steganographic performances were evaluated by detection error $P_E$.
  \begin{equation}\label{eq:pe}
    P_E=\frac{P_{FA}+P_{MD}}{2}
  \end{equation}
where $P_{FA}$ and $P_{MD}$ are the false-alarm and missed-detection probabilities respectively. 
\end{enumerate}

\begin{figure}[!tbp]
\centerline{\includegraphics[width=0.8\columnwidth]{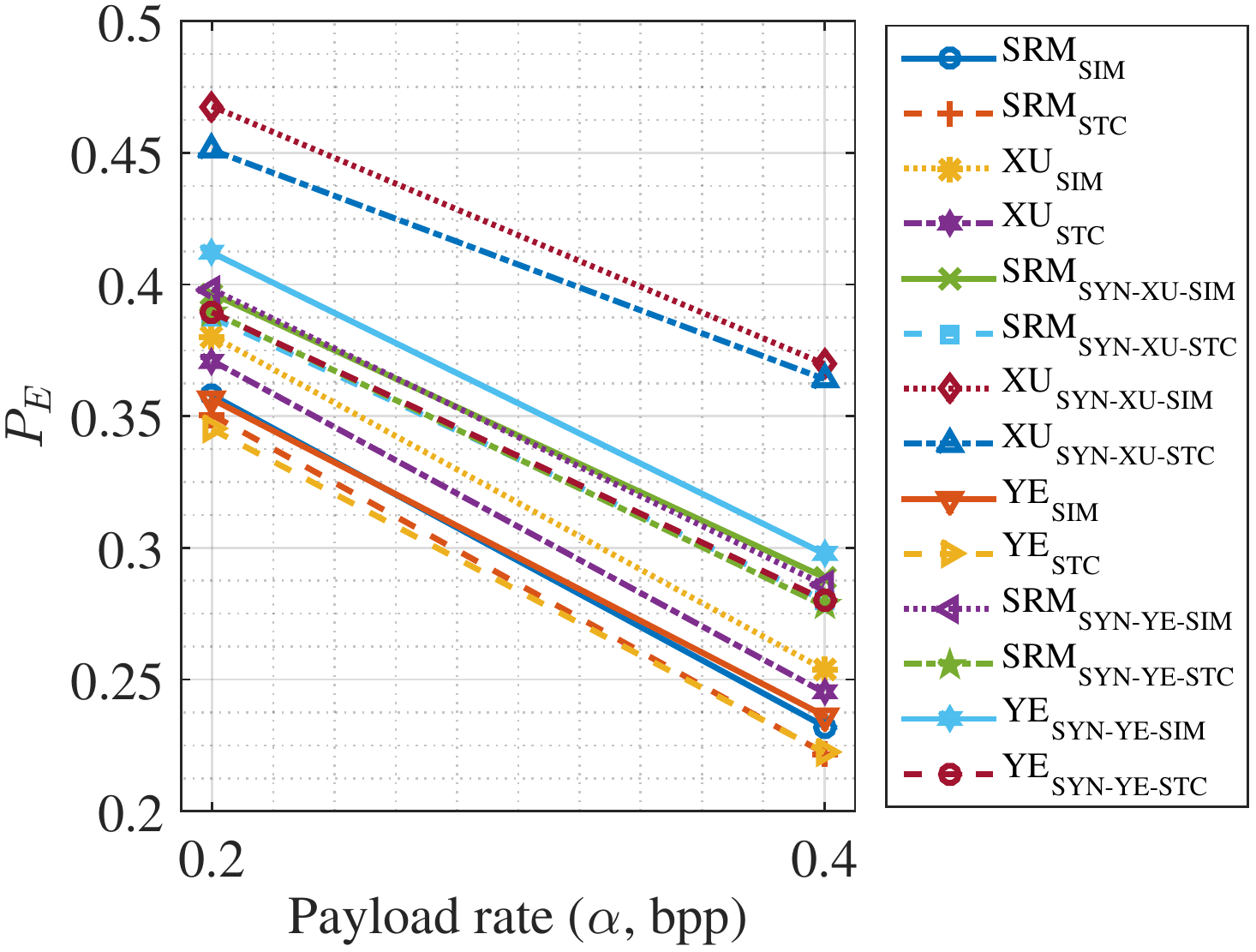}}
\caption{Comparison of detect errors $P_E$ from the simulator and STCs by using S-UNIWARD on BOSS256.}
\label{fig:stc_sun}
\end{figure}

\subsection{Comparison for the simulator and STCs}\label{sec:exper:ssec:stc}
\label{sec:exper:ssec:stcs}

We used the simulator and STCs to produce stego images. The Fig. \ref{fig:stc_sun} reports comparison of detection errors from the simulator and STCs by using S-UNIWARD steganographic scheme under payload rates 0.2 bpp and 0.4 bpp on image set BOSS256. Where XU, YE and SRM represent classifiers XuNet, YeNet and ECs trained by SRM features respectively, and subscripts represent stego image sets. SIM and STC indicate that stego images were created by using the simulator and STCs respectively, and ITE represents that stego images were created by using the proposed method. It is observed that performances of stego images created by the simulator and STCs are consistent. For the space limitation, we only reported results of stego images created by the simulator in rest experiments unless otherwise specified.

\subsection{Impacts of the celling of adversarial intensity factor}
\label{sec:exper:ssec:gamma}

ITE-SYN includes two operations, applying CMD profile and iteratively increasing adversarial perturbations. Employing XuNet as the target CNN classifier and S-UNIWARD as the cost scheme, we produced stego images by using different celling intensity factors under 0.4 bpp on image set BOSS256. There were following spacified stego images were created.
\begin{itemize}
  \item CMD: by using the CMD method described in Algorithm \ref{alg:syn_emb}.
  \item ADV: by using the fixed adversarial intensity $\gamma=1+\gamma_{max}$ to generate adversarial perturbations.
  \item CMD-ADV: using the CMD method and then the generated adversarial perturbations with fixed adversarial intensity $\gamma=1+\gamma_{max}$. The target CNN classifiers were trained by images with CMD.
  \item ITE: by using iteratively increased adversarial intensity.
  \item CMD-ITE: by using the CMD method and then iteratively increased adversarial intensity. The target CNN classifiers were trained by images with CMD.
\end{itemize}

\begin{figure}[tbp]
  \centering
    \begin{minipage}[]{0.48\linewidth}
      \centering
      \centerline{\includegraphics[width=1.0\linewidth]{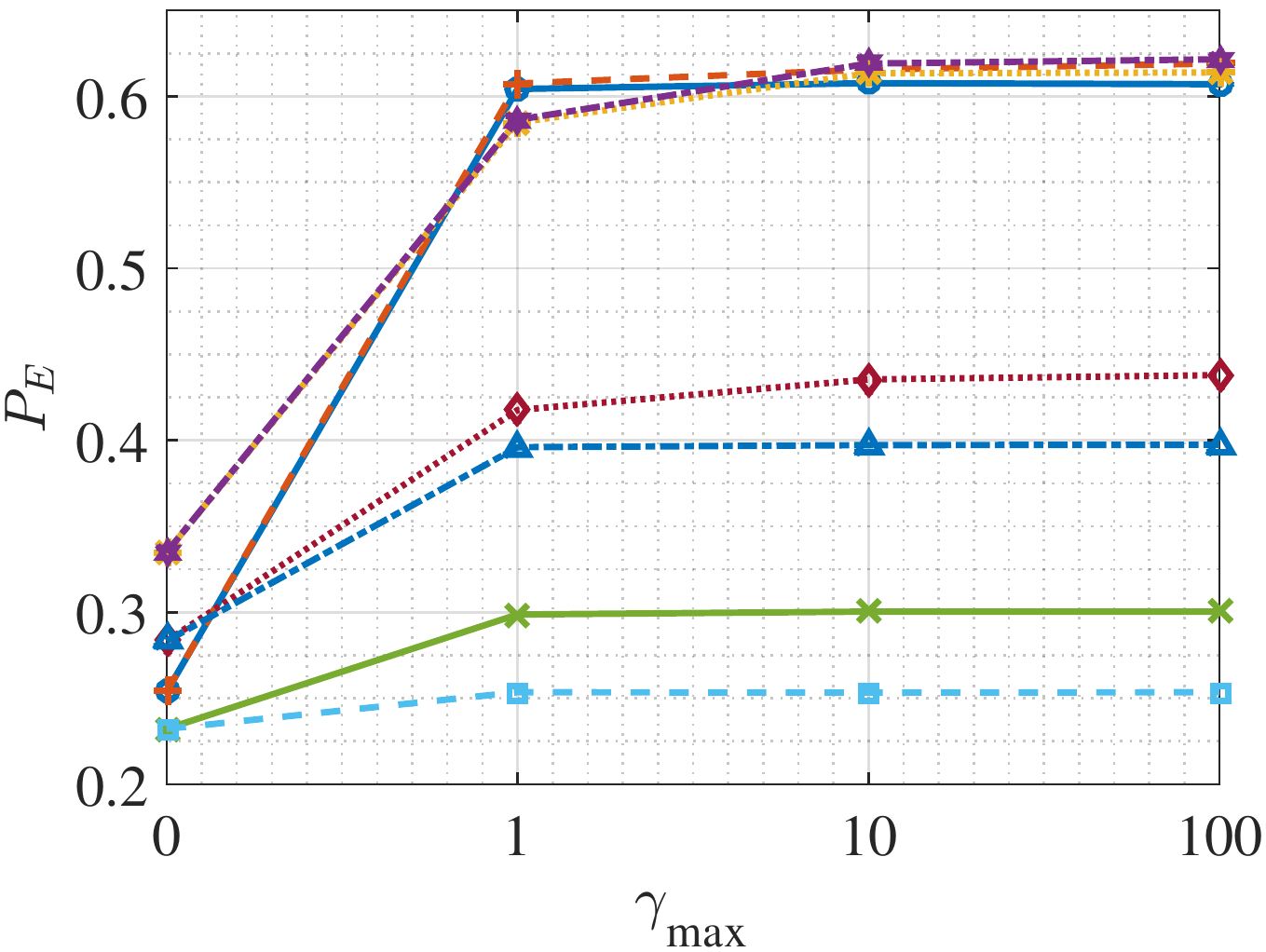}}
      \centerline{(a)}\medskip
    \end{minipage}
    \hfill
    \begin{minipage}[]{0.48\linewidth}
      \centering
      \centerline{\includegraphics[width=1.0\linewidth]{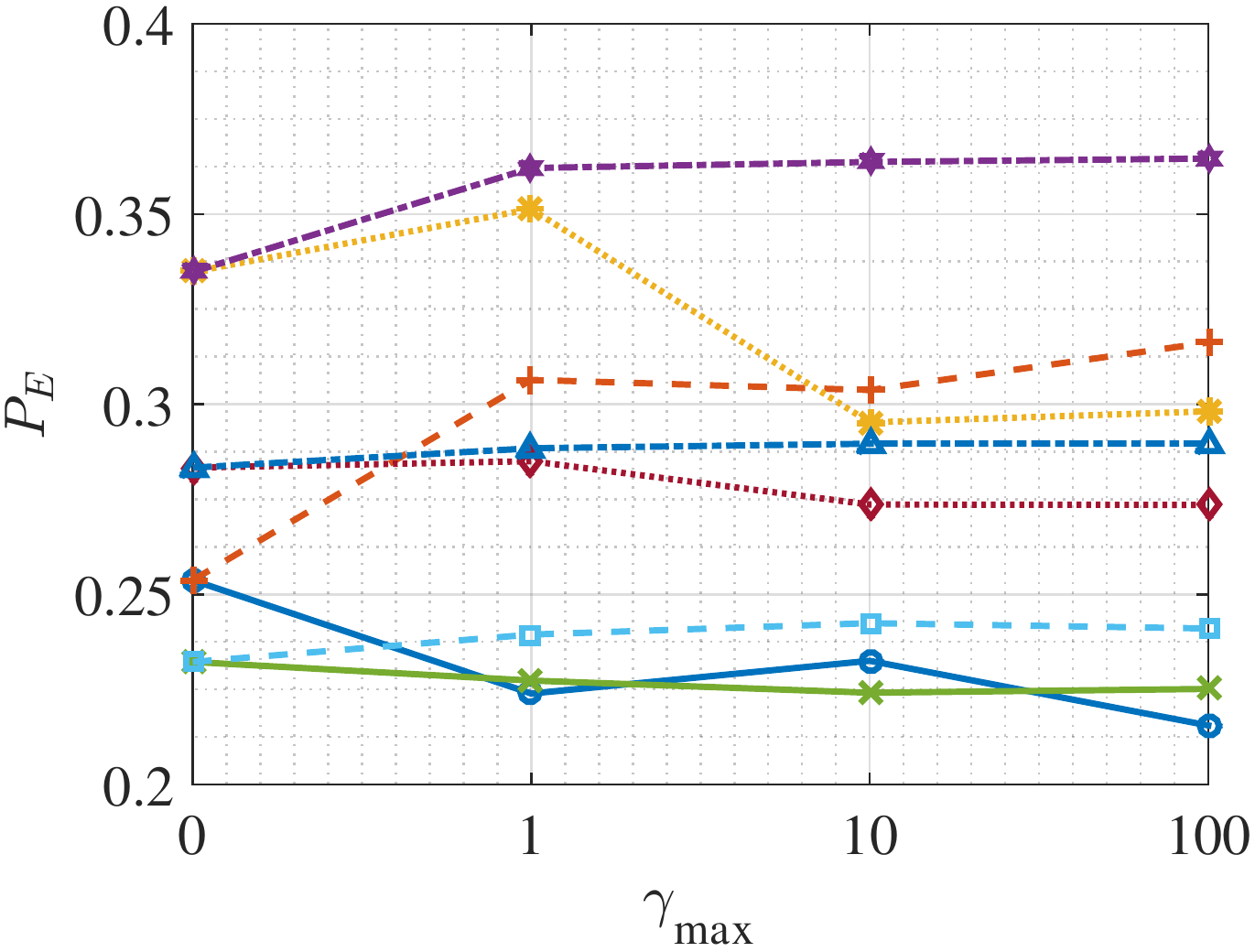}}
      \centerline{(b)}\medskip
    \end{minipage}
    \vfill
    \begin{minipage}[]{0.96\linewidth}
      \centering
      \centerline{\includegraphics[width=1.0\linewidth]{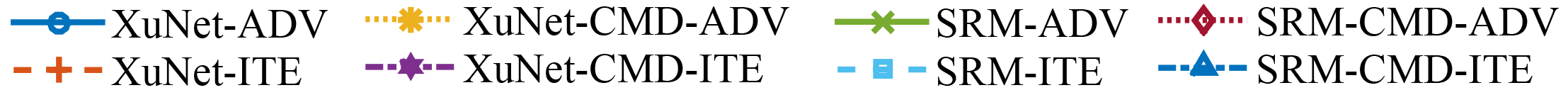}}
    \end{minipage}
  \caption{Performances $P_E$ of different adversarial intensities with S-UNIWARD. (a) Performances of deceiving original classifiers. (b)  Performances of resisting on re-trained classifiers.}
  \label{fig:pe_intensity}
\end{figure}

\begin{figure}[tbp]
  \centering
    \begin{minipage}[]{0.48\linewidth}
      \centering
      \centerline{\includegraphics[width=1.0\linewidth]{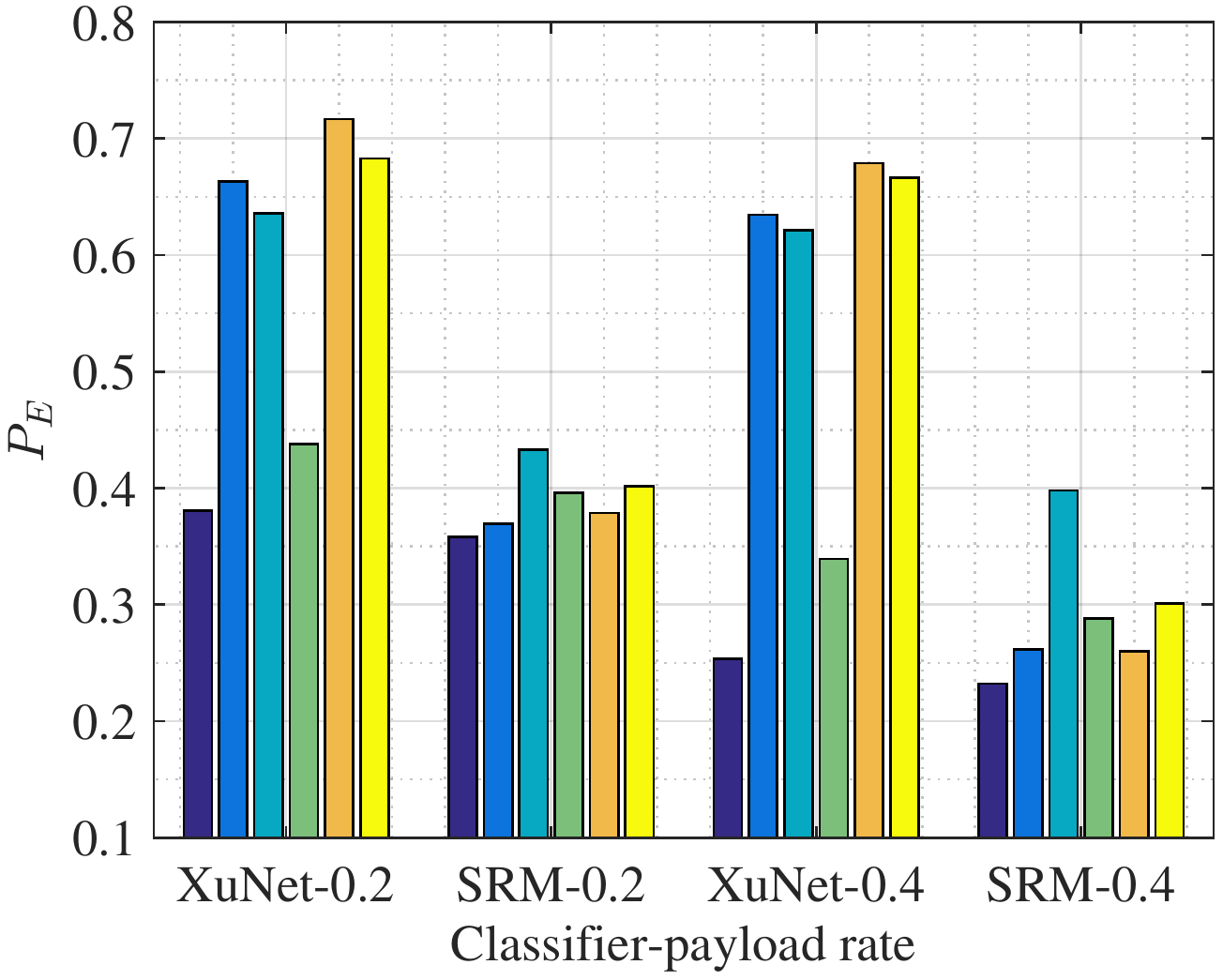}}
      \centerline{(a)}\medskip
    \end{minipage}
    \hfill
    \begin{minipage}[]{0.48\linewidth}
      \centering
      \centerline{\includegraphics[width=1.0\linewidth]{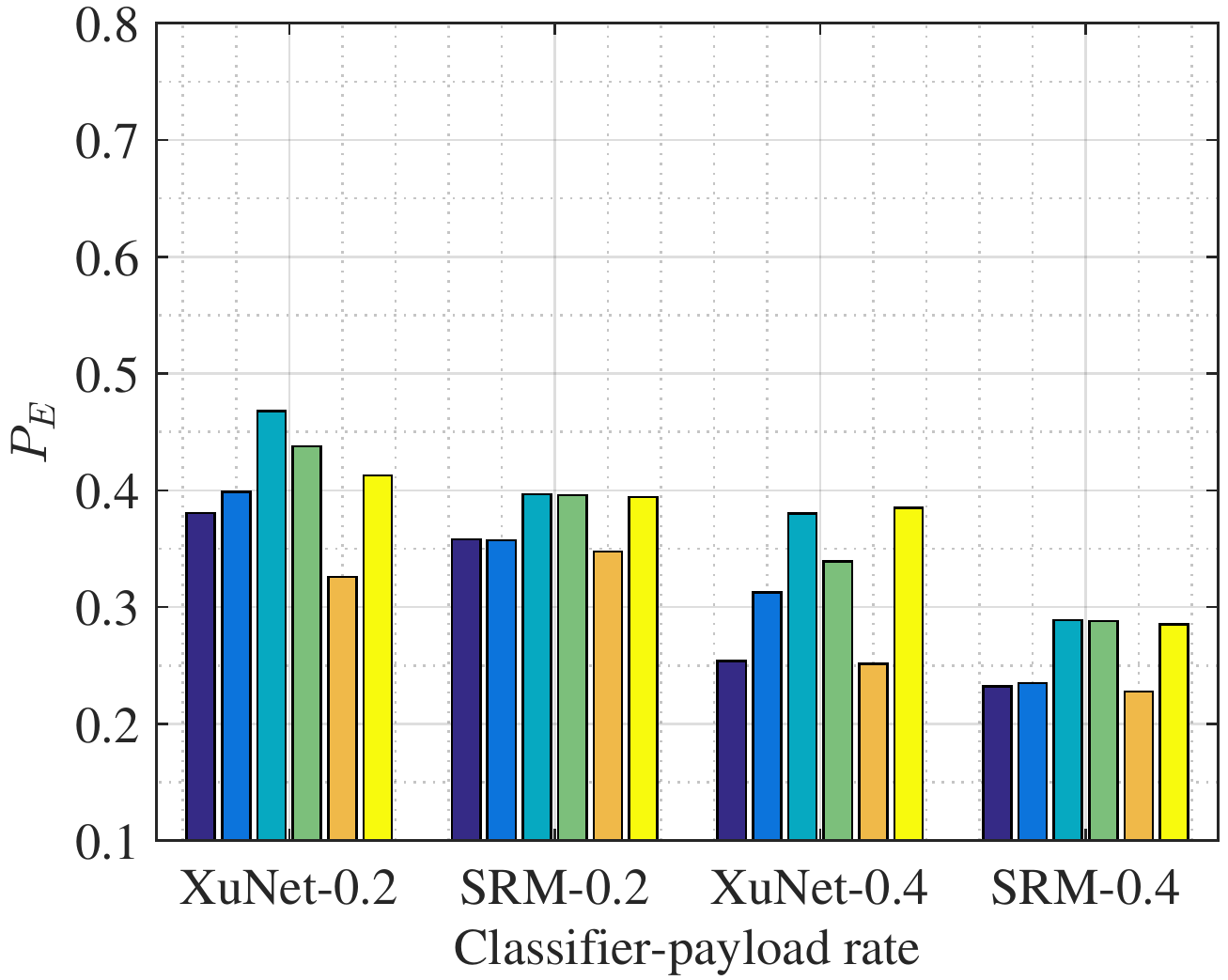}}
      \centerline{(b)}\medskip
    \end{minipage}
    \vfill
    \begin{minipage}[]{0.8\linewidth}
      \centering
      \centerline{\includegraphics[width=1.0\linewidth]{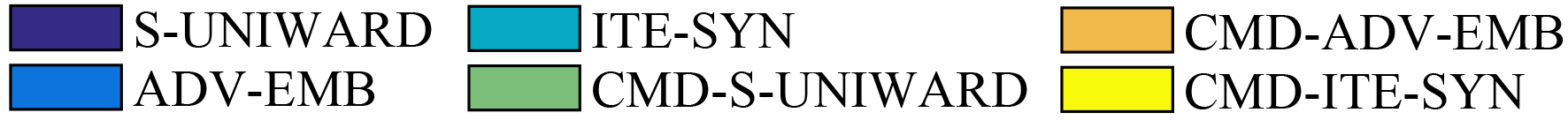}}
    \end{minipage}
  \caption{Comparison for performances $P_E$ of classifiers with non-CMD method from CMD method. The target CNN classifiers are XuNet and cost function is S-UNIWARD. (a) Performances of deceiving original classifiers. (b)  Performances of resisting on re-trained classifiers.}
  \label{fig:pe_cmd}
\end{figure}

Here, for each image we selected the top-left sub-image as the start sub-image and only applied adversarial perturbations onto the top-left sub-image. Results are shown in  Fig. \ref{fig:pe_intensity}, where $\langle classifier\rangle$-$\langle stego\quad image\quad set\rangle$ are used to indicate classifiers and corresponding stego image sets. 
There are following observations.
\begin{itemize}
  \item[1)] For deceiving the target CNN classifier, ITE versions perform superior or similar to corresponding ADV versions. But for resisting on detection of original feature-based steganalystic models, ADV versions perform superior to corresponding ITE versions. 
  \item[2)] For adversarial training, ITE versions always perform superior to corresponding ADV versions.
  \item[3)] For ADV, the attack is insufficient for a small $\gamma$, but is overdone for a big $\gamma$. It is obvious when resist on adversarial trained models.
  \item[4)] For purely iterative adversarial perturbations, the gain is insufficient even thought the $\gamma_{max}$ is big enough.
\end{itemize}

Actually, for iterative adversarial perturbations, including ITE and CMD-ITE, too big $\gamma_{max}$ is not necessary, because most adversarial stego images are produced at proper intensity factors, which is illustrated in section \ref{sec:exper:ssec:time}.
Refer to these results, it is reasonable that we select $\gamma_{max}=10$.

For investigating performances of attacking CNN classifiers trained by non-CMD and CMD version stego images, we trained XuNet models for non-CMD and CMD stego images with S-UNIWARD under different payload rates respectively. And then we created adversarial stego images by using ADV-EMB\cite{Tang2019_ADS} and the proposed method to attack those trained CNN classifiers respectively. Results of comparison of performances are shown in Fig. \ref{fig:pe_cmd}, where $\textit{CMD}-<\bullet>$ represent stego images created with CMD method or adversarial stego images created to attack CMD version CNN models, and $<\bullet>$ represent the cost function or adversarial methods. It is observed that ADV-EMB with CMD version performs better when it deceives the target CNN classifiers. However, it is worthly noticed as followings. 
\begin{itemize}
  \item[1)] For deceiving original non-target classifiers, ADV-EMB with CMD underperforms corresponding baselines. Meanwhile, ITE-SYN overperforms corresponding baselines.
  \item[2)] For resisting on adversarial trained classifiers, ADV-EMB with CMD underperforms corresponding baseline even non-CMD baselines. Meanwhile, ITE-SYN with non-CMD performs superrior to corresponding baselines and ADV-EMB, neither inferior to corresponding ITE-SYN with CMD nor CMD baselines.
\end{itemize}
Therefore, it is insignificant that creating adversarial stego images to attack CNN classifiers trained by CMD version stego images. In rest of this paper, all target CNN classifiers were trained by non-CMD version stego images unless otherwise specified.

\subsection{Comparison to state-of-the-art methods}
\label{sec:exper:ssec:comparison}

\begin{figure*}[tbp]
  \centering
    \begin{minipage}[]{0.24\linewidth}
      \centering
      \centerline{\includegraphics[width=1.0\linewidth]{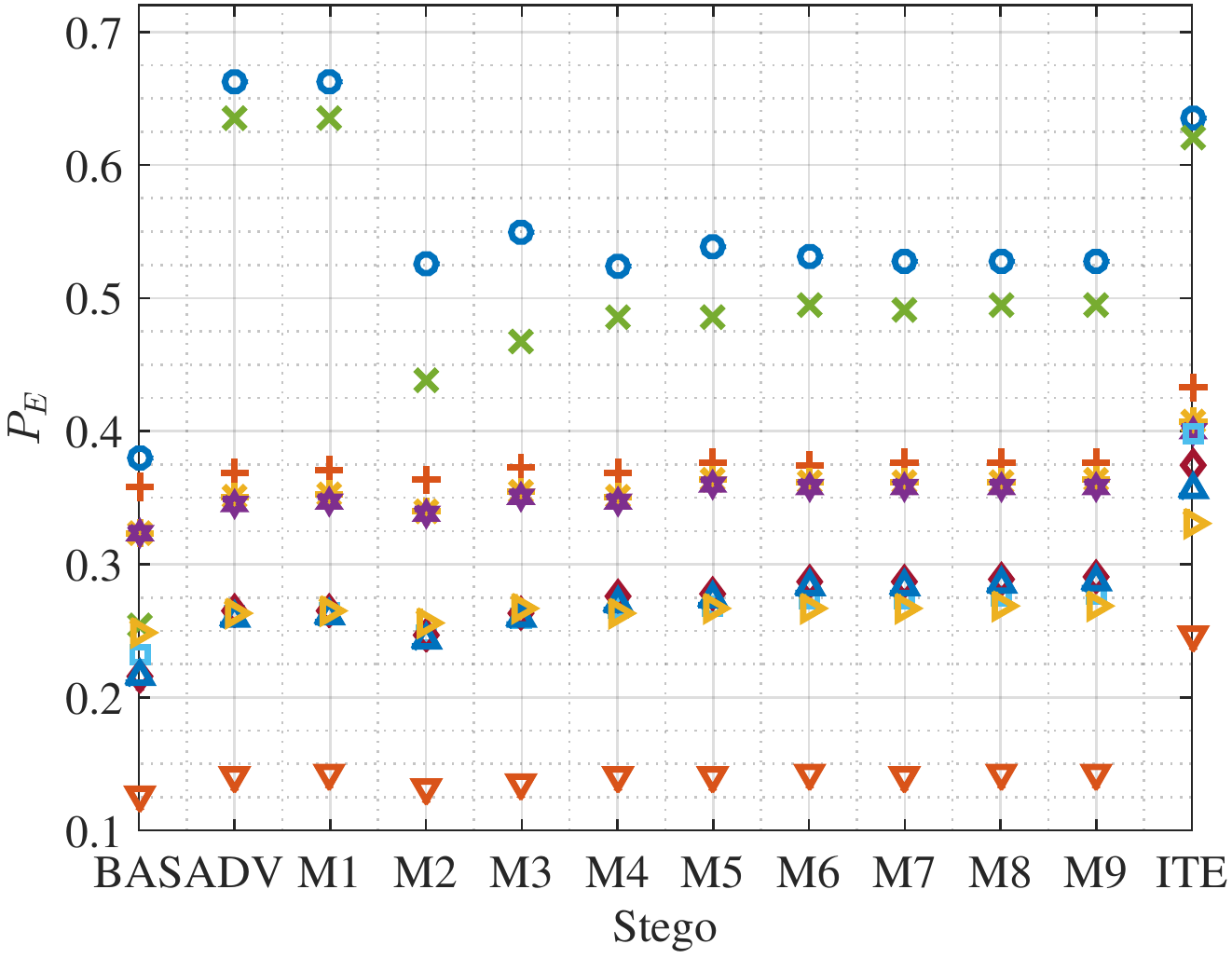}}
      \centerline{(a) S-UNIWARD}\medskip
    \end{minipage}
    \hfil
    \begin{minipage}[]{0.24\linewidth}
      \centering
      \centerline{\includegraphics[width=1.0\linewidth]{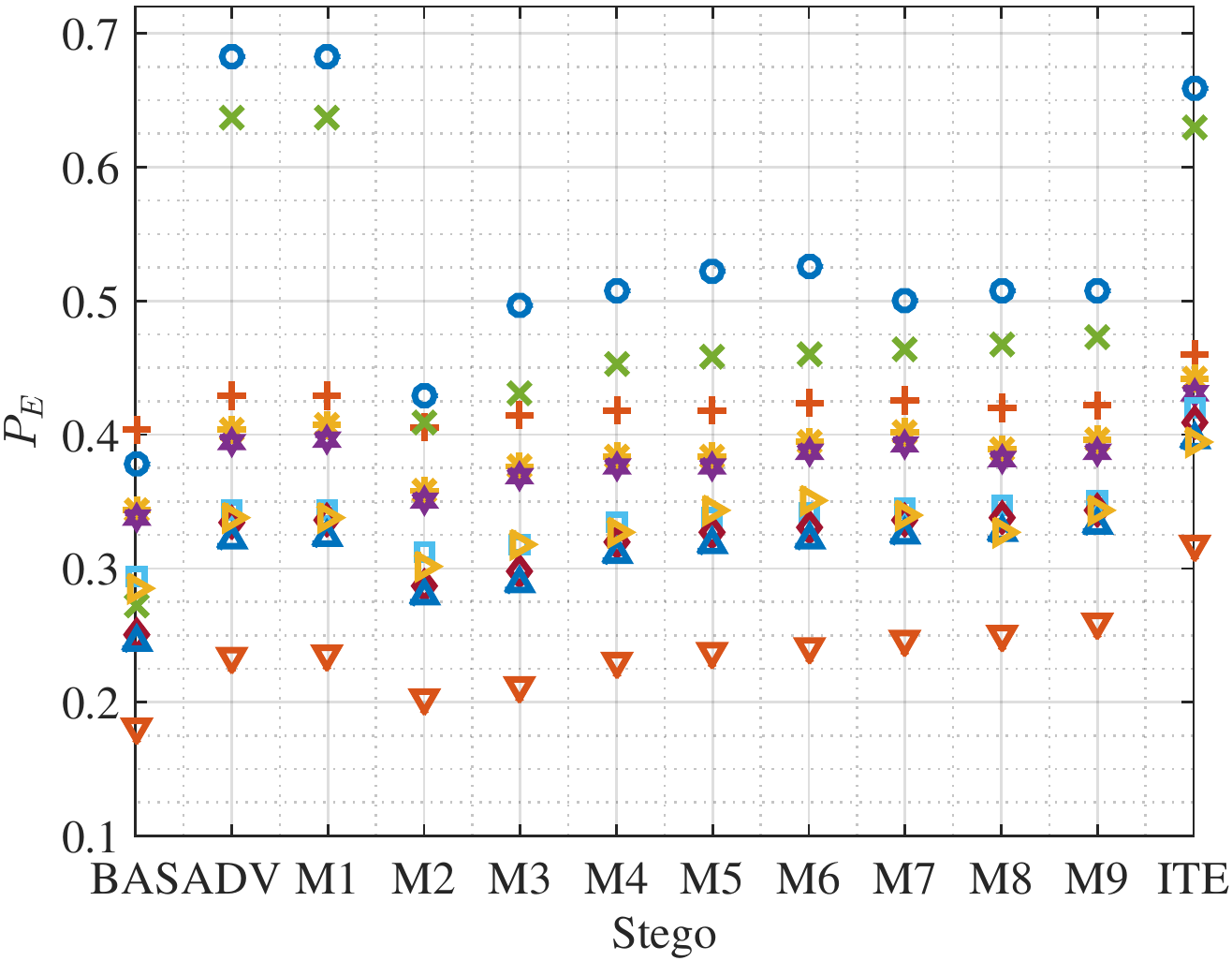}}
      \centerline{(b) HILL}\medskip
    \end{minipage}
    \hfil
    \begin{minipage}[]{0.24\linewidth}
      \centering
      \centerline{\includegraphics[width=1.0\linewidth]{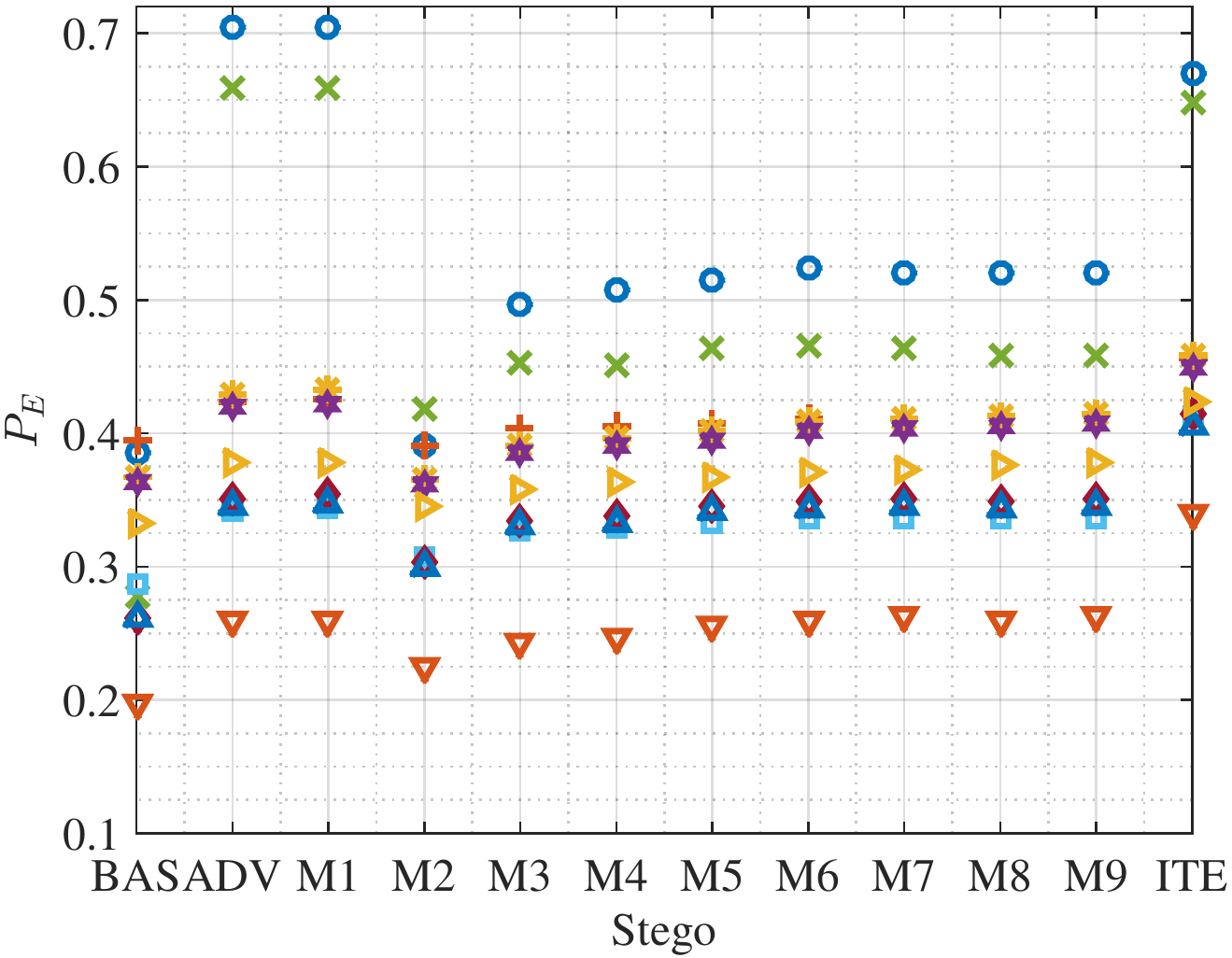}}
      \centerline{(c) MiPOD}\medskip
    \end{minipage}
    \hfil
    \begin{minipage}[]{0.24\linewidth}
      \centering
      \centerline{\includegraphics[width=1.0\linewidth]{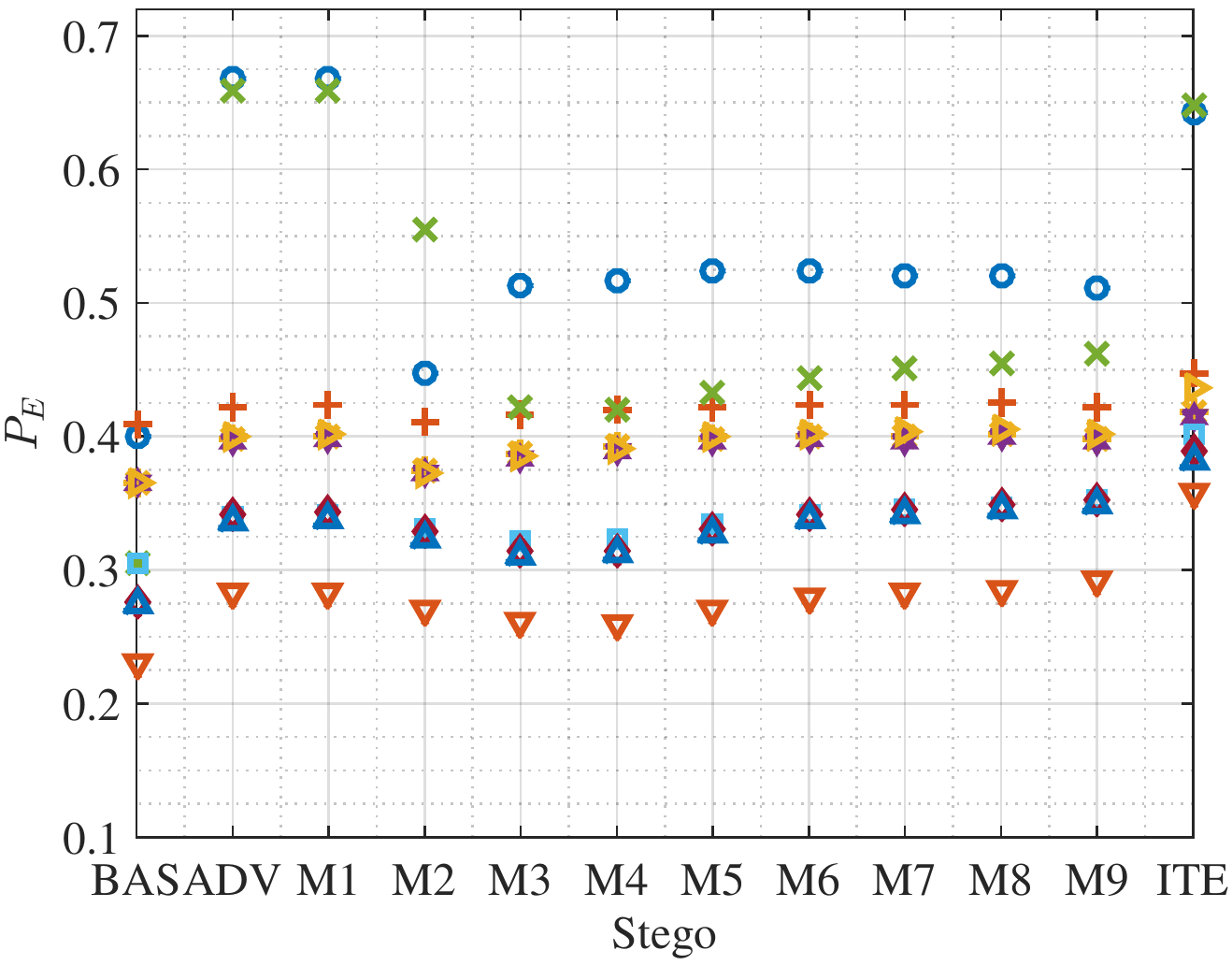}}
      \centerline{(d) MGR}\medskip
    \end{minipage}
    \vfil
    \begin{minipage}[]{0.8\linewidth}
      \centering
      \centerline{\includegraphics[width=1.0\linewidth]{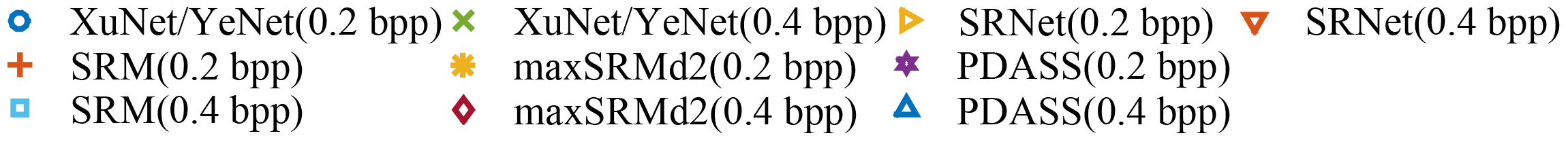}}
      \centerline{}\medskip
    \end{minipage}
  \caption{Performances $P_E$ of deceiving original classifiers with the target XuNet for BOSS256.}
  \label{fig:pe_adv_xu}
\end{figure*}
\begin{figure*}[tbp]
  \centering
    \begin{minipage}[]{0.24\linewidth}
      \centering
      \centerline{\includegraphics[width=1.0\linewidth]{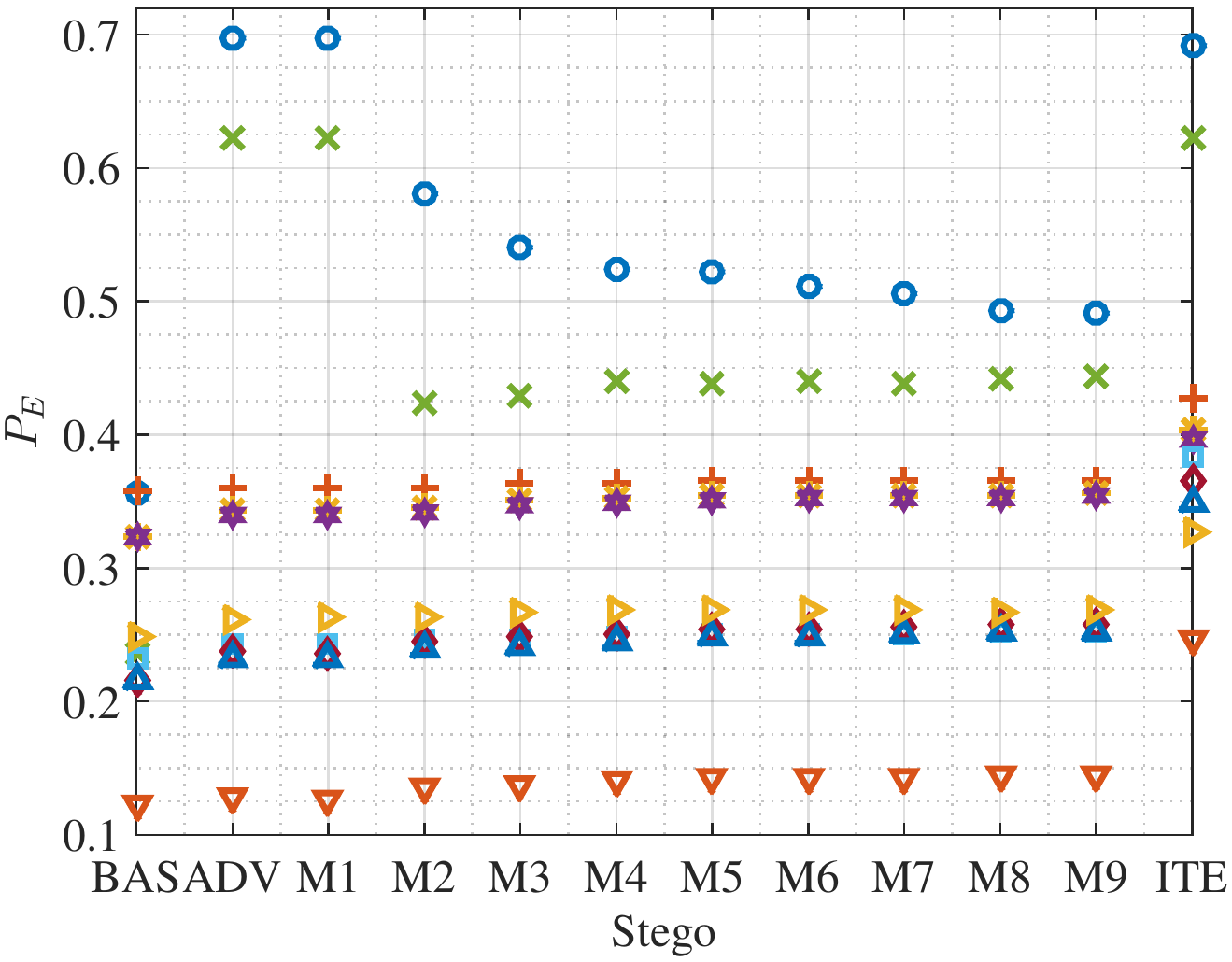}}
      \centerline{(a) S-UNIWARD}\medskip
    \end{minipage}
    \hfil
    \begin{minipage}[]{0.24\linewidth}
      \centering
      \centerline{\includegraphics[width=1.0\linewidth]{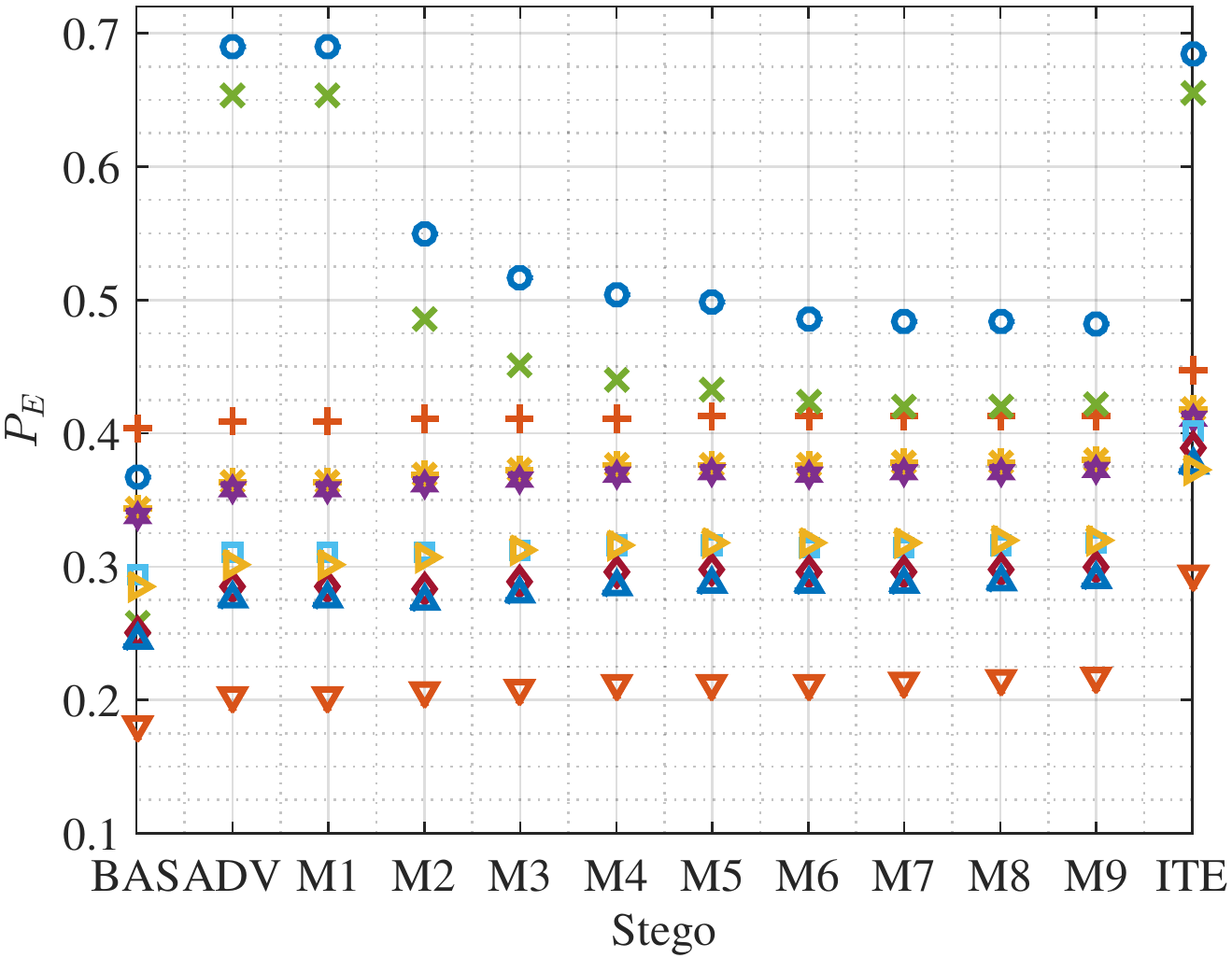}}
      \centerline{(b) HILL}\medskip
    \end{minipage}
    \hfil
    \begin{minipage}[]{0.24\linewidth}
      \centering
      \centerline{\includegraphics[width=1.0\linewidth]{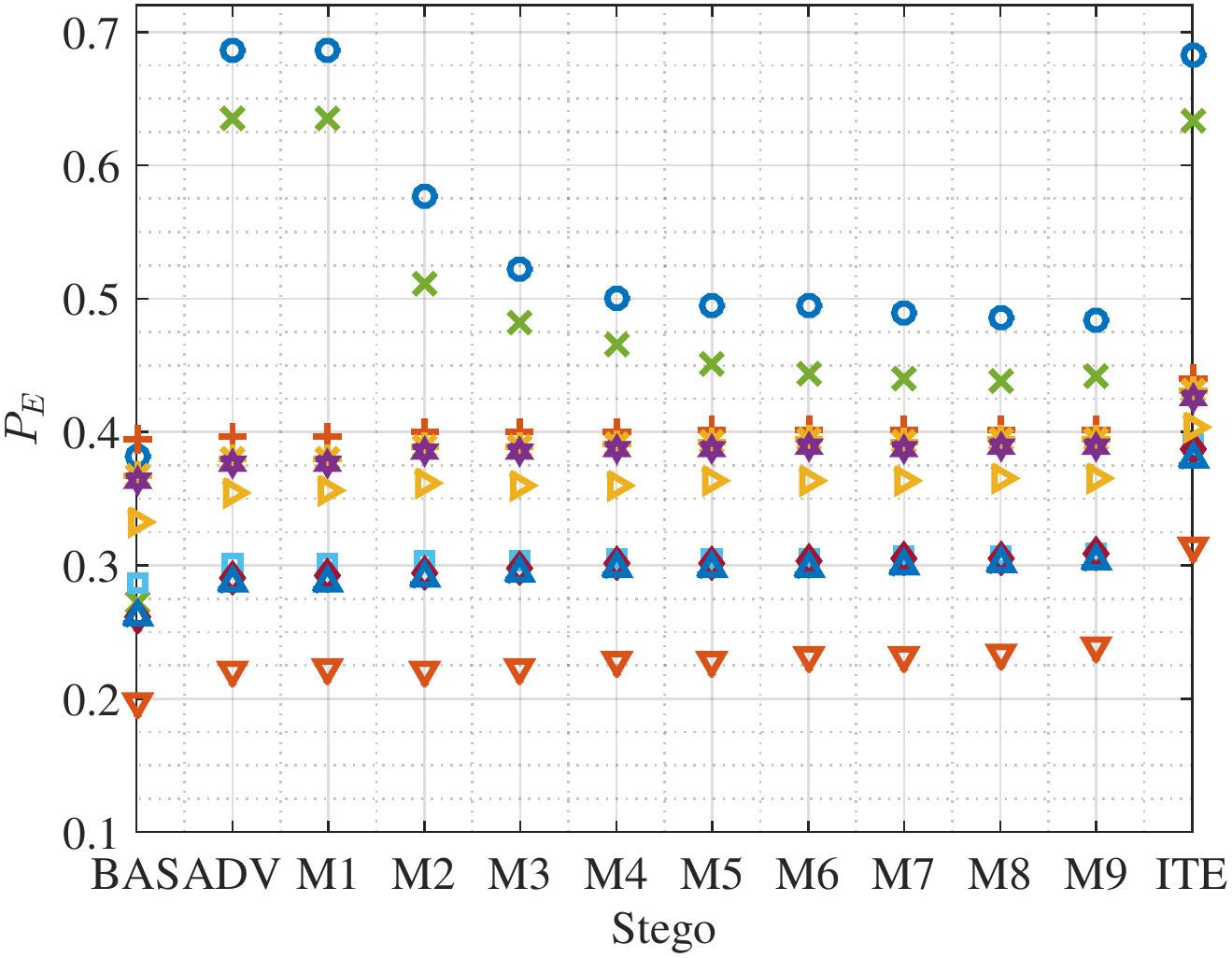}}
      \centerline{(c) MiPOD}\medskip
    \end{minipage}
    \hfil
    \begin{minipage}[]{0.24\linewidth}
      \centering
      \centerline{\includegraphics[width=1.0\linewidth]{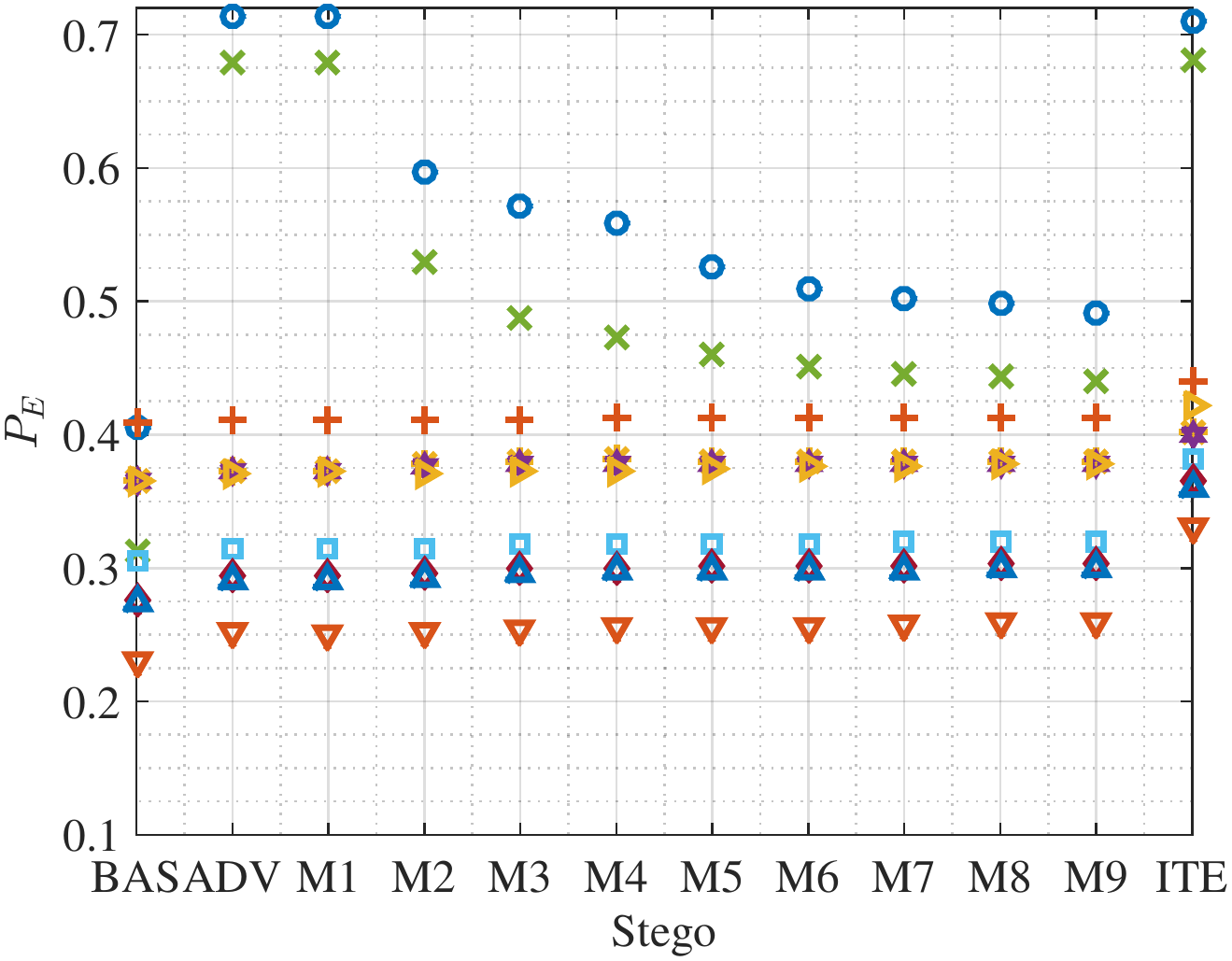}}
      \centerline{(d) MGR}\medskip
    \end{minipage}
    \vfil
    \begin{minipage}[]{0.8\linewidth}
      \centering
      \centerline{\includegraphics[width=1.0\linewidth]{fig_legend_adv_n}}
      \centerline{}\medskip
    \end{minipage}
  \caption{Performances $P_E$ of deceiving original classifiers with the target YeNet for BOSS256.}
  \label{fig:pe_adv_ye}
\end{figure*}

\begin{figure*}[tbp]
  \centering
    \begin{minipage}[]{0.23\linewidth}
      \centering
      \centerline{\includegraphics[width=1.0\linewidth]{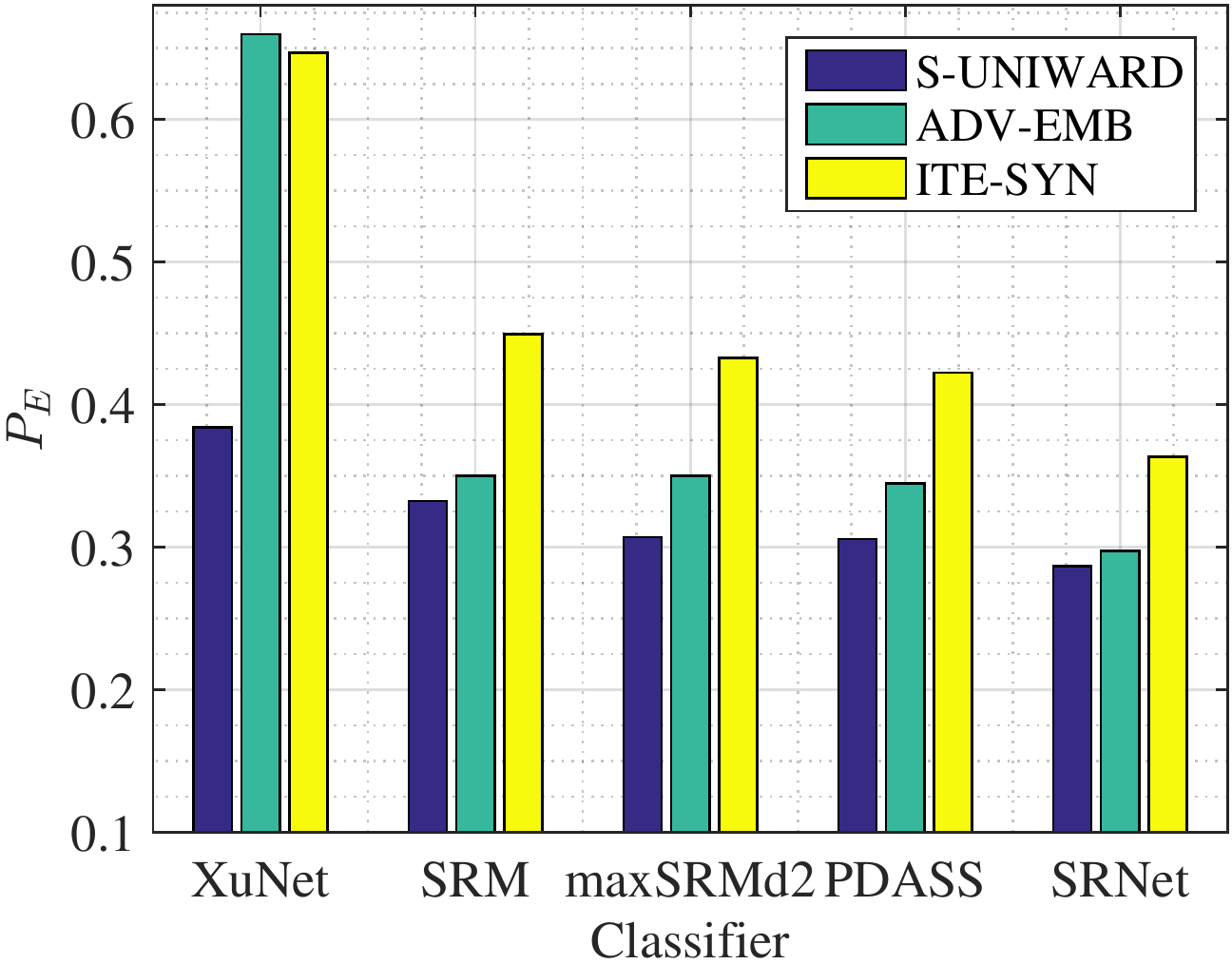}}
      \centerline{(a)}\medskip
    \end{minipage}
    \hfil
    \begin{minipage}[]{0.23\linewidth}
      \centering
      \centerline{\includegraphics[width=1.0\linewidth]{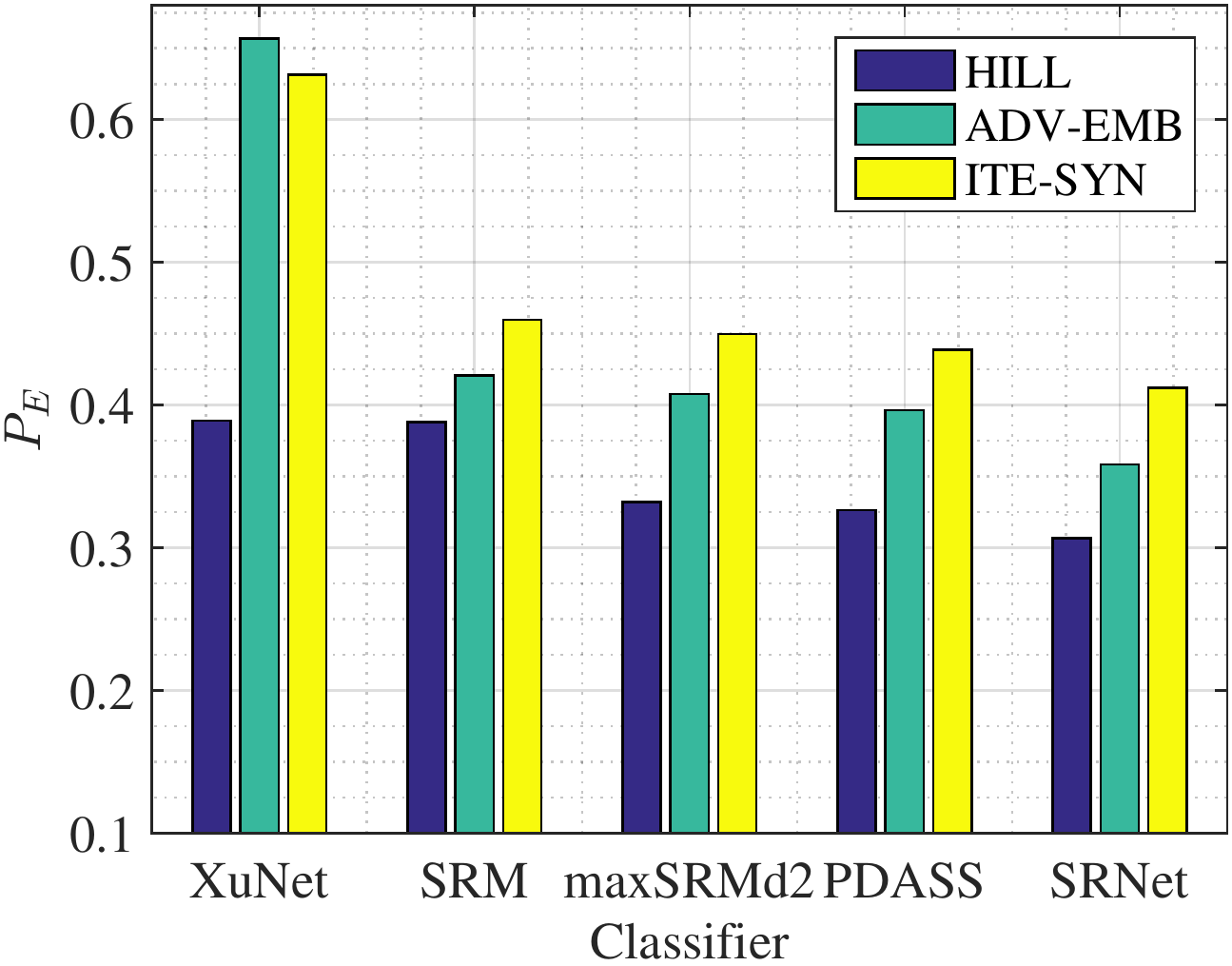}}
      \centerline{(b)}\medskip
    \end{minipage}
    \hfil
    \begin{minipage}[]{0.23\linewidth}
      \centering
      \centerline{\includegraphics[width=1.0\linewidth]{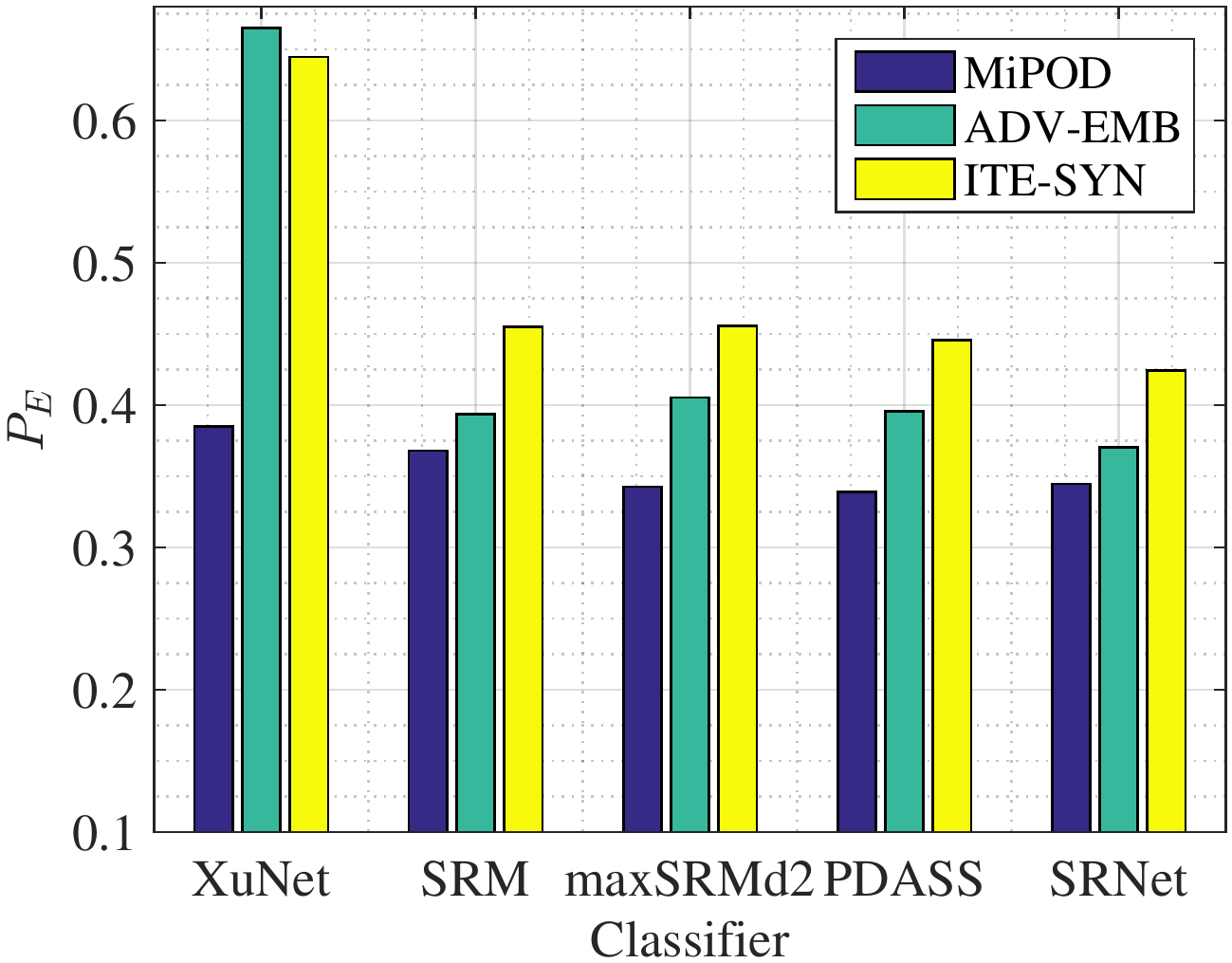}}
      \centerline{(c)}\medskip
    \end{minipage}
    \hfil
    \begin{minipage}[]{0.23\linewidth}
      \centering
      \centerline{\includegraphics[width=1.0\linewidth]{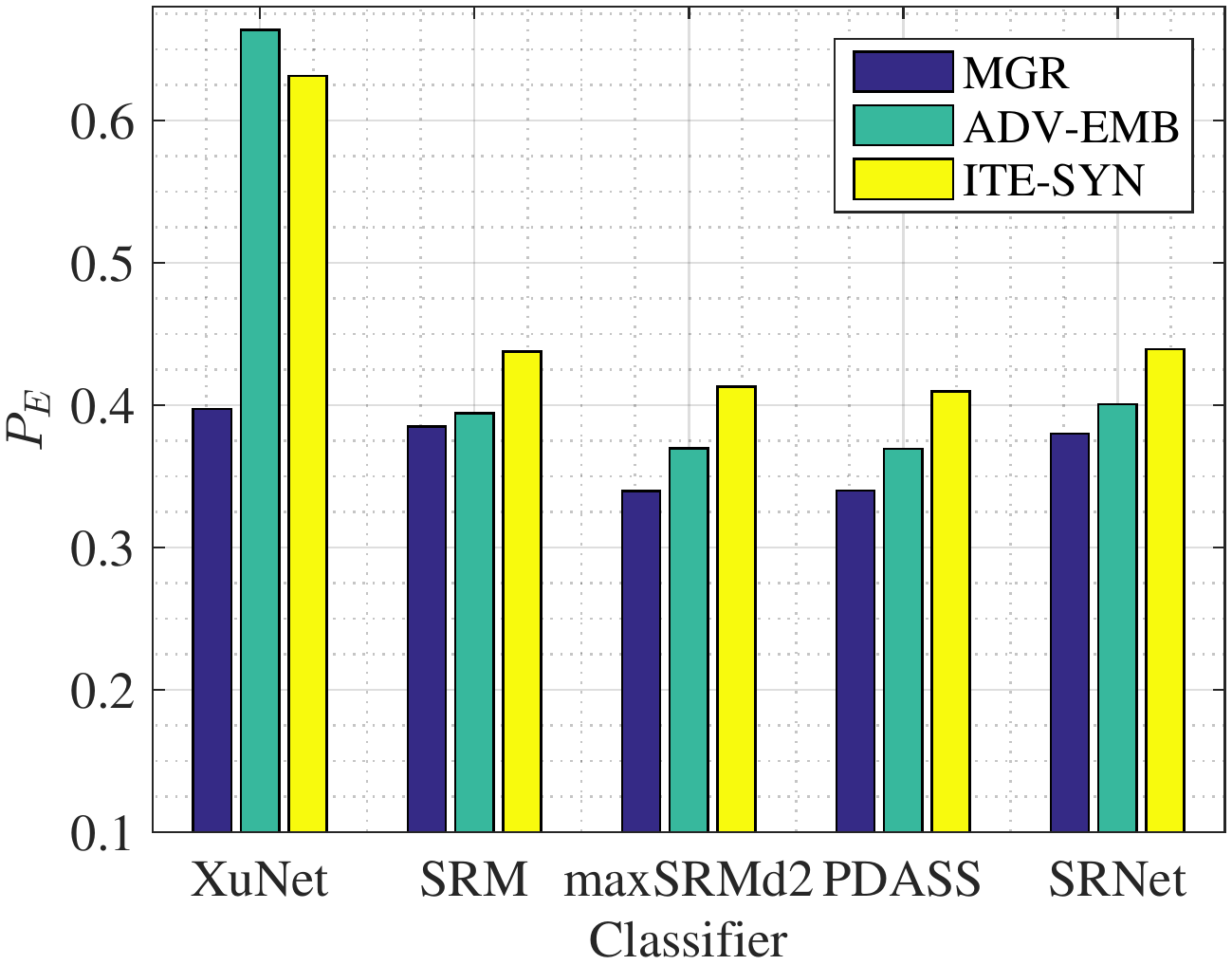}}
      \centerline{(d)}\medskip
    \end{minipage}
    \vfil
    \begin{minipage}[]{0.23\linewidth}
      \centering
      \centerline{\includegraphics[width=1.0\linewidth]{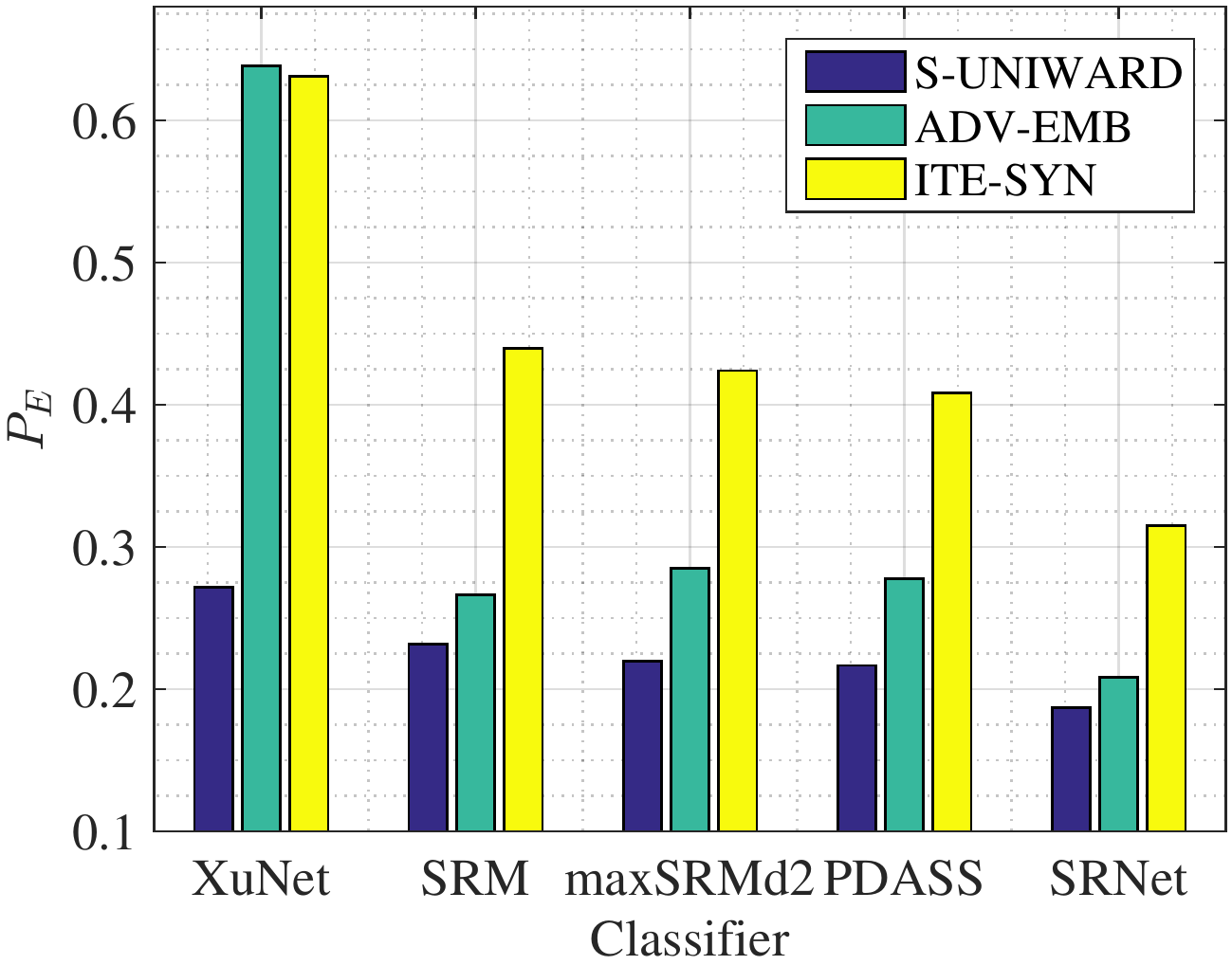}}
      \centerline{(e)}\medskip
    \end{minipage}
    \hfil
    \begin{minipage}[]{0.23\linewidth}
      \centering
      \centerline{\includegraphics[width=1.0\linewidth]{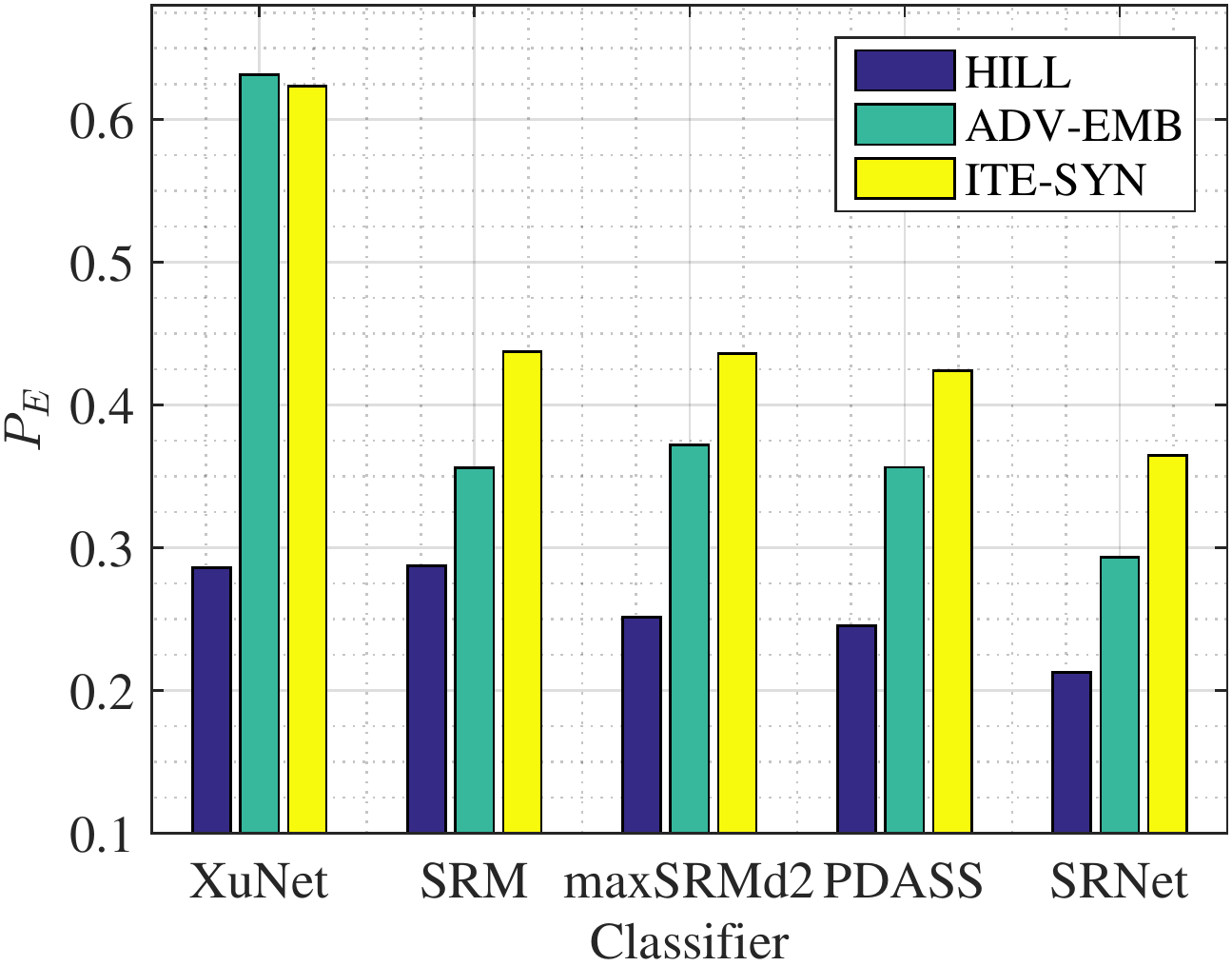}}
      \centerline{(f)}\medskip
    \end{minipage}
    \hfil
    \begin{minipage}[]{0.23\linewidth}
      \centering
      \centerline{\includegraphics[width=1.0\linewidth]{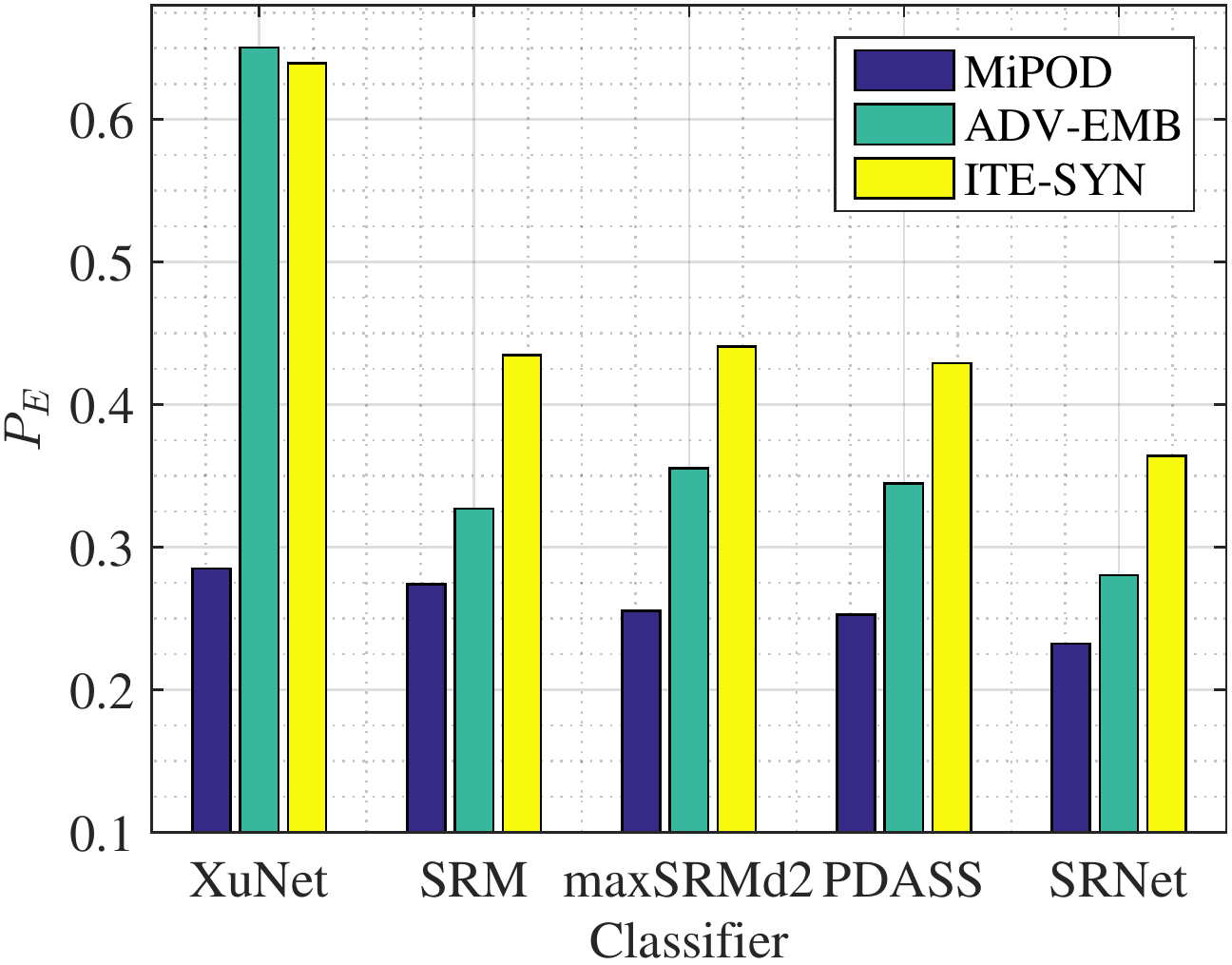}}
      \centerline{(g)}\medskip
    \end{minipage}
    \hfil
    \begin{minipage}[]{0.23\linewidth}
      \centering
      \centerline{\includegraphics[width=1.0\linewidth]{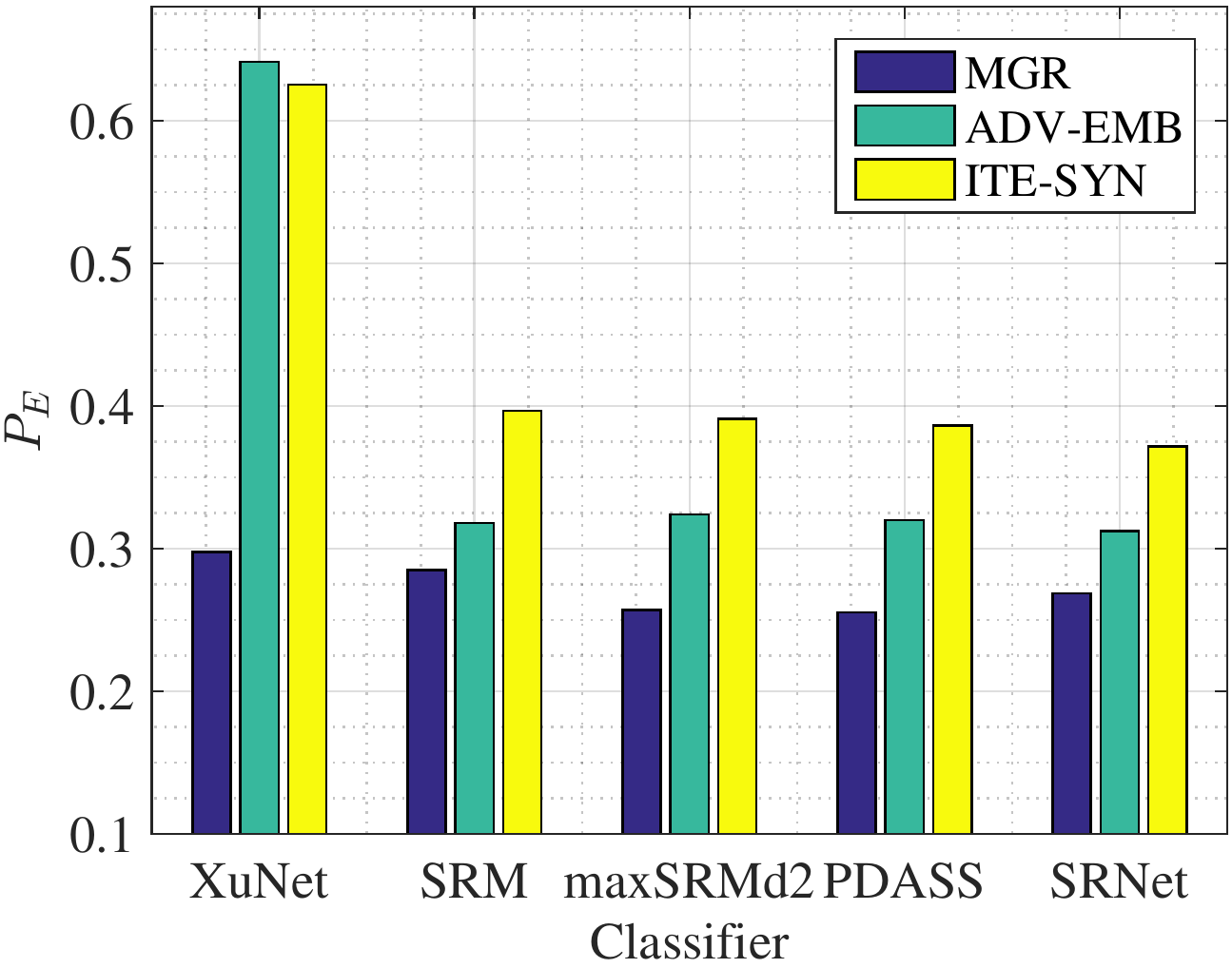}}
      \centerline{(h)}\medskip
    \end{minipage}
  \caption{Performances $P_E$ of deceiving original classifiers with the target XuNet for ALASKA256. For (a)-(d), stego images and adversarial stego images were produced with corresponding steganographic schemes under payload rate 0.2 bpp respectively. And (e)-(h) are under payload rate 0.4 bpp respectively.}
  \label{fig:pe_adv_xu_alaska}
\end{figure*}
\begin{figure*}[tbp]
  \centering
    \begin{minipage}[]{0.23\linewidth}
      \centering
      \centerline{\includegraphics[width=1.0\linewidth]{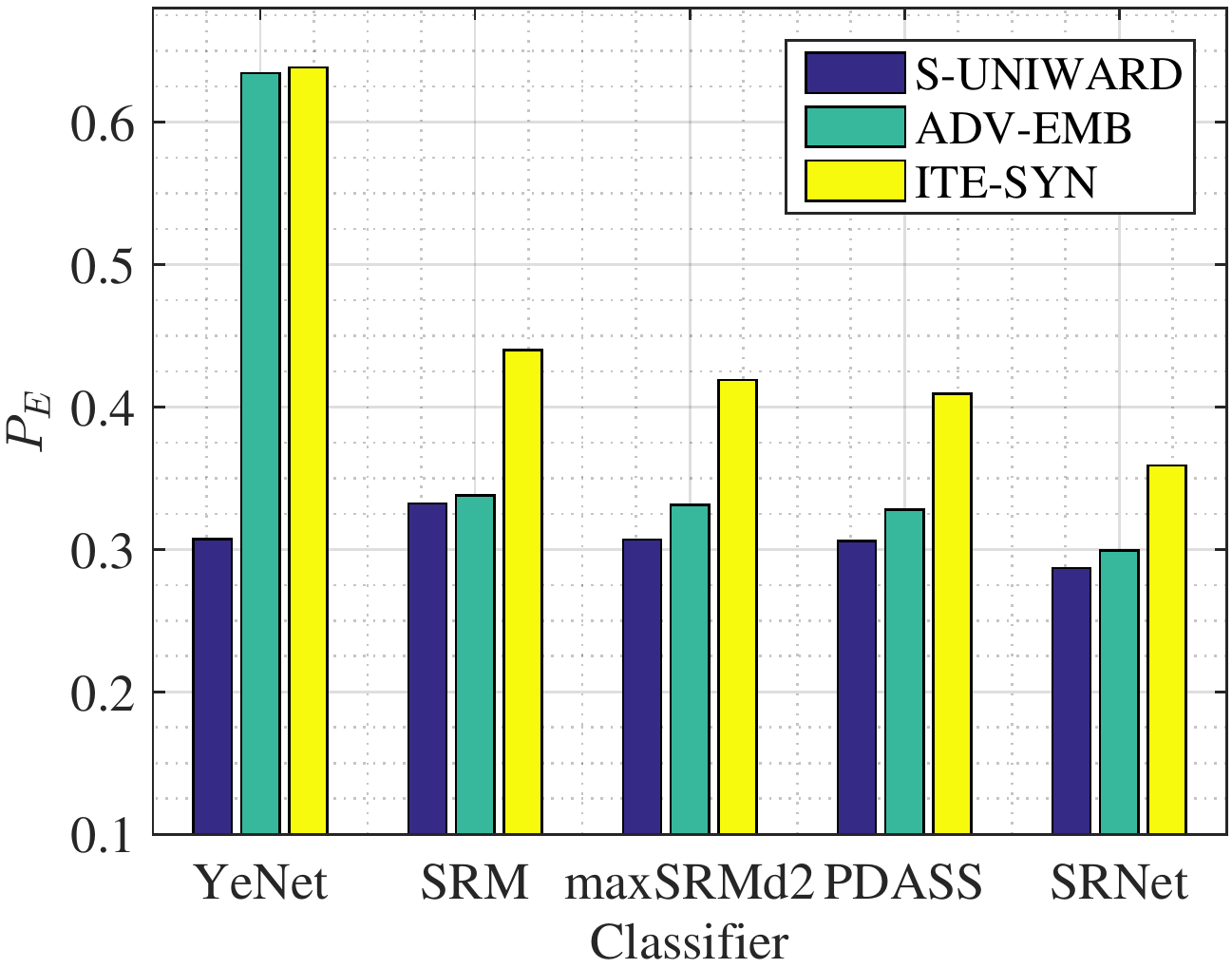}}
      \centerline{(a)}\medskip
    \end{minipage}
    \hfil
    \begin{minipage}[]{0.23\linewidth}
      \centering
      \centerline{\includegraphics[width=1.0\linewidth]{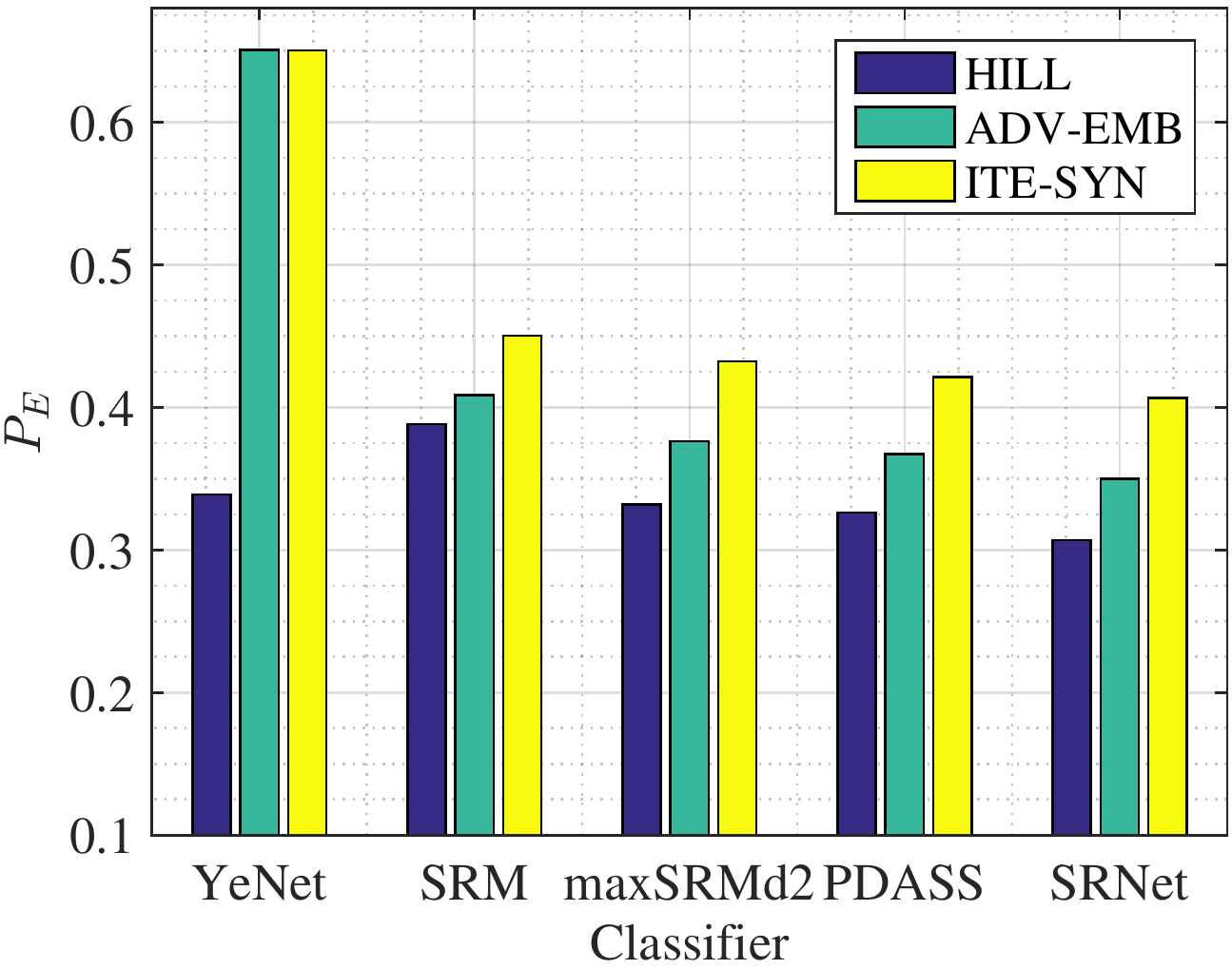}}
      \centerline{(b)}\medskip
    \end{minipage}
    \hfil
    \begin{minipage}[]{0.23\linewidth}
      \centering
      \centerline{\includegraphics[width=1.0\linewidth]{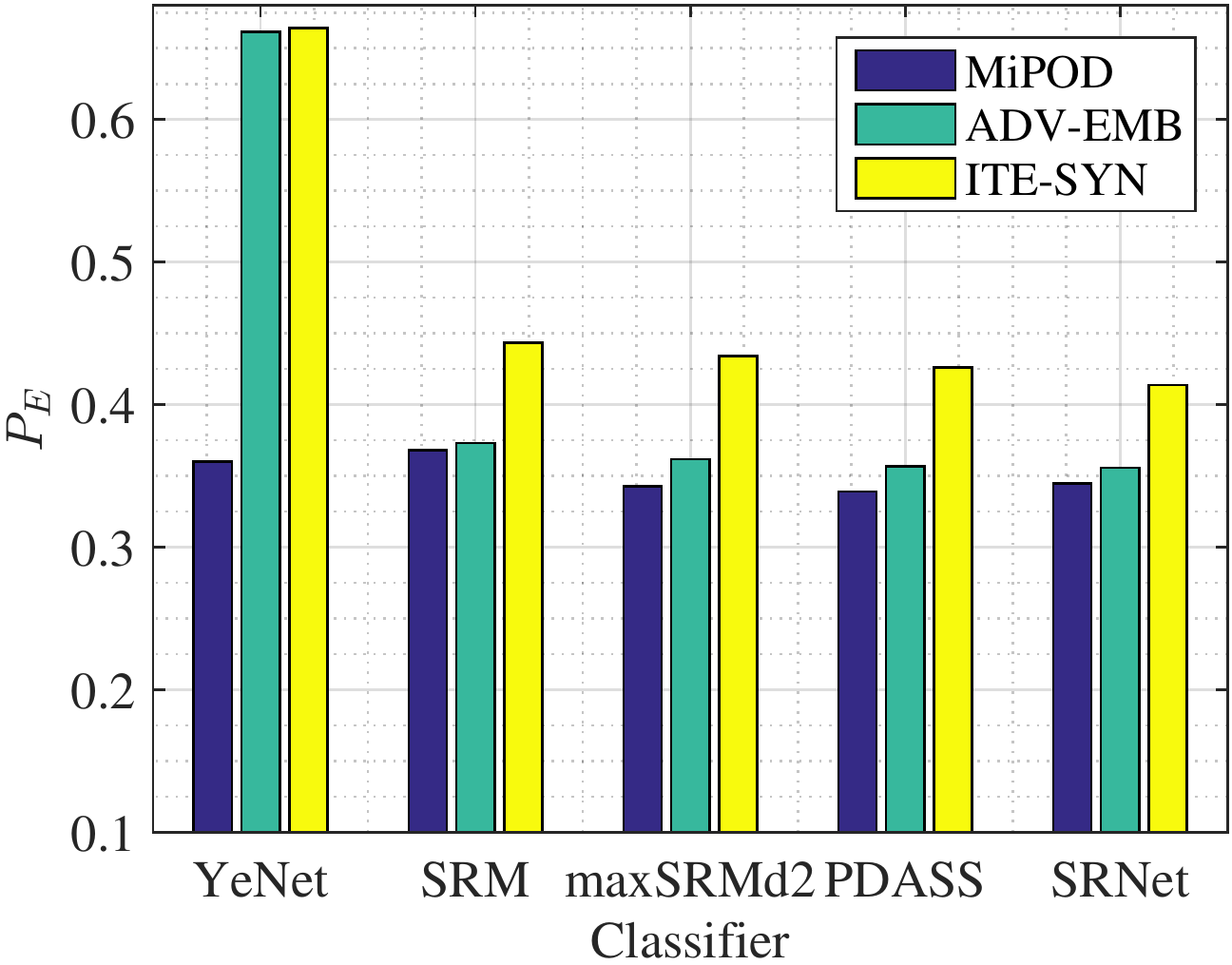}}
      \centerline{(c)}\medskip
    \end{minipage}
    \hfil
    \begin{minipage}[]{0.23\linewidth}
      \centering
      \centerline{\includegraphics[width=1.0\linewidth]{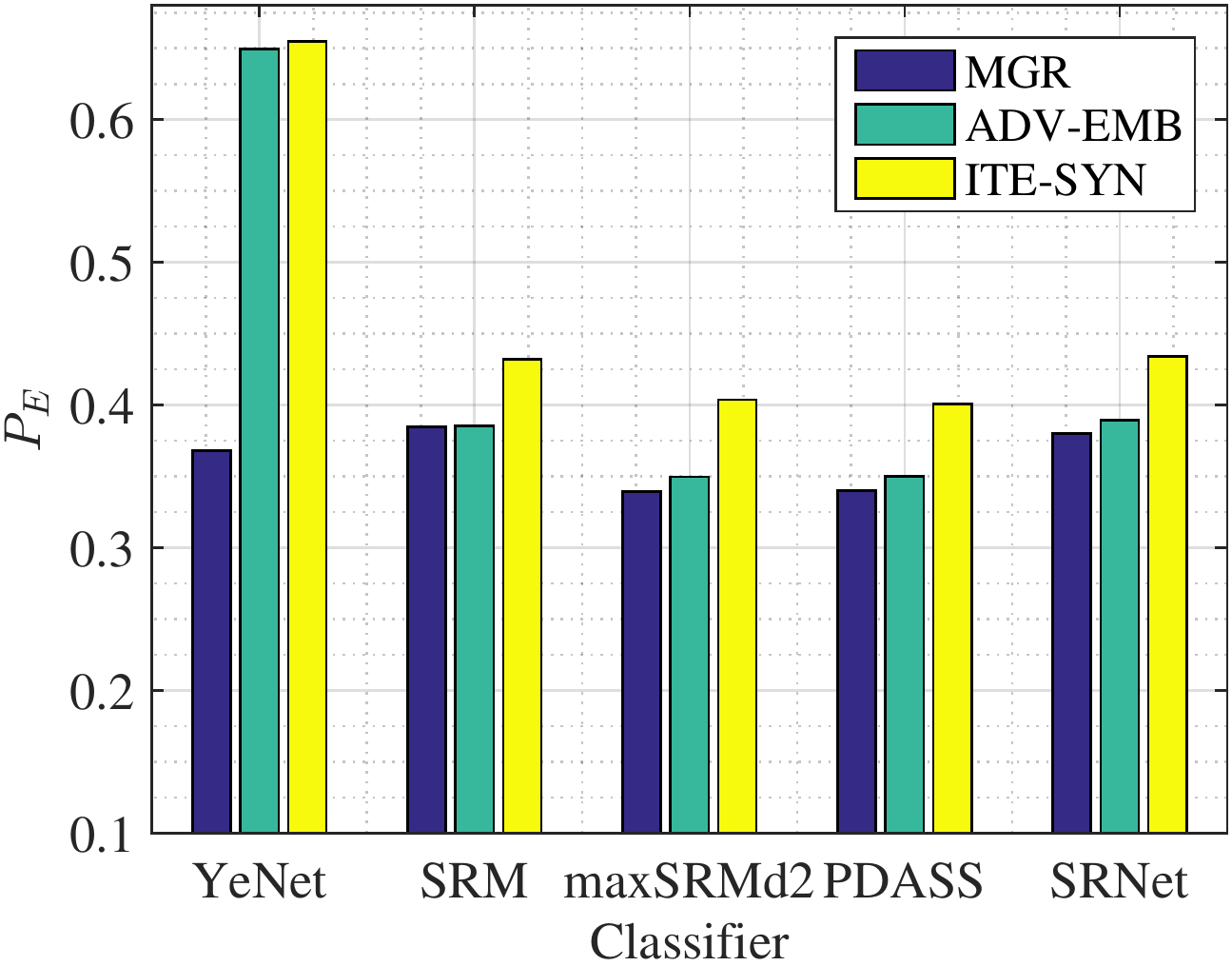}}
      \centerline{(d)}\medskip
    \end{minipage}
    \vfil
    \begin{minipage}[]{0.23\linewidth}
      \centering
      \centerline{\includegraphics[width=1.0\linewidth]{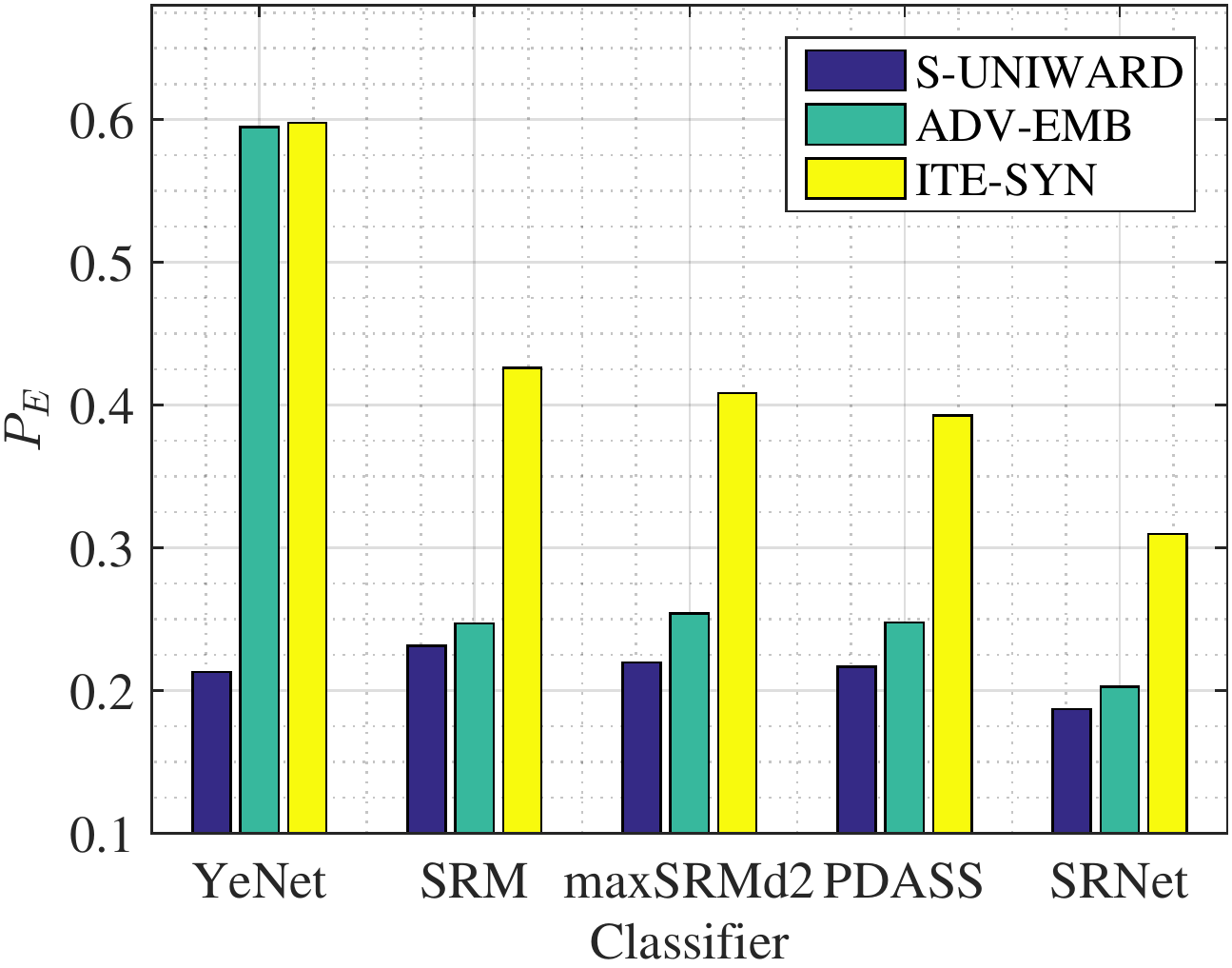}}
      \centerline{(e)}\medskip
    \end{minipage}
    \hfil
    \begin{minipage}[]{0.23\linewidth}
      \centering
      \centerline{\includegraphics[width=1.0\linewidth]{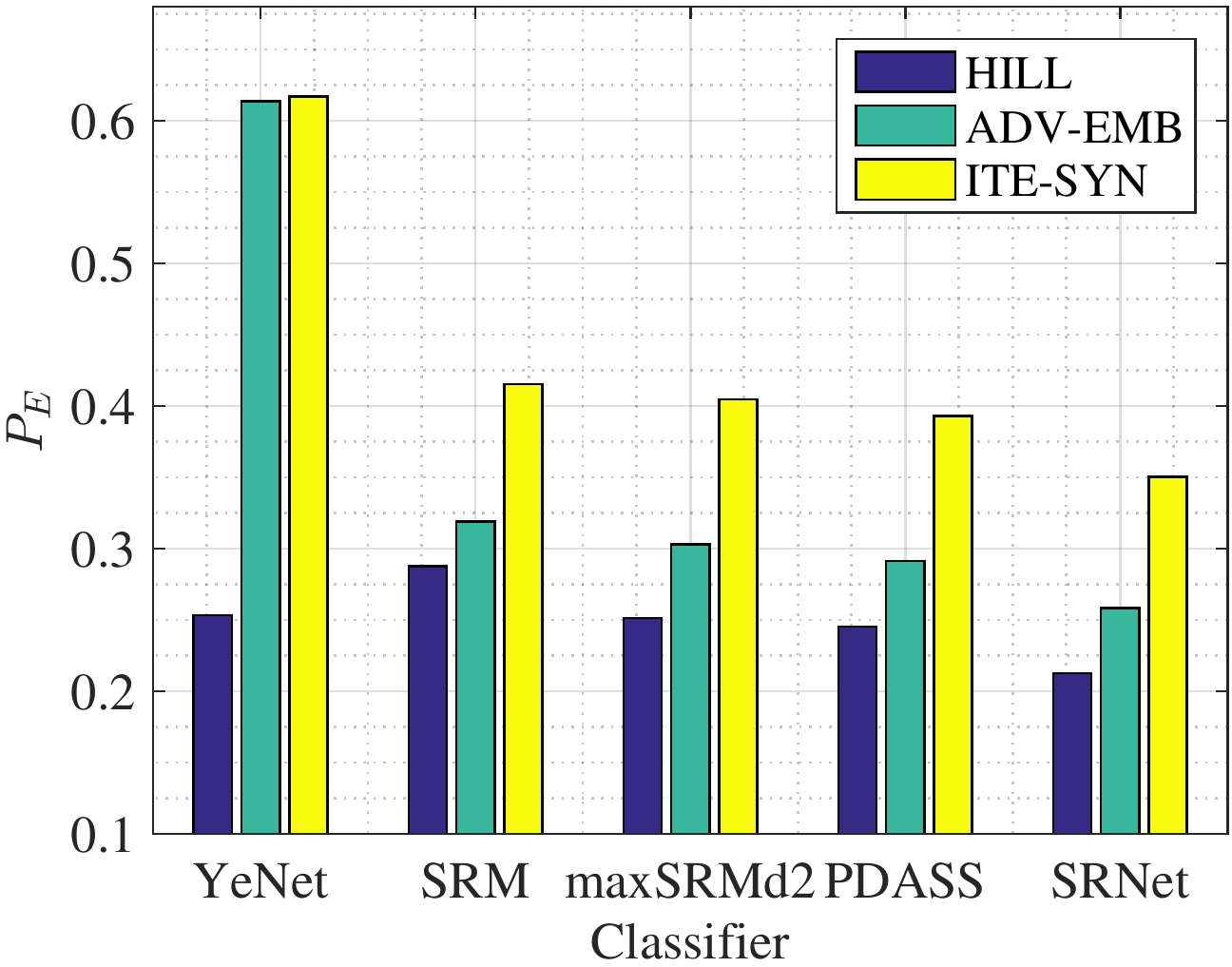}}
      \centerline{(f)}\medskip
    \end{minipage}
    \hfil
    \begin{minipage}[]{0.23\linewidth}
      \centering
      \centerline{\includegraphics[width=1.0\linewidth]{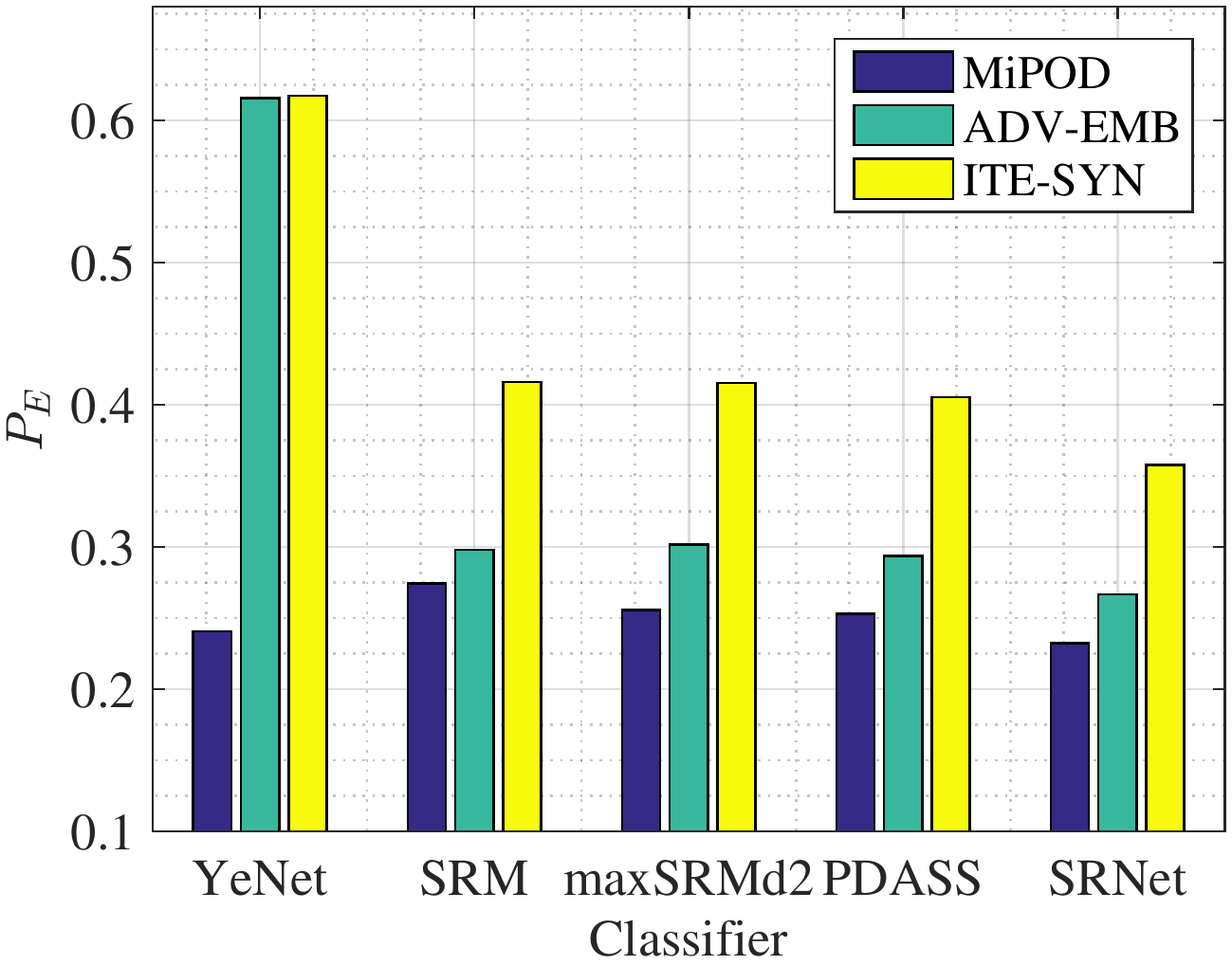}}
      \centerline{(g)}\medskip
    \end{minipage}
    \hfil
    \begin{minipage}[]{0.23\linewidth}
      \centering
      \centerline{\includegraphics[width=1.0\linewidth]{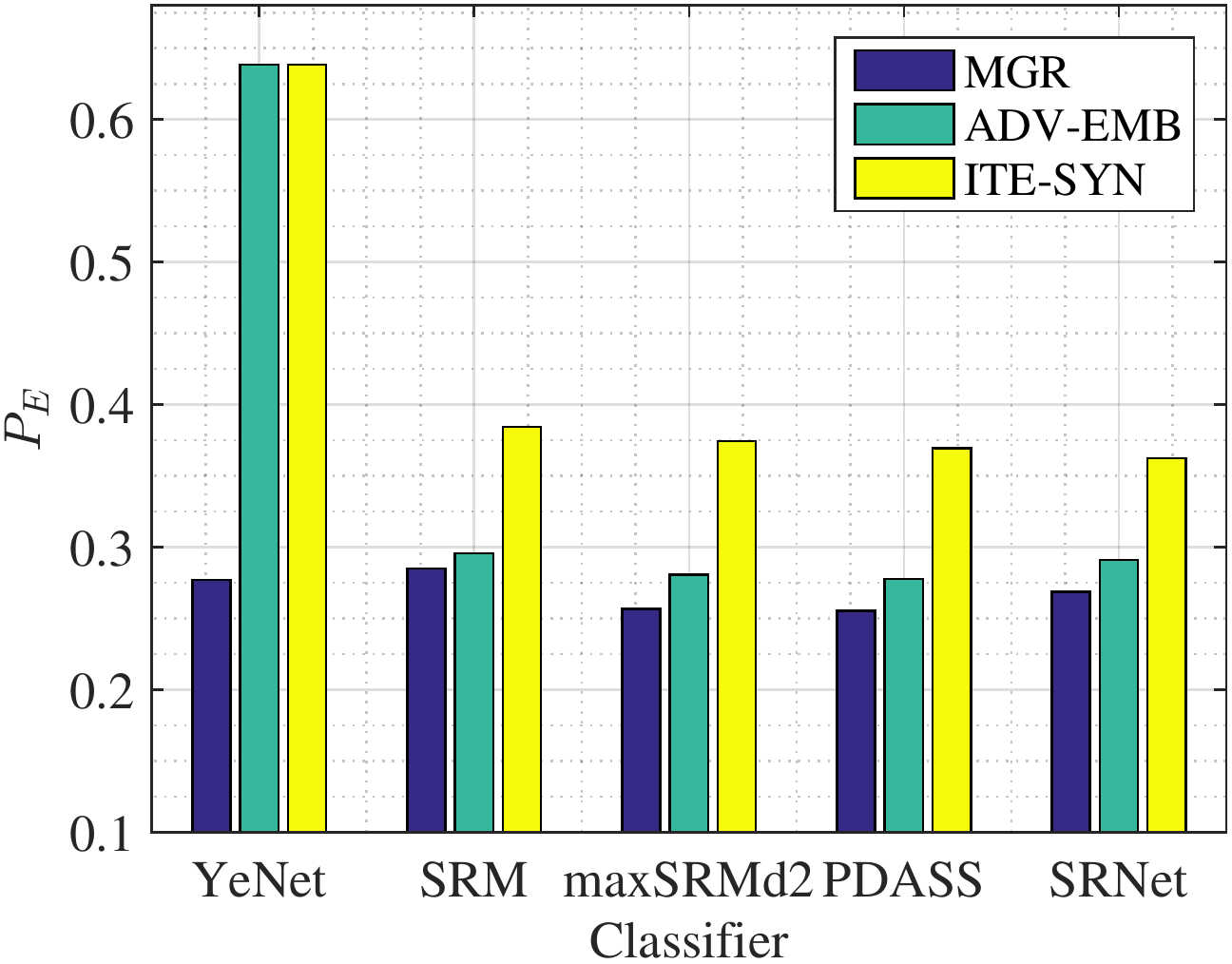}}
      \centerline{(h)}\medskip
    \end{minipage}
  \caption{Performances $P_E$ of deceiving original classifiers with the target YeNet for ALASKA256. For (a)-(d), stego images and adversarial stego images were produced with corresponding steganographic schemes under payload rate 0.2 bpp respectively. And (e)-(h) are under payload rate 0.4 bpp respectively.}
  \label{fig:pe_adv_ye_alaska}
\end{figure*}
Employing given cost functions, we created corresponding stego images.
We used these image sets to train original classifiers, including CNN classifiers and ECs with conventional features. Using XuNet classifiers and YeNet classifiers as target classifiers, we created adversarial stego images by using ADV-EMB method and the proposed ITE-SYN method respectively. 
In \cite{Bernard2019_MM} Bernard et al. proposed Min-Max function to enhance steganographic security by selecting optimal stego images created by using ADV-EMB method to resist on a sequence of classifiers, where there were nine rounds of Min-Max processes. For Min-Max function, we executed nine rounds of processes as well. 
Results of adversarial samples deceiving original detectors for BOSS256 are shown in 
Fig. \ref{fig:pe_adv_xu} and Fig. \ref{fig:pe_adv_ye}, where stego version \textit{BAS} represents baselines, which are tested for original stego images by original classifiers, stego versions \textit{M1-M9} represent each round of Min-Max, and \textit{ADV} and \textit{ITE} represent adversarial stego images produced by ADV-EMB and ITE-SYN. 
As described in \cite{Boroumand2019_SRNet}, SRNet under payload rate 0.2 bpp was trained by adapting curriculum learning from $\mathscr{F}(0.4 bpp)\Rightarrow \mathscr{F}(0.3 bpp)\Rightarrow \mathscr{F}(0.2 bpp)$. The basic classifiers under payload rate 0.4 bpp were trained for $500k$ iterations with decreasing learning rate boundary $450k$ iteration. Remaining classifiers were trained by curriculum learning with $150k$ iterations and decreasing learning rate boundary $100k$ iteration. We selected the best steganalystic performance iterations as classifiers from the last $50k$ iterations of validation results.
Results of deceiving initial detectors for ALASKA256 are shown in Fig. \ref{fig:pe_adv_xu_alaska} and \ref{fig:pe_adv_ye_alaska} as well. For saving experiments' resources, we did not execute Min-Max function for images database ALASKA256.
There are observations as following.
\begin{itemize}
\item $P_E$ of ITE-SYN deceiving target CNN classifiers, including both XuNet and YeNet, are significantly improved from baselines, which indicate that the proposed method can deceive target classifiers successfully.
\item For deceiving target CNN classifiers, ITE-SYN versions perform inferior to corresponding ADV-EMB but superior to MinMax versions. It indicates that success rates of creating adversarial stego images by using ITE-SYN are lower than corresponding ADV-EMB. For ITE-SYN, we only apply adversarial perturbations on a sub-image for each image. It is possible that apply adversarial perturbations on more sub-images to improve success rates. We will investigate it in the future research. For Min-Max versions, stego images are most difficult onces selected from original stego images and previous versions adversarial stego images. The more original stego images are selected as most difficult images, $P_E$ of Min-Max versions deceiving the original target CNN classifiers are inferior.
\item For resisting on detecting by other original classifiers, ITE-SYN versions perform superior to corresponding ADV-EMB and MinMax versions, including both CNN classifiers SRNet and conventional ensemble classifiers SRM, maxSRMd2 and PDASS. 
\end{itemize}

\begin{figure*}[tbp]
  \centering
    \begin{minipage}[]{0.23\linewidth}
      \centering
      \centerline{\includegraphics[width=1.0\linewidth]{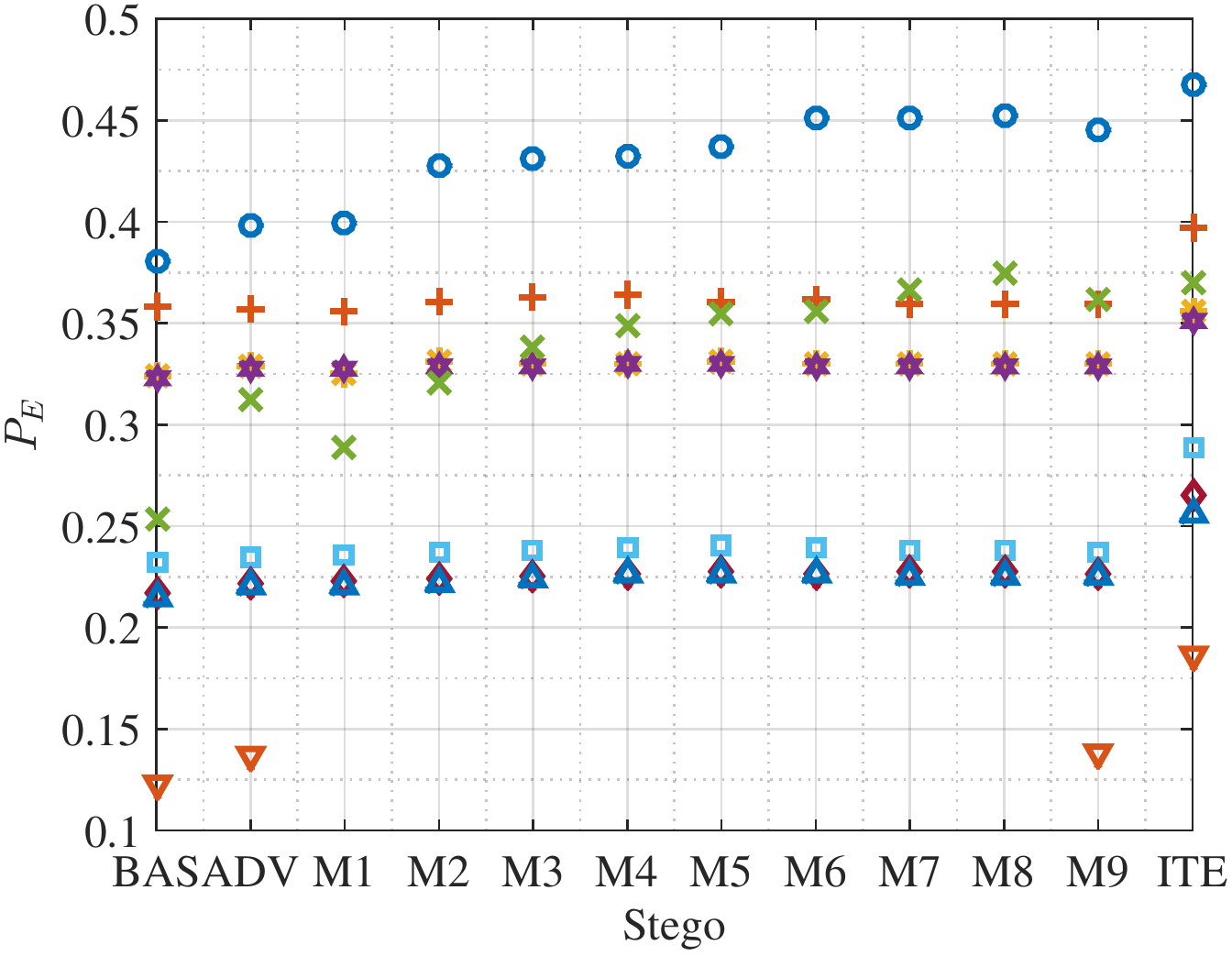}}
      \centerline{(a) S-UNIWARD}\medskip
    \end{minipage}
    \hfil
    \begin{minipage}[]{0.23\linewidth}
      \centering
      \centerline{\includegraphics[width=1.0\linewidth]{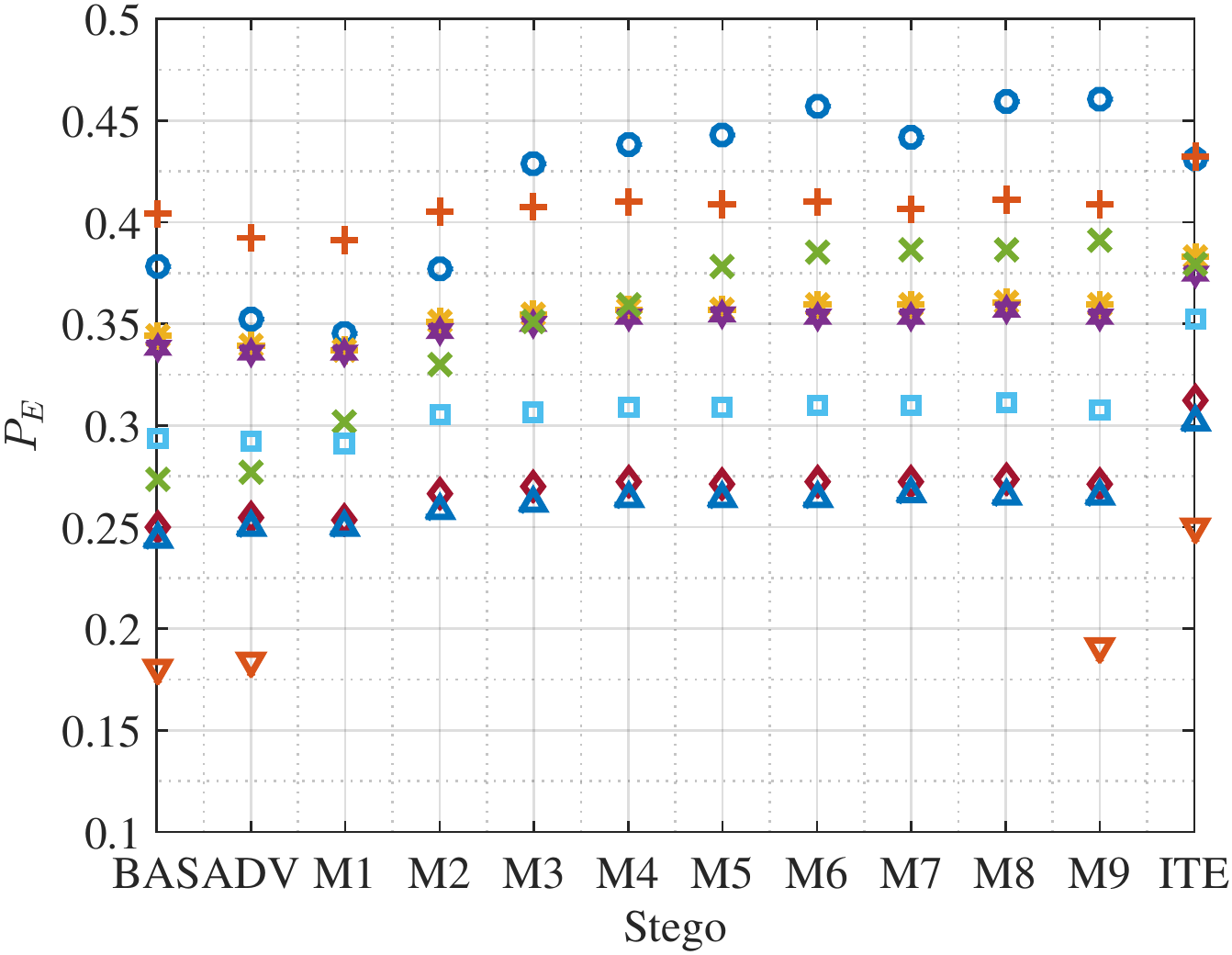}}
      \centerline{(b) HILL}\medskip
    \end{minipage}
    \hfil
    \begin{minipage}[]{0.23\linewidth}
      \centering
      \centerline{\includegraphics[width=1.0\linewidth]{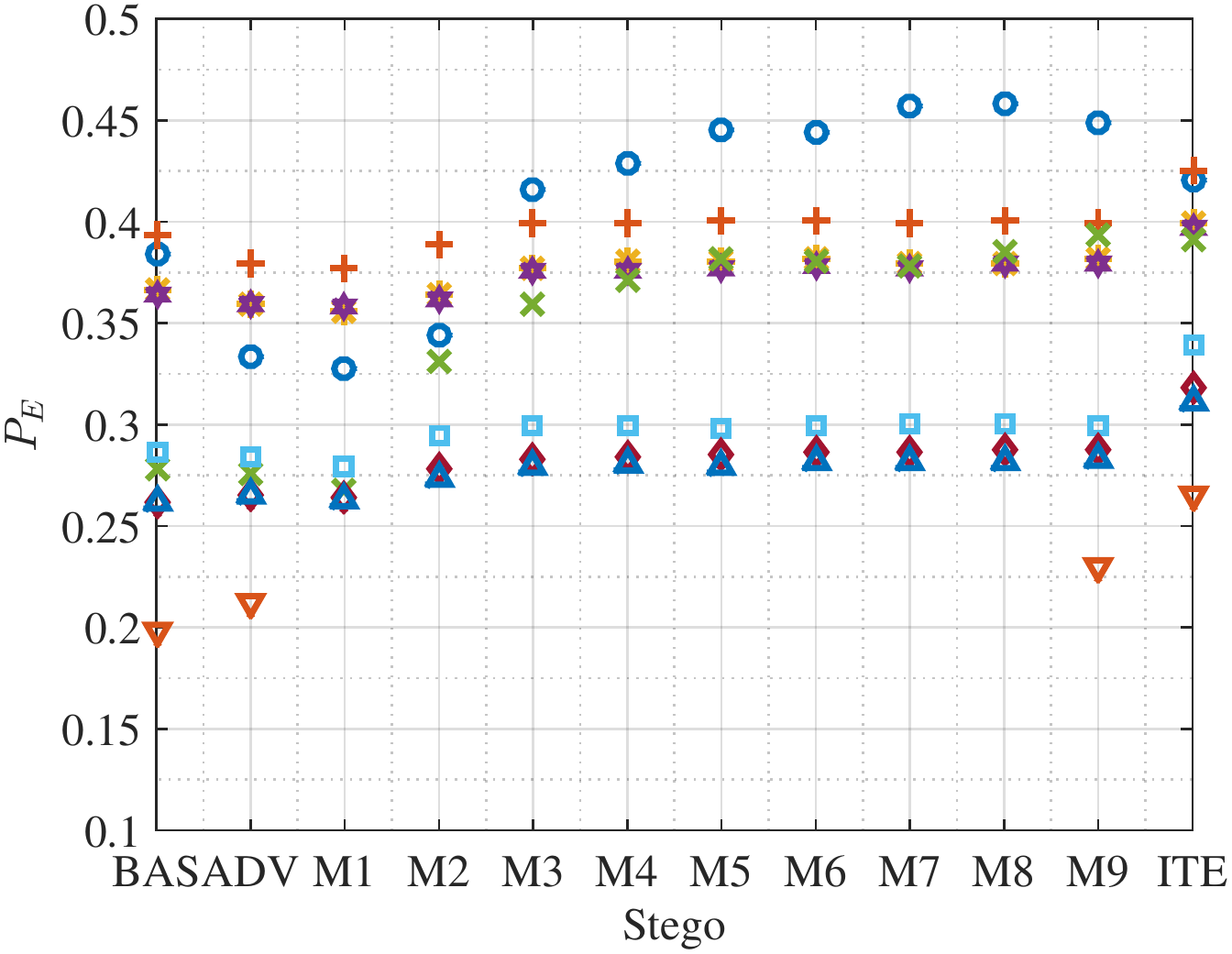}}
      \centerline{(c) MiPOD}\medskip
    \end{minipage}
    \hfil
    \begin{minipage}[]{0.23\linewidth}
      \centering
      \centerline{\includegraphics[width=1.0\linewidth]{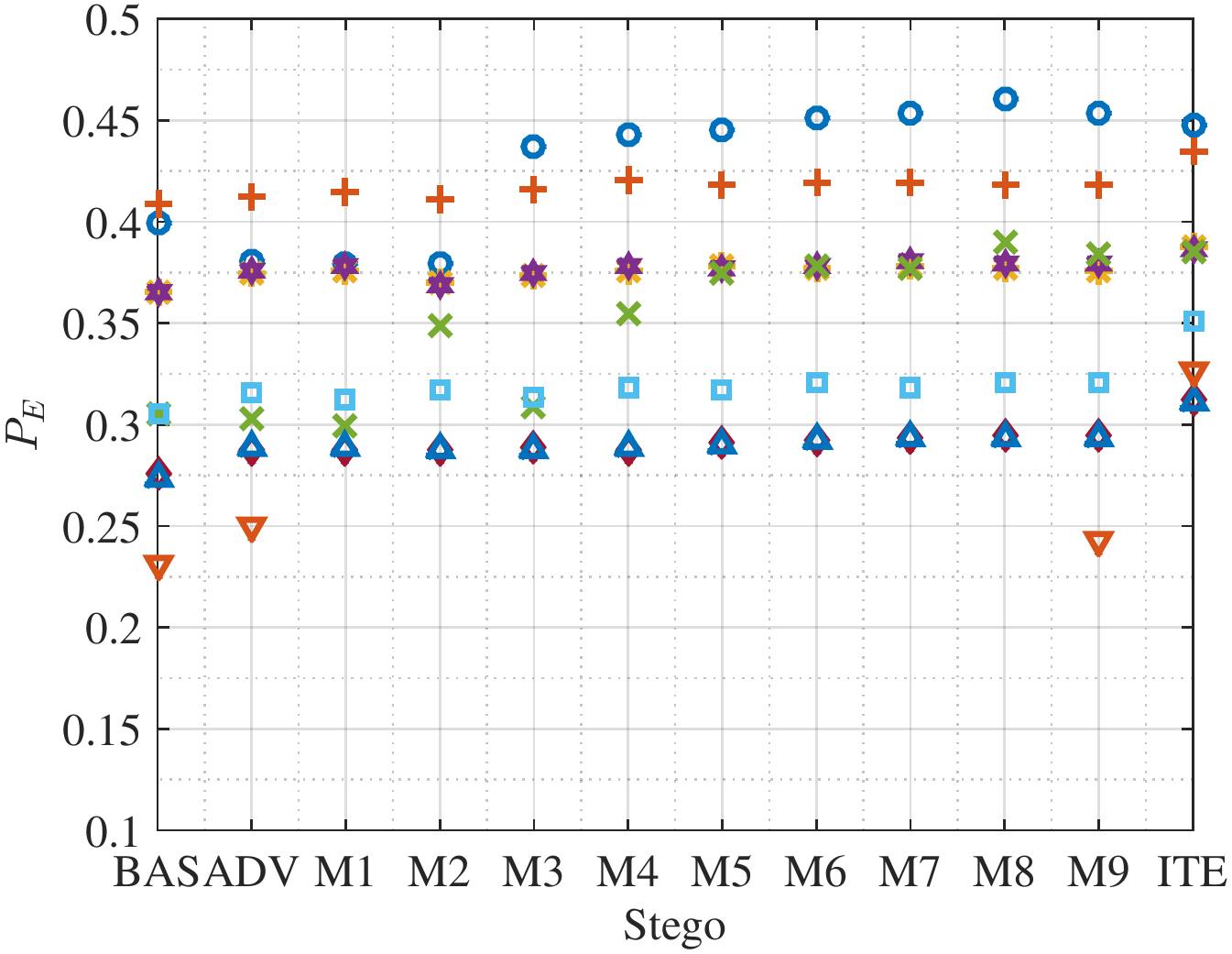}}
      \centerline{(d) MGR}\medskip
    \end{minipage}
    \vfil
    \begin{minipage}[]{0.6\linewidth}
      \centering
      \centerline{\includegraphics[width=1.0\linewidth]{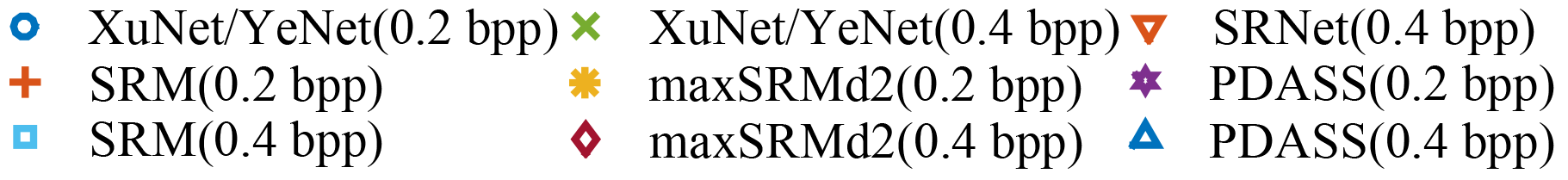}}
      \centerline{}\medskip
    \end{minipage}
  \caption{Performances $P_E$ of resisting on adversarial trained classifiers with the target XuNet for BOSS256. Decimal fractions in parenthesis indicate payload rates 0.2 bpp and 0.4 bpp respectively. For stego versions, 0 represents the baseline, 1-9 represent rounds of Min-Max function, ADV and ITE represent ADV-EMB and ITE-SYN respectively.}
  \label{fig:pe_retrain_xu}
\end{figure*}

\begin{figure*}[tbp]
  \centering
    \begin{minipage}[]{0.23\linewidth}
      \centering
      \centerline{\includegraphics[width=1.0\linewidth]{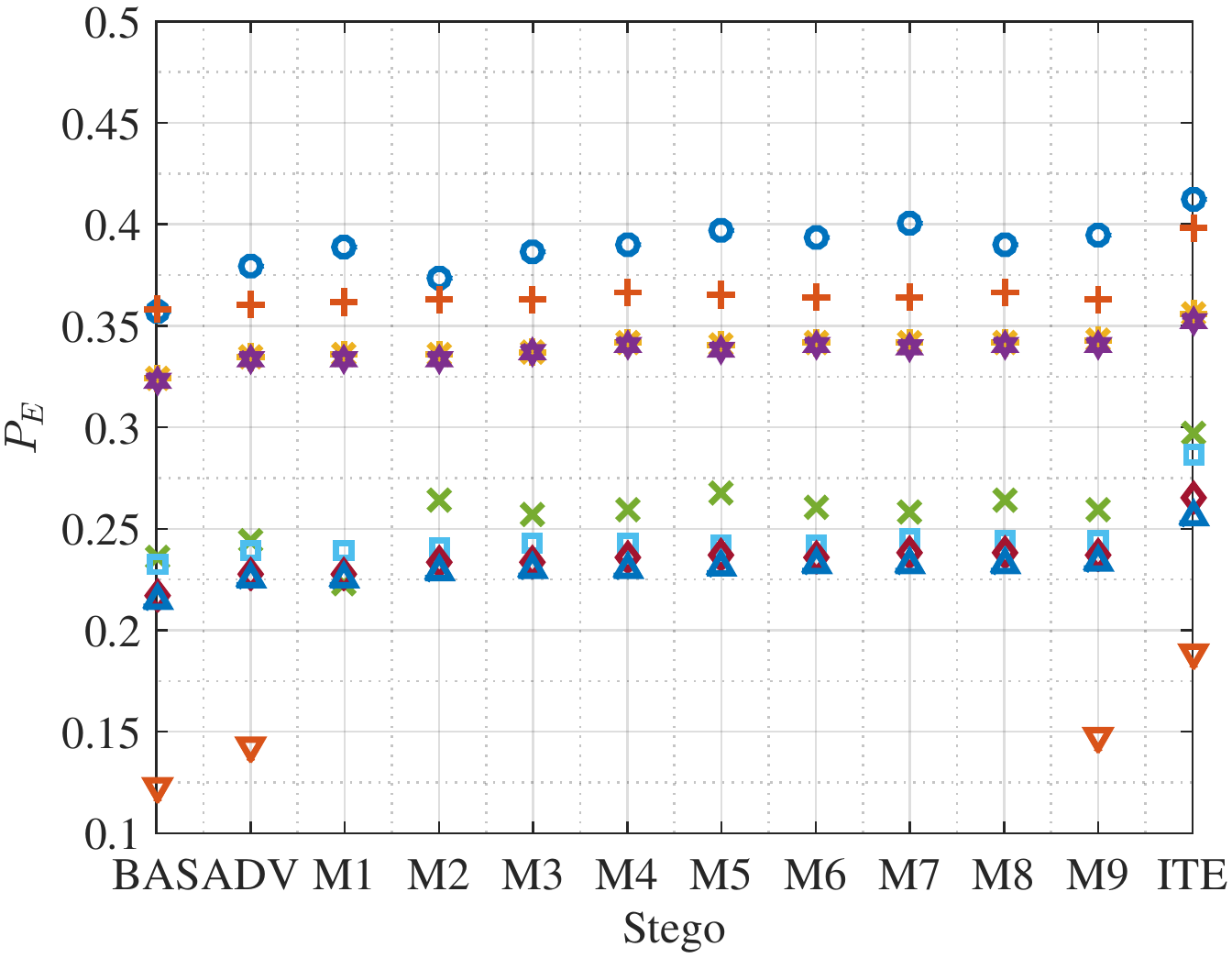}}
      \centerline{(a) S-UNIWARD}\medskip
    \end{minipage}
    \hfil
    \begin{minipage}[]{0.23\linewidth}
      \centering
      \centerline{\includegraphics[width=1.0\linewidth]{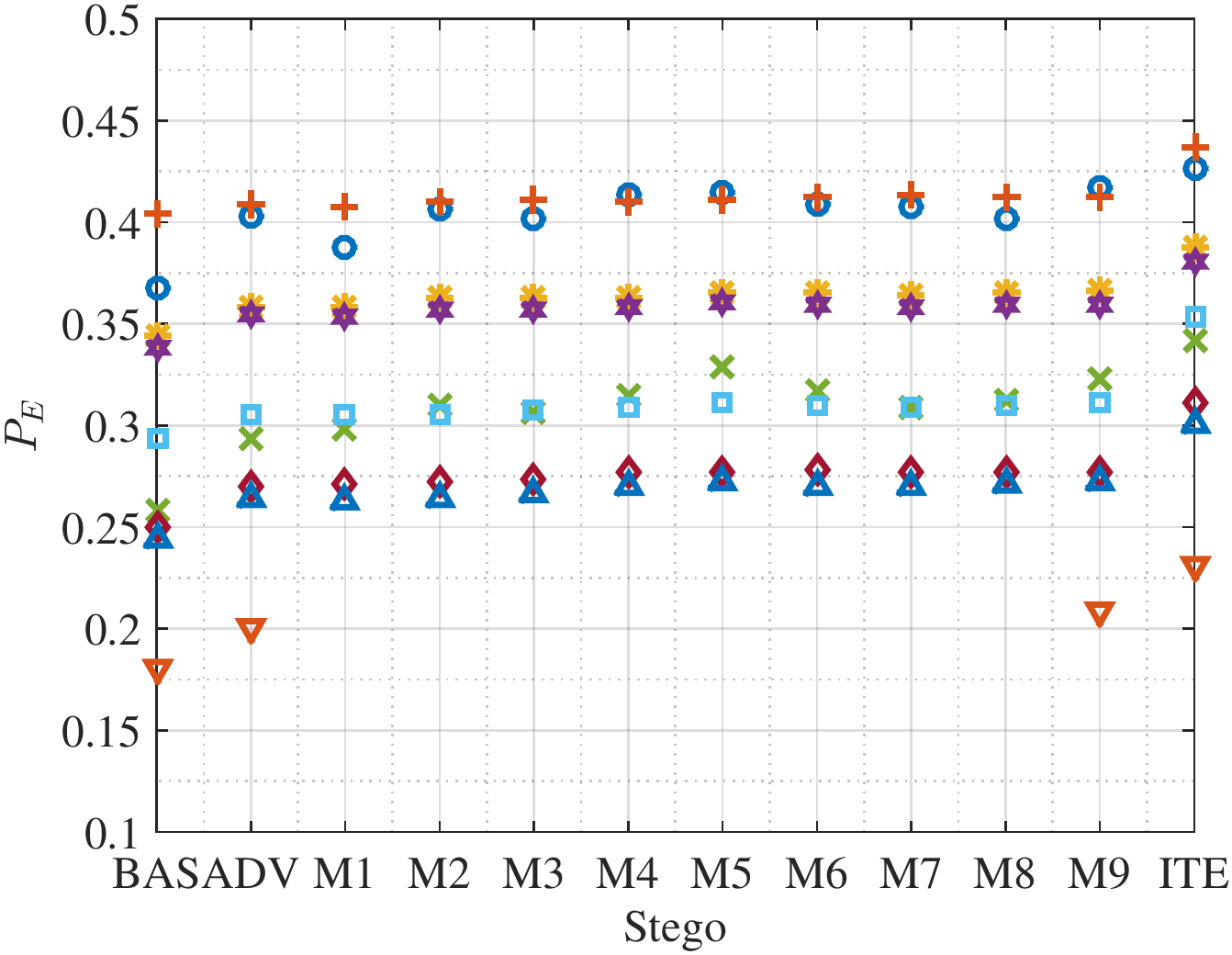}}
      \centerline{(b) HILL}\medskip
    \end{minipage}
    \hfil
    \begin{minipage}[]{0.23\linewidth}
      \centering
      \centerline{\includegraphics[width=1.0\linewidth]{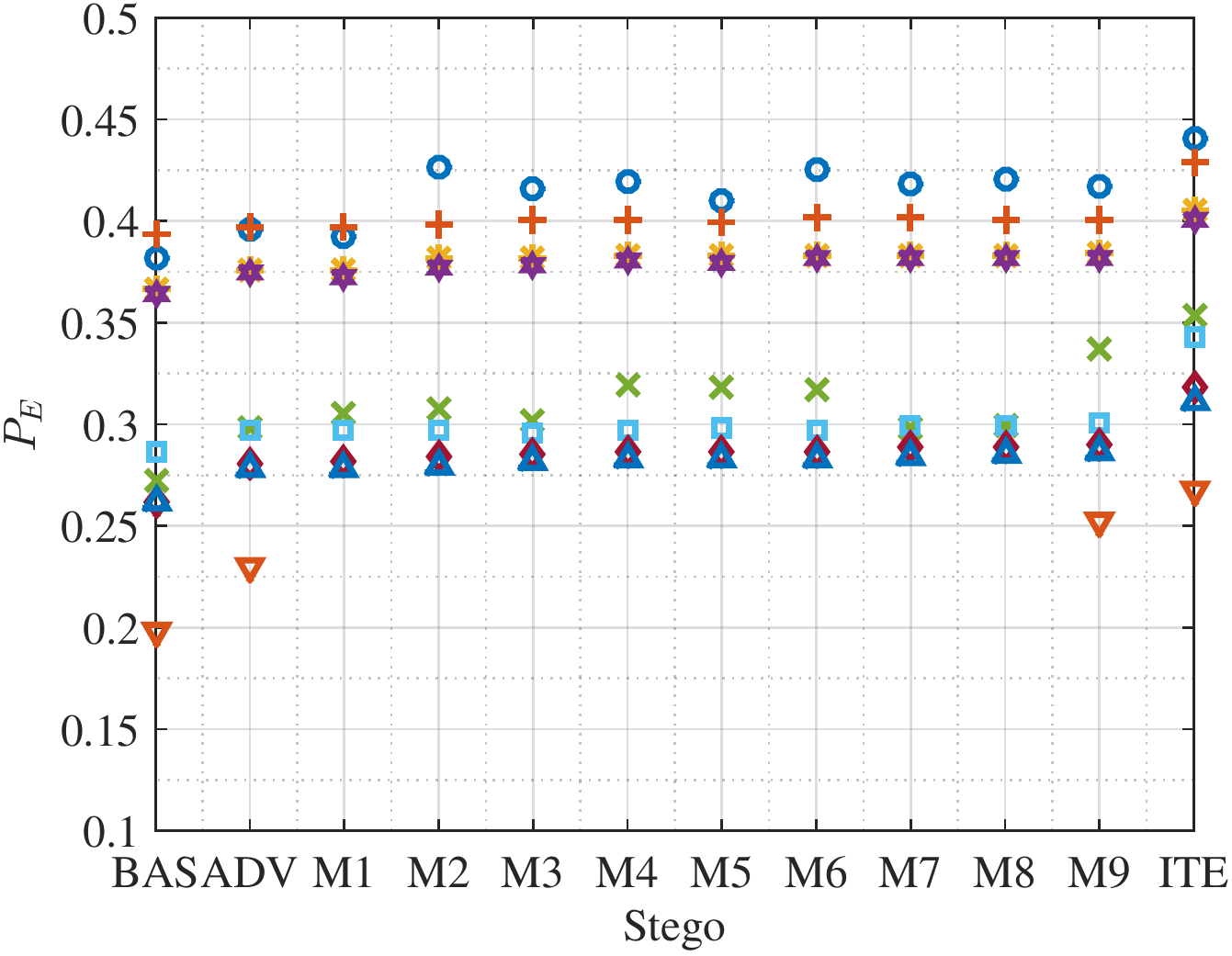}}
      \centerline{(c) MiPOD}\medskip
    \end{minipage}
    \hfil
    \begin{minipage}[]{0.23\linewidth}
      \centering
      \centerline{\includegraphics[width=1.0\linewidth]{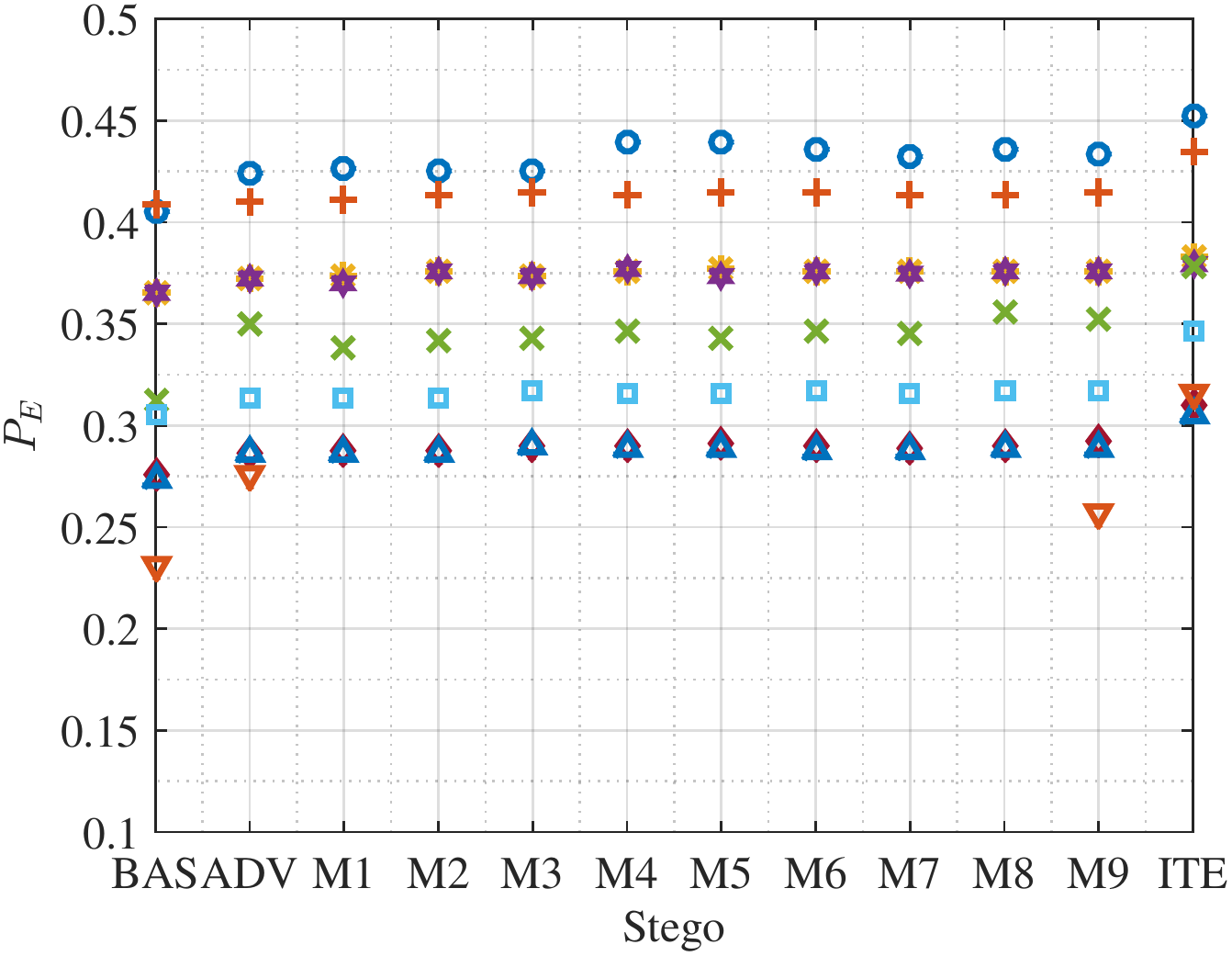}}
      \centerline{(d) MGR}\medskip
    \end{minipage}
    \vfil
    \begin{minipage}[]{0.60\linewidth}
      \centering
      \centerline{\includegraphics[width=1.0\linewidth]{fig_legend_trn_n}}
      \centerline{}\medskip
    \end{minipage}
  \caption{Performances $P_E$ of resisting on adversarial trained classifiers with the target YeNet for BOSS256. Decimal fractions in parenthesis indicate payload rates 0.2 bpp and 0.4 bpp respectively. For stego versions, 0 represents the baseline, 1-9 represent rounds of Min-Max function, ADV and ITE represent ADV-EMB and ITE-SYN respectively.}
  \label{fig:pe_retrain_ye}
\end{figure*}

\begin{figure*}[tbp]
  \centering
    \begin{minipage}[]{0.23\linewidth}
      \centering
      \centerline{\includegraphics[width=1.0\linewidth]{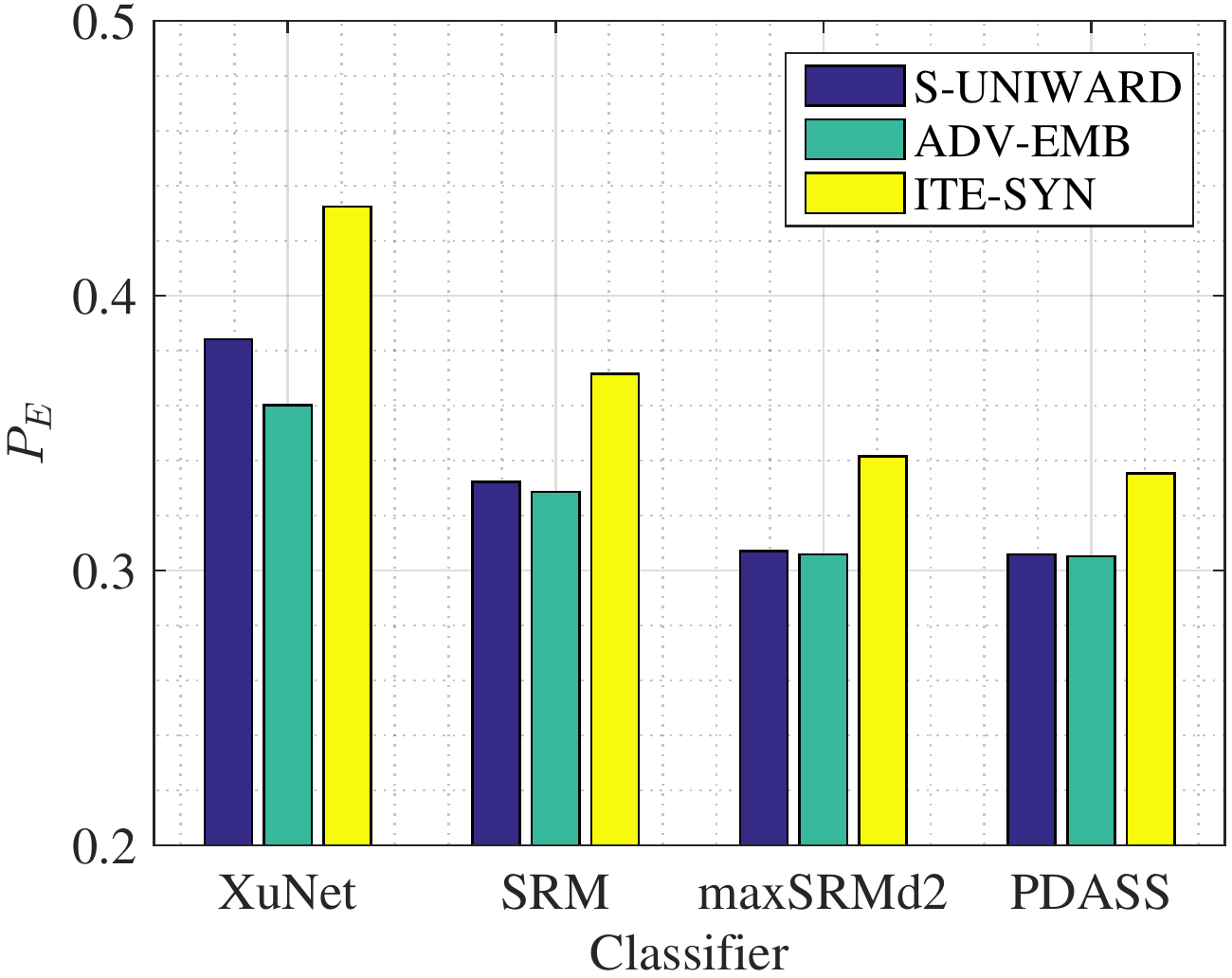}}
      \centerline{(a)}\medskip
    \end{minipage}
    \hfil
    \begin{minipage}[]{0.23\linewidth}
      \centering
      \centerline{\includegraphics[width=1.0\linewidth]{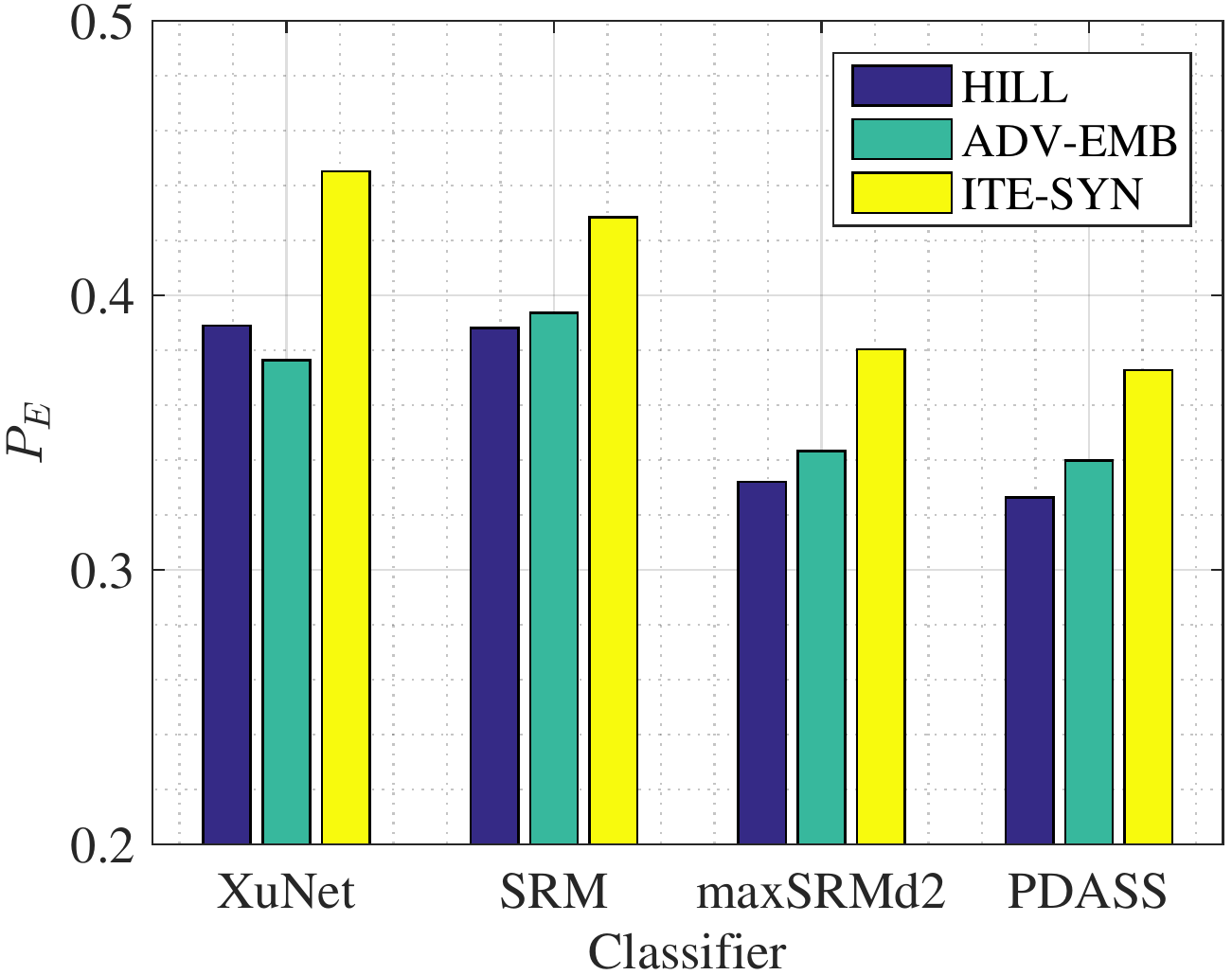}}
      \centerline{(b)}\medskip
    \end{minipage}
    \hfil
    \begin{minipage}[]{0.23\linewidth}
      \centering
      \centerline{\includegraphics[width=1.0\linewidth]{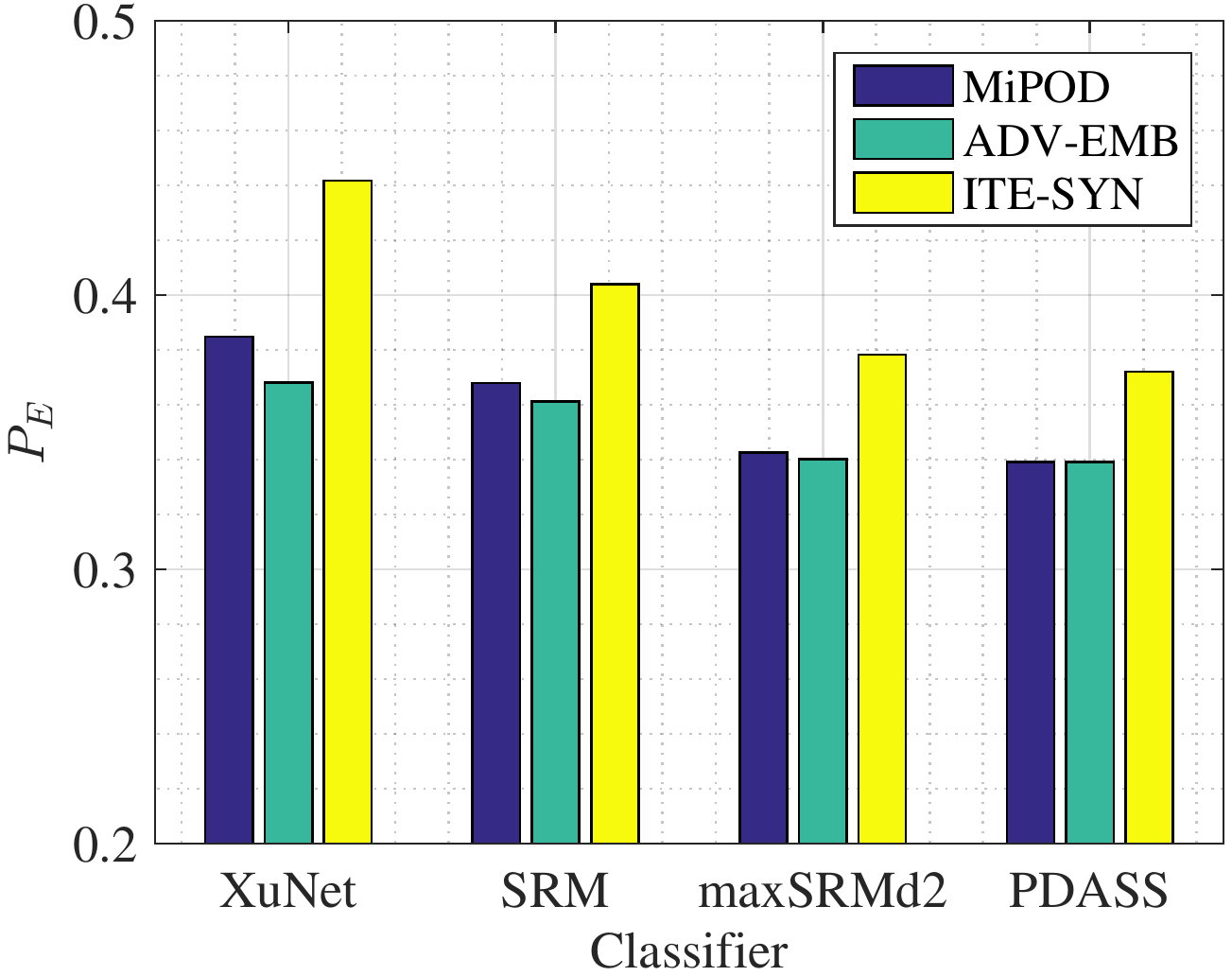}}
      \centerline{(c)}\medskip
    \end{minipage}
    \hfil
    \begin{minipage}[]{0.23\linewidth}
      \centering
      \centerline{\includegraphics[width=1.0\linewidth]{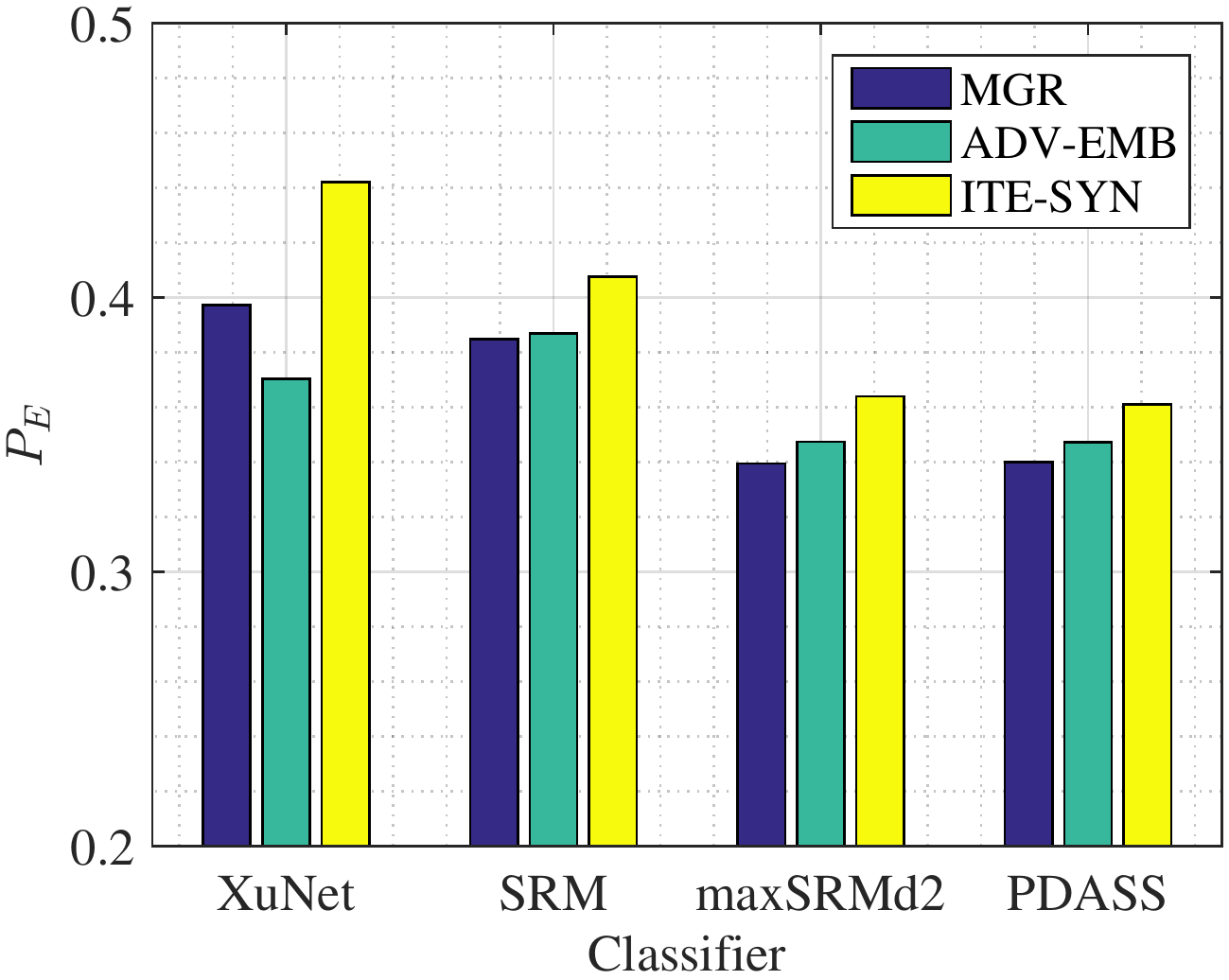}}
      \centerline{(d)}\medskip
    \end{minipage}
    \vfil
    \begin{minipage}[]{0.23\linewidth}
      \centering
      \centerline{\includegraphics[width=1.0\linewidth]{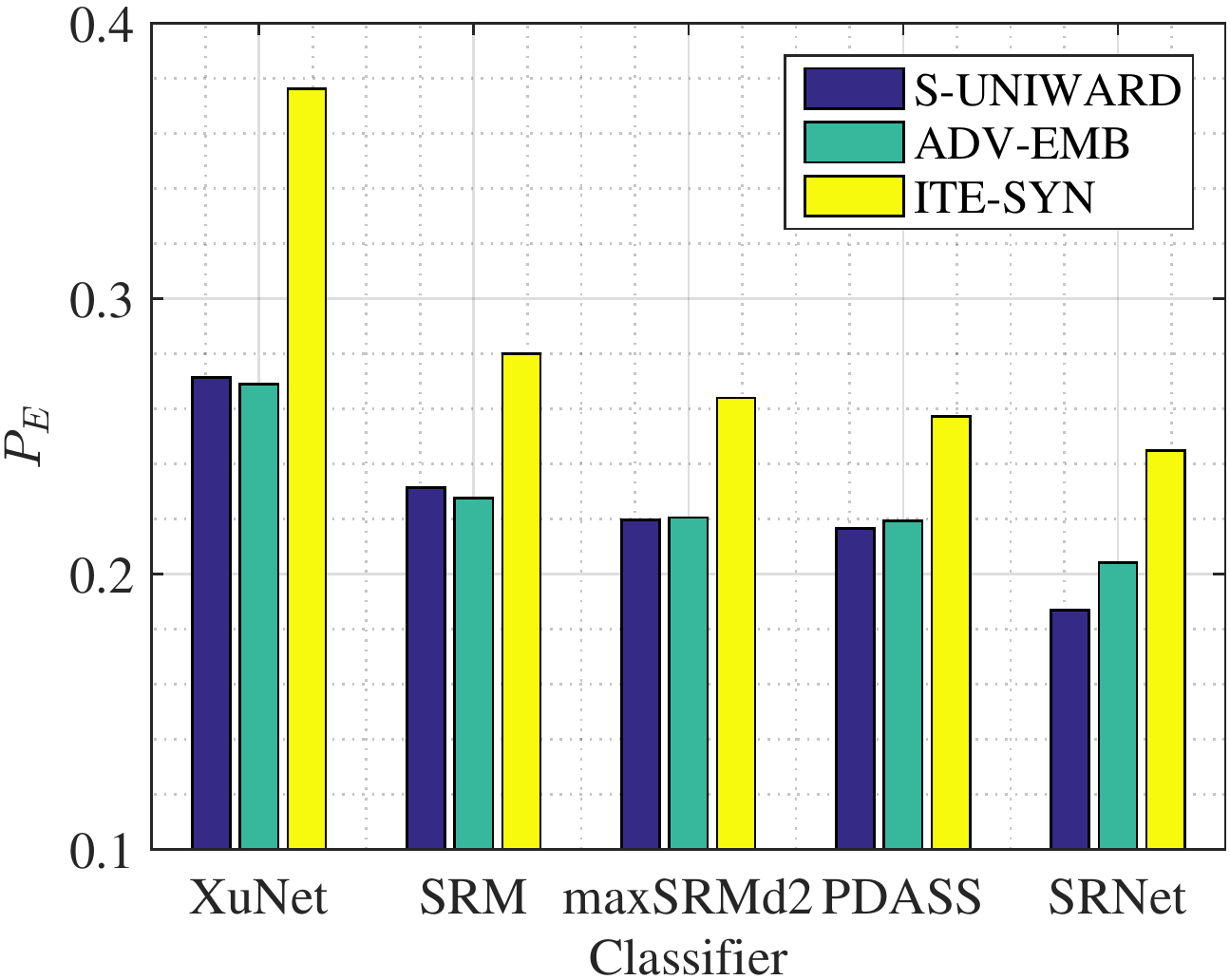}}
      \centerline{(e)}\medskip
    \end{minipage}
    \hfil
    \begin{minipage}[]{0.23\linewidth}
      \centering
      \centerline{\includegraphics[width=1.0\linewidth]{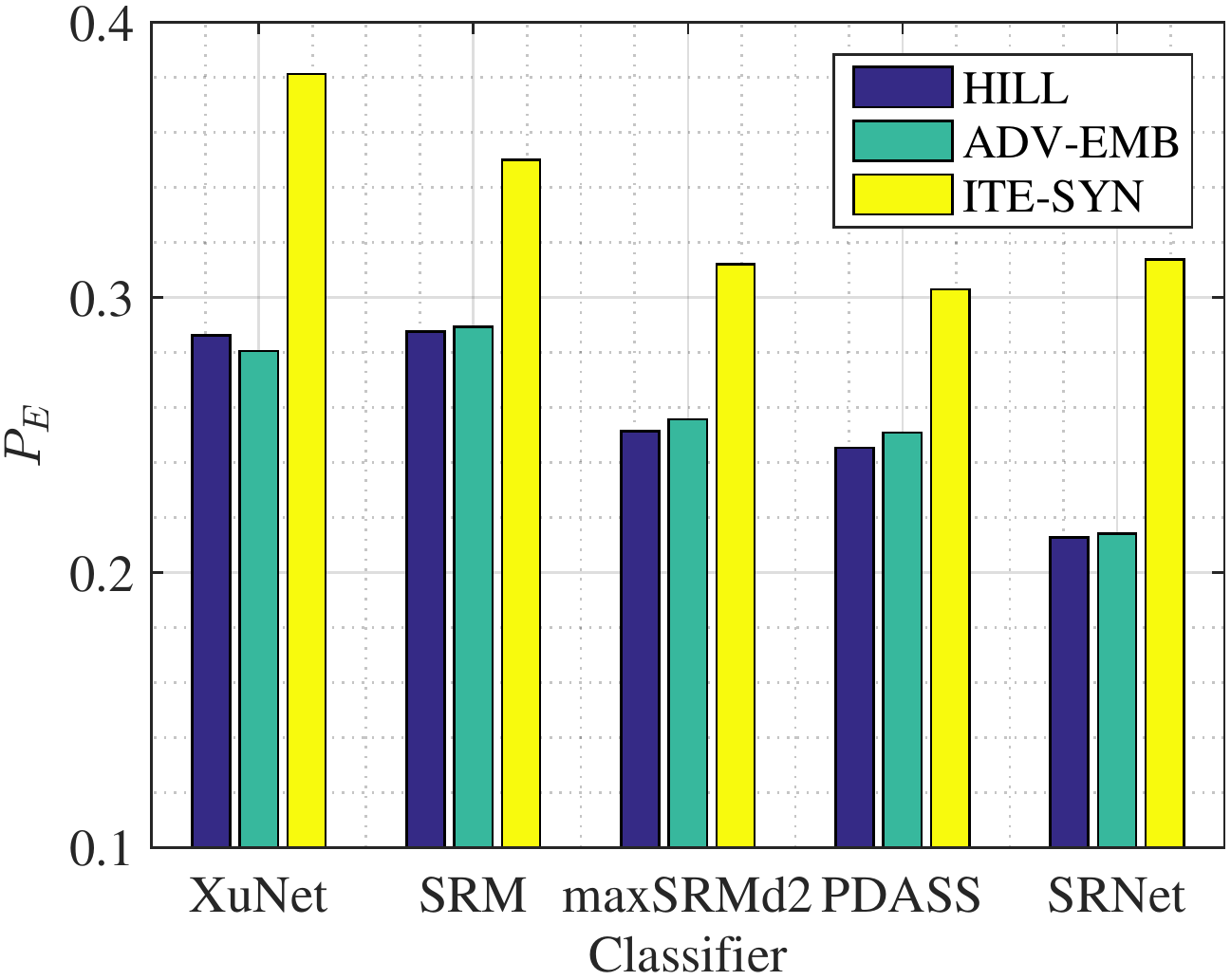}}
      \centerline{(f)}\medskip
    \end{minipage}
    \hfil
    \begin{minipage}[]{0.23\linewidth}
      \centering
      \centerline{\includegraphics[width=1.0\linewidth]{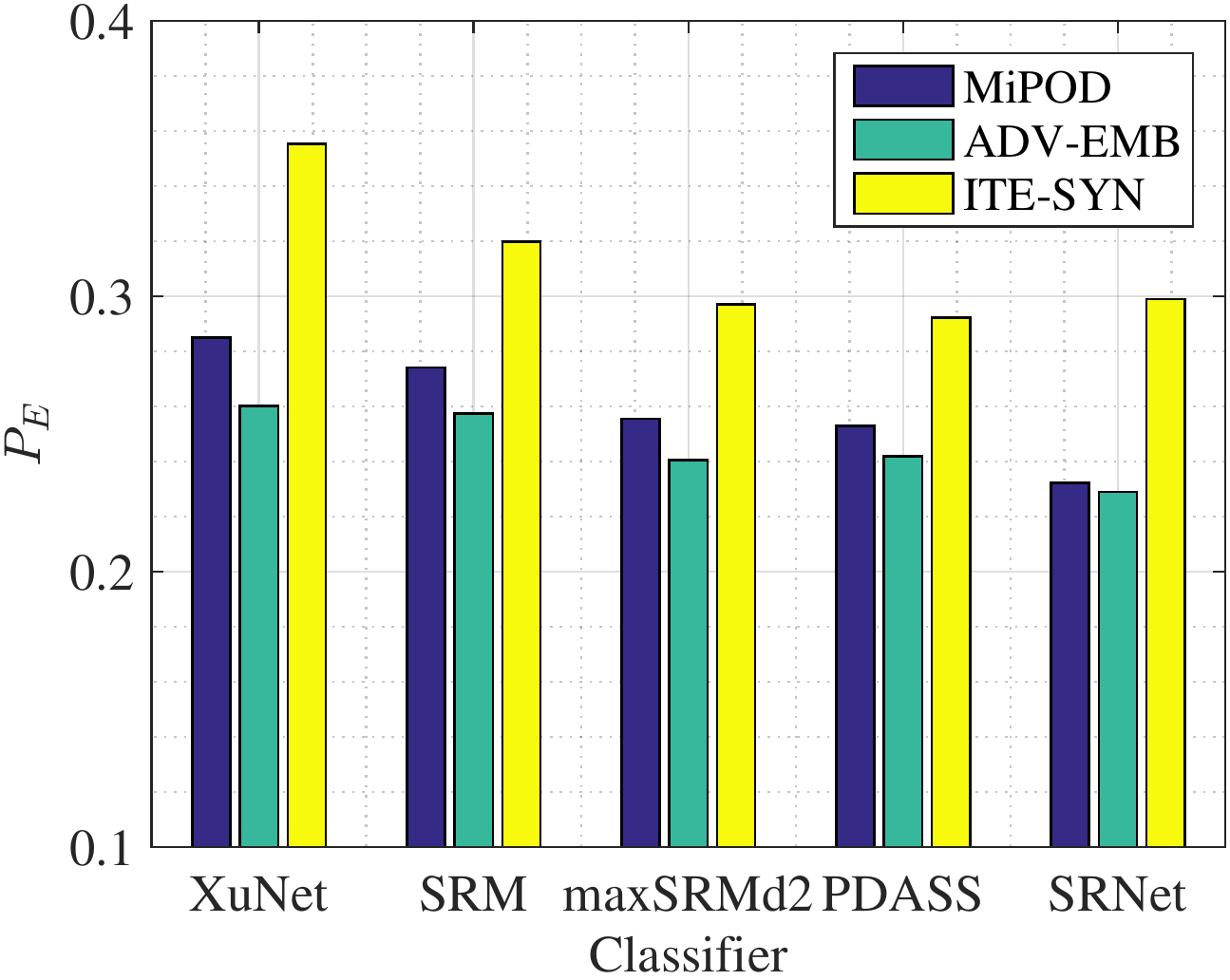}}
      \centerline{(g)}\medskip
    \end{minipage}
    \hfil
    \begin{minipage}[]{0.23\linewidth}
      \centering
      \centerline{\includegraphics[width=1.0\linewidth]{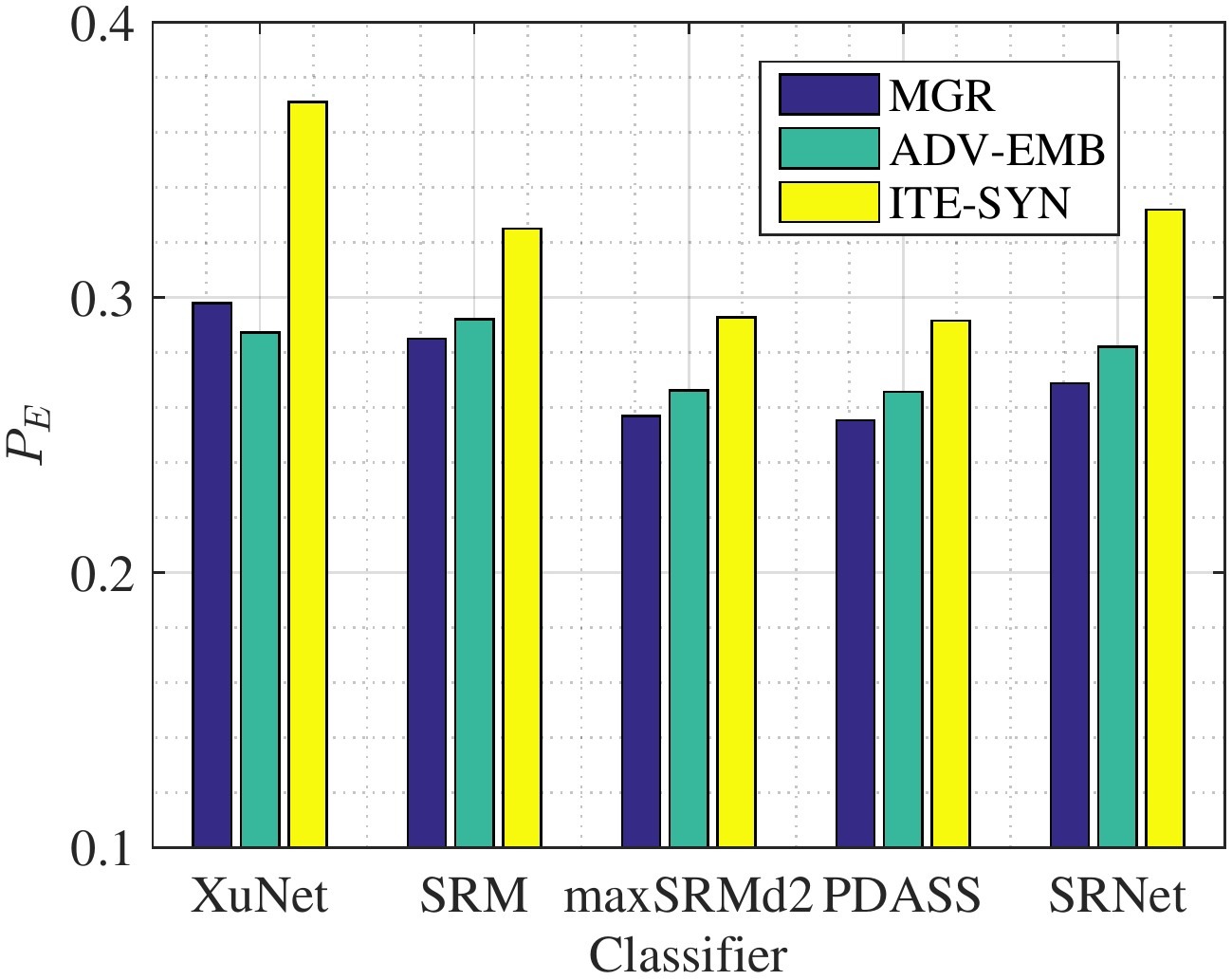}}
      \centerline{(h)}\medskip
    \end{minipage}
  \caption{Performances $P_E$ of resisting on adversarial trained classifiers with the target XuNet for ALASKA256. For (a)-(d), stego images and adversarial stego images were produced with corresponding steganographic schemes under payload rate 0.2 bpp respectively. And (e)-(h) are under payload rate 0.4 bpp respectively.}
  \label{fig:pe_xu_alaska}
\end{figure*}

\begin{figure*}[tbp]
  \centering
    \begin{minipage}[]{0.23\linewidth}
      \centering
      \centerline{\includegraphics[width=1.0\linewidth]{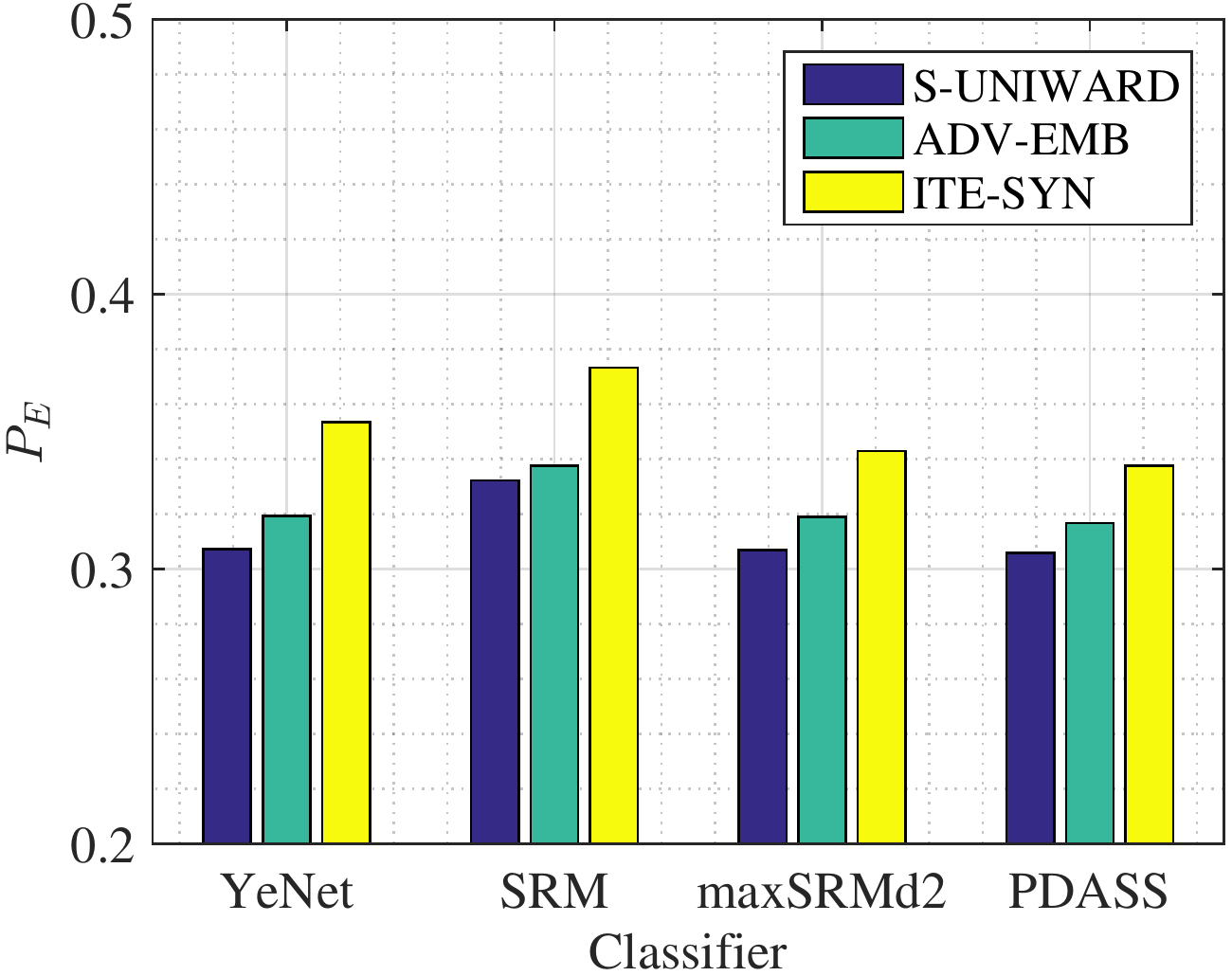}}
      \centerline{(a)}\medskip
    \end{minipage}
    \hfil
    \begin{minipage}[]{0.23\linewidth}
      \centering
      \centerline{\includegraphics[width=1.0\linewidth]{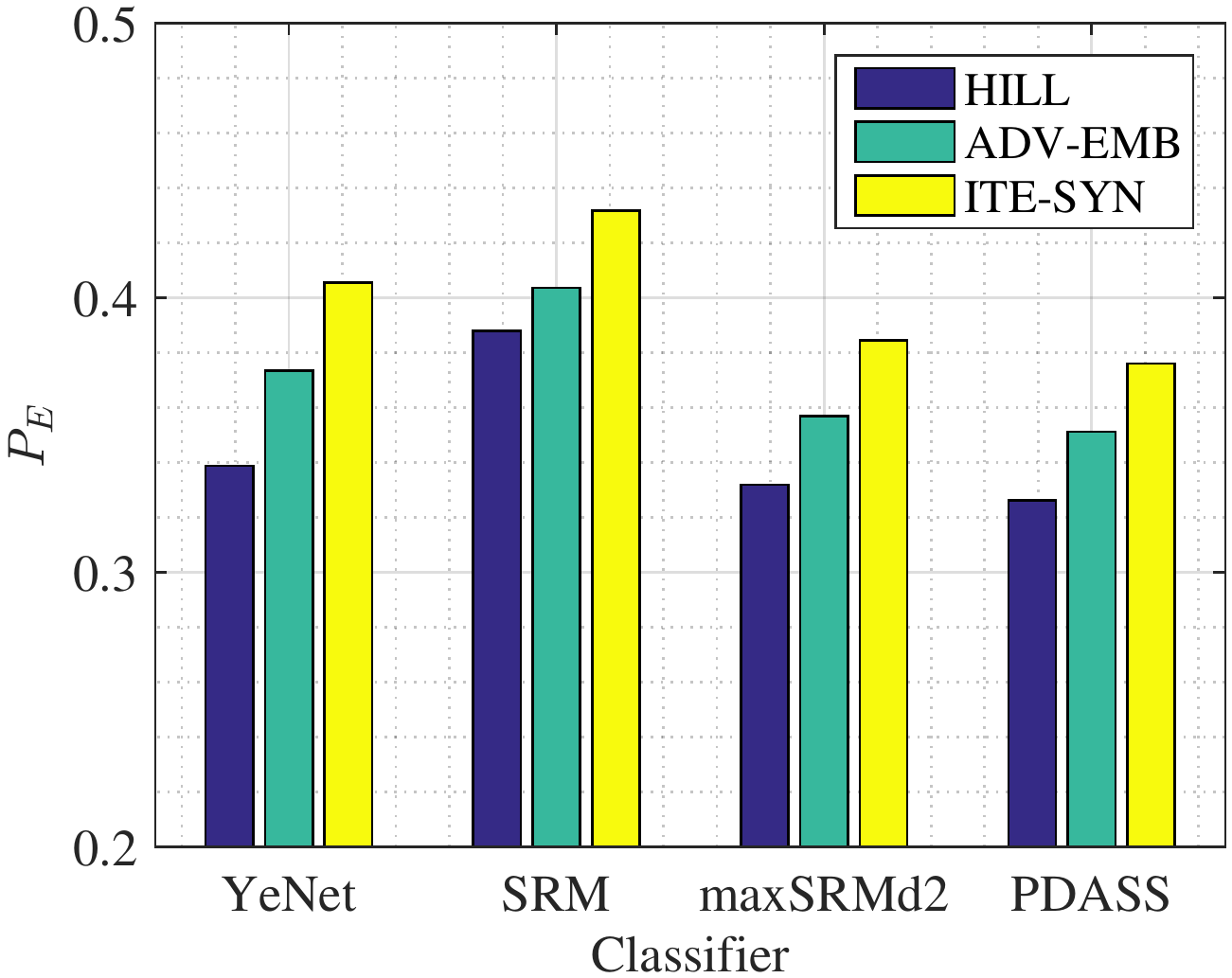}}
      \centerline{(b)}\medskip
    \end{minipage}
    \hfil
    \begin{minipage}[]{0.23\linewidth}
      \centering
      \centerline{\includegraphics[width=1.0\linewidth]{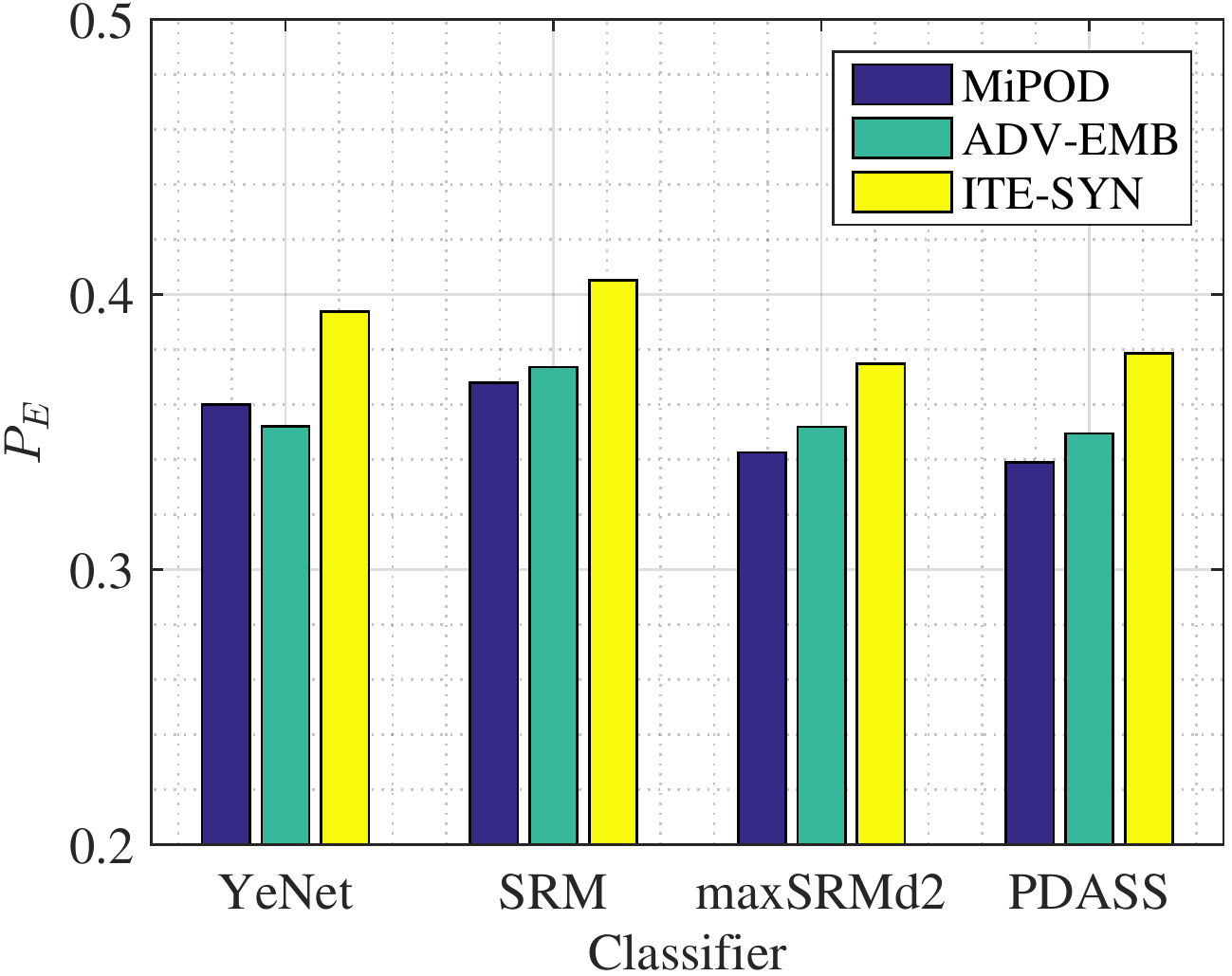}}
      \centerline{(c)}\medskip
    \end{minipage}
    \hfil
    \begin{minipage}[]{0.23\linewidth}
      \centering
      \centerline{\includegraphics[width=1.0\linewidth]{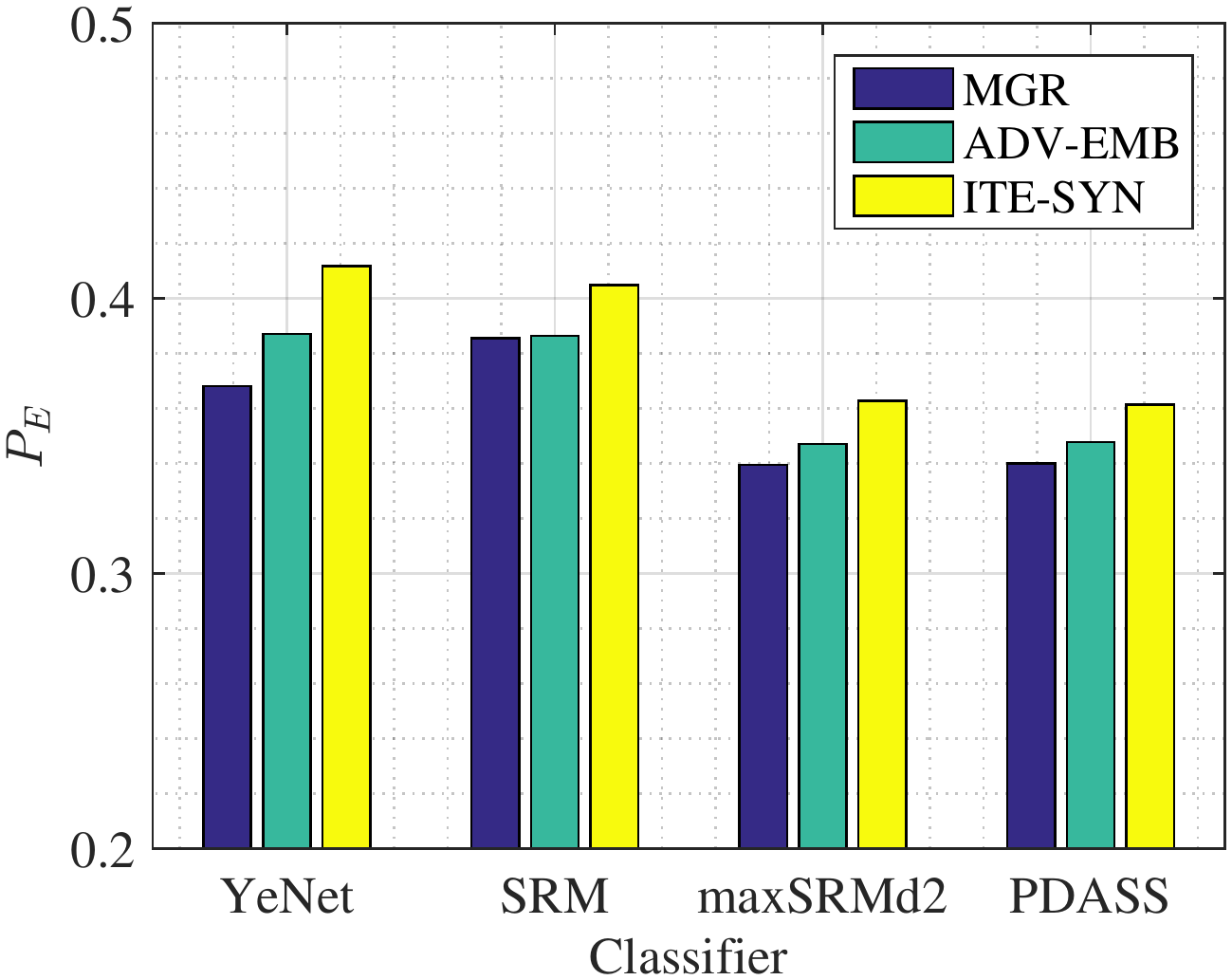}}
      \centerline{(d)}\medskip
    \end{minipage}
    \vfil
    \begin{minipage}[]{0.23\linewidth}
      \centering
      \centerline{\includegraphics[width=1.0\linewidth]{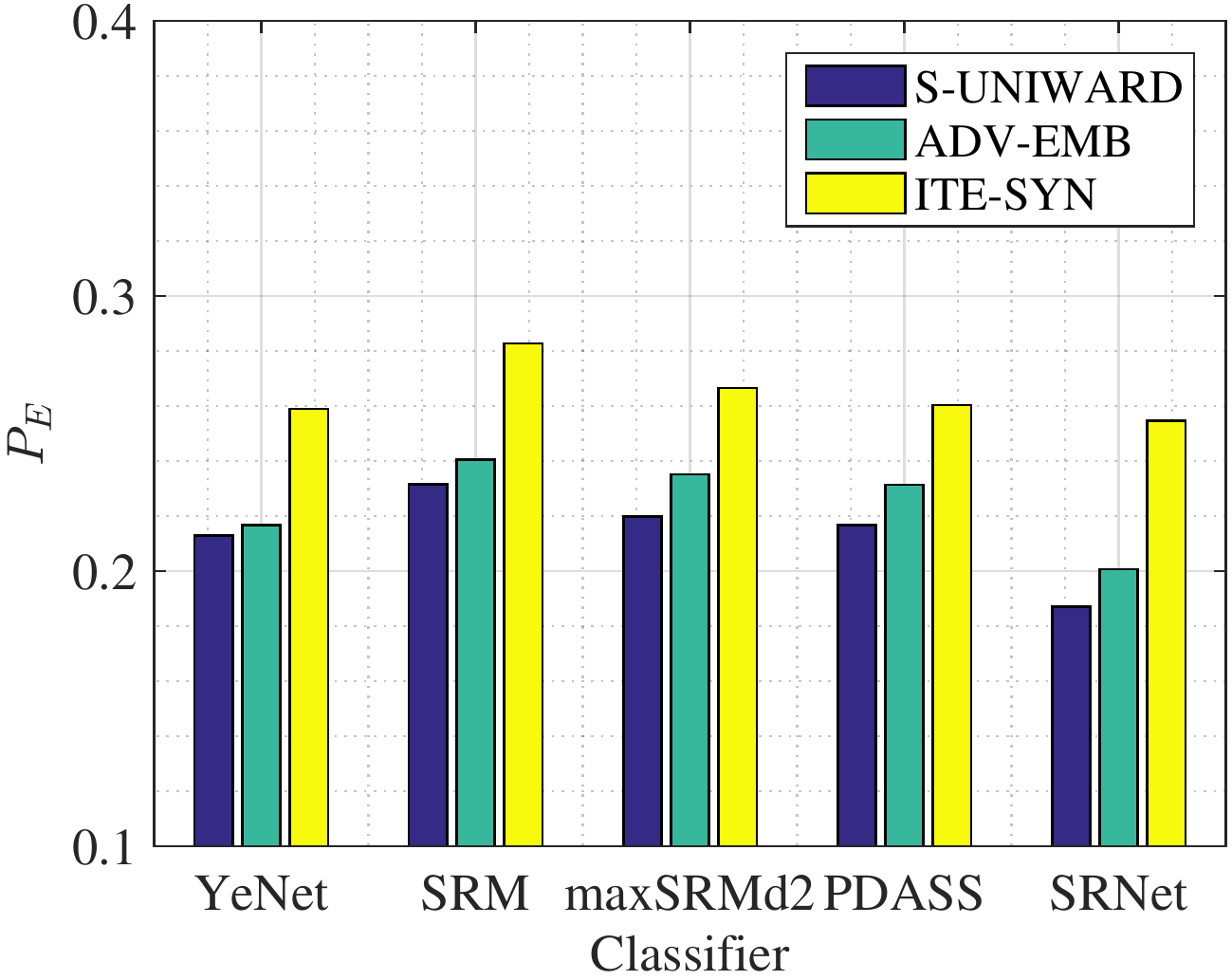}}
      \centerline{(e)}\medskip
    \end{minipage}
    \hfil
    \begin{minipage}[]{0.23\linewidth}
      \centering
      \centerline{\includegraphics[width=1.0\linewidth]{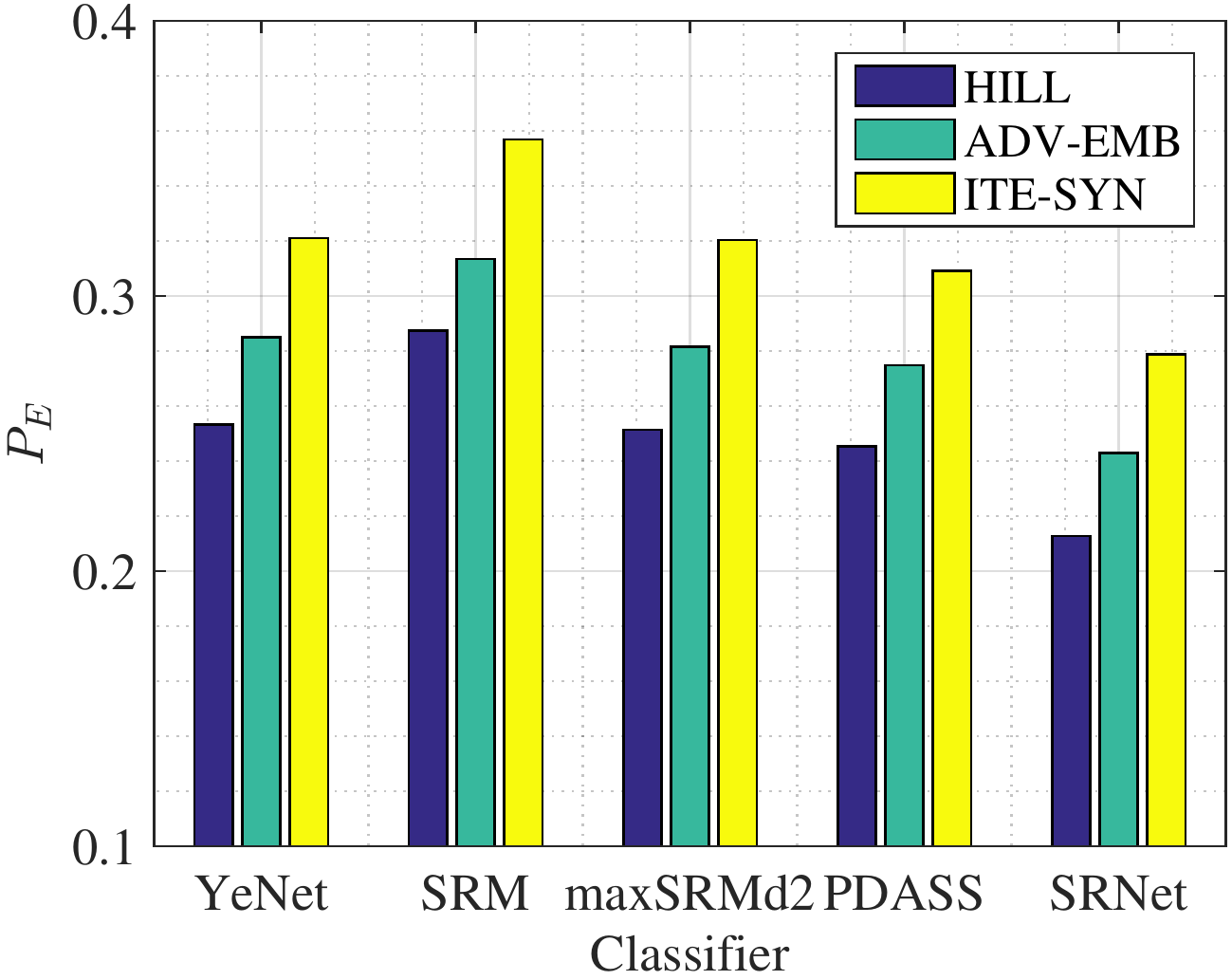}}
      \centerline{(f)}\medskip
    \end{minipage}
    \hfil
    \begin{minipage}[]{0.23\linewidth}
      \centering
      \centerline{\includegraphics[width=1.0\linewidth]{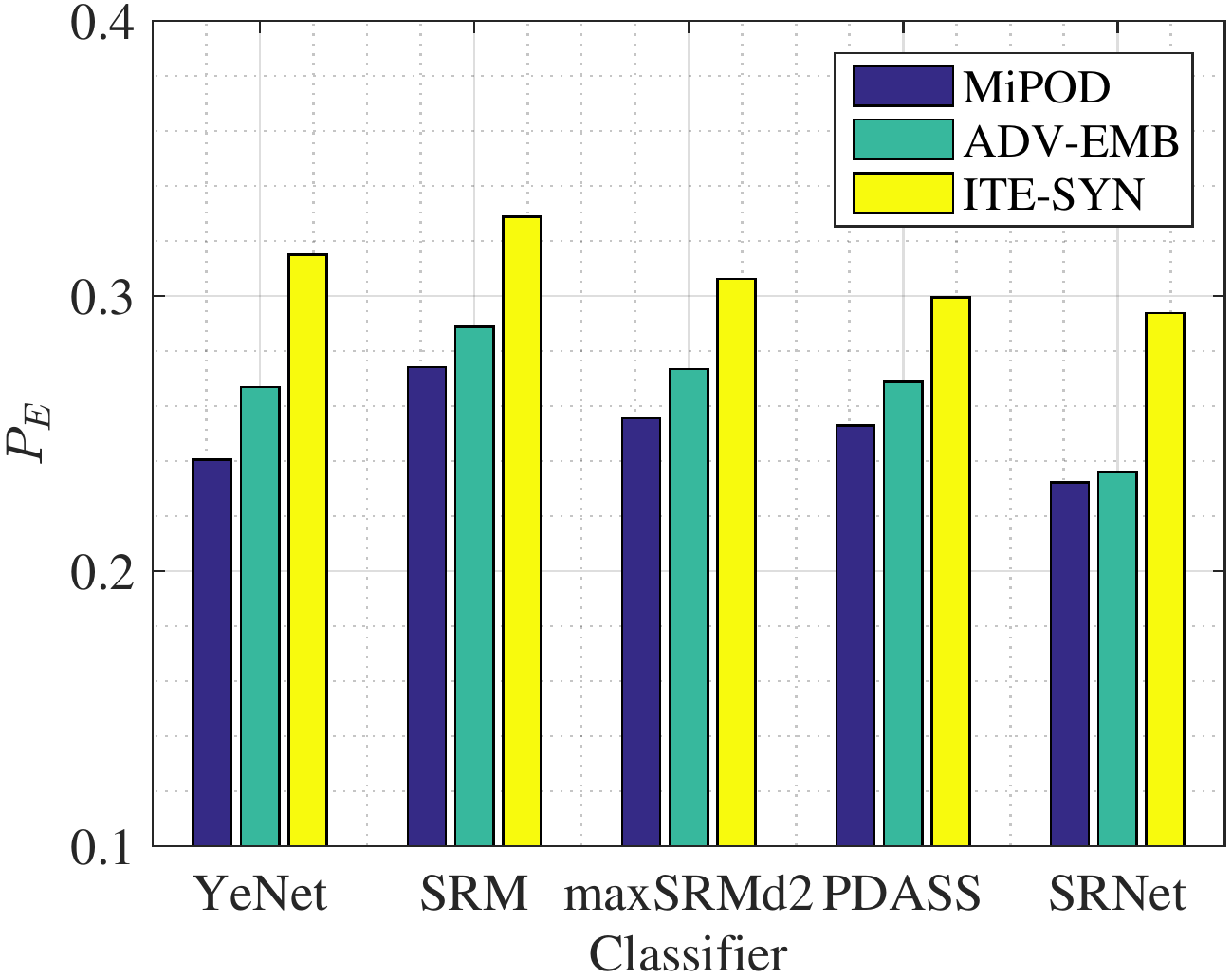}}
      \centerline{(g)}\medskip
    \end{minipage}
    \hfil
    \begin{minipage}[]{0.23\linewidth}
      \centering
      \centerline{\includegraphics[width=1.0\linewidth]{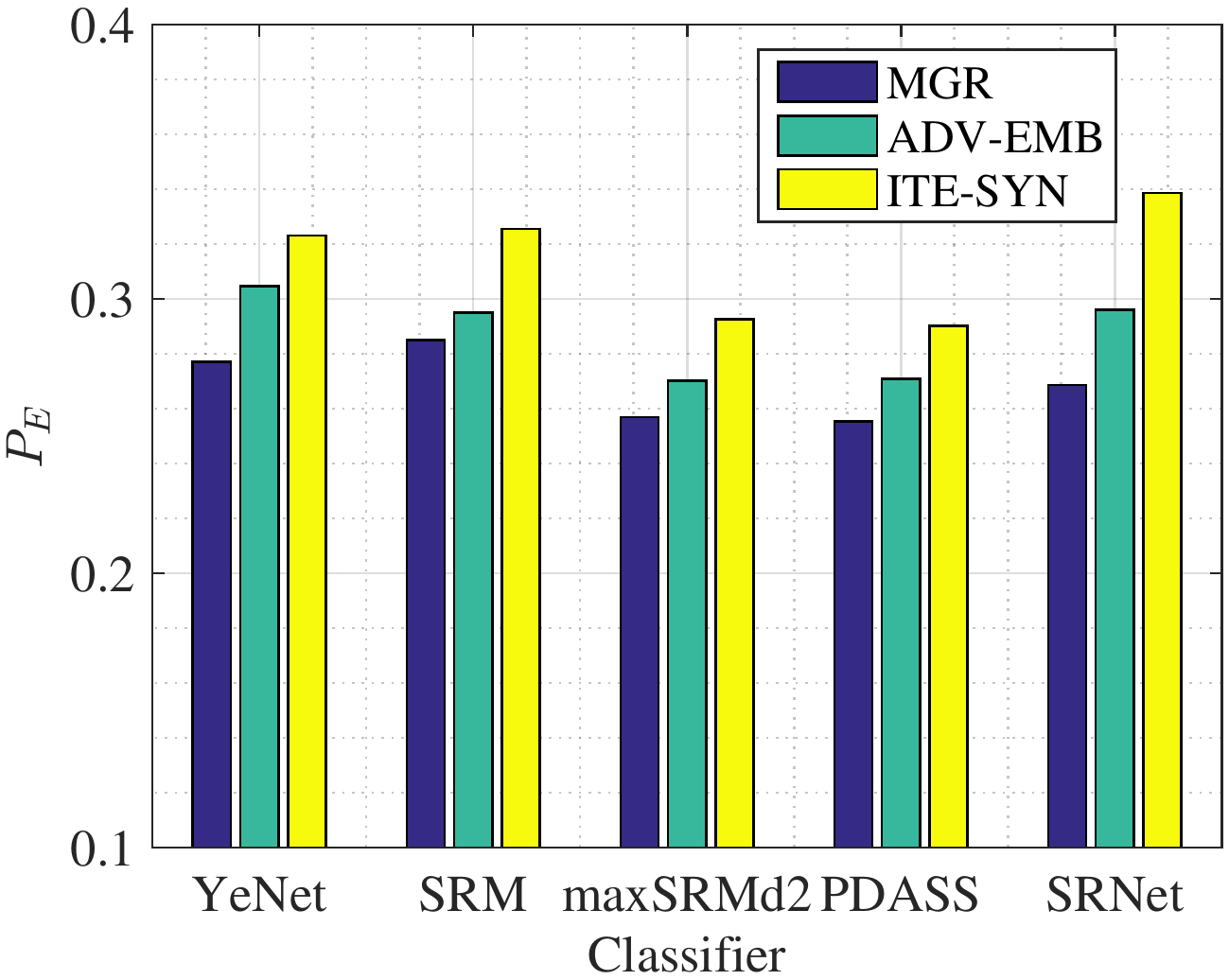}}
      \centerline{(h)}\medskip
    \end{minipage}
  \caption{Performances $P_E$ of resisting on adversarial trained classifiers with the target YeNet for ALASKA256. For (a)-(d), stego images and adversarial stego images were produced with corresponding steganographic schemes under payload rate 0.2 bpp respectively. And (e)-(h) are under payload rate 0.4 bpp respectively.}
  \label{fig:pe_ye_alaska}
\end{figure*}

Many effective methods have been proposed to defend adversarial samples. Adversarial training\cite{Tramer2017} is one of the easiest defensive methods by directly adding adversarial samples to training samples.
We re-trained new classifiers by using adversarial stego images and executed classifications. 
Results of re-trained classifiers for BOSS256 are shown in 
Fig. \ref{fig:pe_retrain_xu} and Fig. \ref{fig:pe_retrain_ye}, and results for ALASKA256 are shown in Fig. \ref{fig:pe_xu_alaska} and Fig \ref{fig:pe_ye_alaska}. Since we skipped testing of payload rate 0.3 bpp, we only re-trained SRNet classifiers under payload rate 0.4 bpp for adversarial stego image sets. For Min-Max function, we only re-trained SRNet classifiers for the ninth round of stego image sets.
There are observations as following.
\begin{itemize}
\item For resisting on target CNN classifiers, ITE-SYN versions perform superior to corresponding ADV-EMB versions.
\item For resisting on XuNet detecting, ITE-SYN versions perform inferior to Min-Max versions sometimes, such as S-UNIWARD under payload rate 0.2 bpp, and HILL, MiPOD and MGR under payload rates 0.2 bpp and 0.4 bpp. However, at least four rounds Min-Max optimal selecting most difficult stego images, Min-Max versions maybe superior to corresponding ITE-SYN versions.
\item For resisting on YeNet detecting, ITE-SYN versions perform superior to corresponding Min-Max versions.
\item For resisting on detecting by other classifiers, including CNN classifiers SRNet and conventional ensemble classifiers, ITE-SYN versions perform superior to corresponding ADV-EMB versions and Min-Max versions.
\end{itemize}

Above all, adversarial stego images created by ITE-SYN can effectively fool target CNN classifiers, and significantly resist on conventional ECs detecting. After adversarial training, ITE-SYN also achieve high steganographic performances to counter both CNN classifiers and conventional models. 
It illustrates that ITE-SYN is benefited from both iteratively increasing adversarial intensity and synchronizing modification directions, including embedding both common sub-images and the adversarial perturbed sub-image.


There is only one adversarial stego image created for a single cover image by using ITE-SYN. However, Min-Max methods have to train a serial of classifiers by using ADV-EMB scheme to optimally select the most difficult stego images.
Taking computational time exhibited in section \ref{sec:exper:ssec:time} into consideration, it indicates that computational complexity of Min-Max method is more than ITE-SYN. 
It is indicated that ITE-SYN performs prior to ADV-EMB and even ADV-EMB applied Min-Max function in summary.
Furthermore, it is predicted that steganographic performances of adversarial training are improved by applying Min-Max function to ITE-SYN scheme to select the most difficult stego images.


\subsection{Computation time}
\label{sec:exper:ssec:time}

\begin{table}[tbp] 
\centering
\caption{Computation time (in seconds) of creating an adversarial stego image.}
\setlength{\tabcolsep}{1.5 pt}
\begin{tabular}{c | c | c c c c}
\hline
\multirow{2}{*}{Target} & \multirow{2}{*}{Scheme} & Success rate(\%) & Minimum & Maximum & Average \\
\cline{3-6}
 & & \multicolumn{4}{|c}{0.2 bpp}  \\
\hline 
\multirow{2}{*}{XuNet} & ADV-EMB & 96.32 & 1.95 & 62.38 & $\mathbf{10.61}$ \\
 & ITE-SYN & 91.01 & 1.48 & 262.38 & 24.80 \\
\hline 
\multirow{2}{*}{YeNet} & ADV-EMB & 99.79 & 1.90 & 61.08 & 6.95 \\
 & ITE-SYN & 98.80 & 1.98 & 254.16 & $\mathbf{6.41}$ \\
\hline 
 & & \multicolumn{4}{|c}{0.4 bpp}  \\
\hline 
\multirow{2}{*}{XuNet} & ADV-EMB & 99.57 & 2.03 & 64.56 & 7.84 \\
 & ITE-SYN & 97.99 & 1.26 & 260.91 & $\mathbf{7.84}$ \\
\hline 
\multirow{2}{*}{YeNet} & ADV-EMB & 99.77 & 1.96 & 70.20 & 6.95 \\
 & ITE-SYN & 99.75 & 1.36 & 252.52 & $\mathbf{3.84}$ \\
 \hline
\multicolumn{6}{p{220 pt}}{Bold values are the minimum average computation time of countering one target classifier.}
\end{tabular}
\label{tab:time_stc}
\end{table}

\begin{figure}[tbp]
  \centering
    \begin{minipage}[]{0.8\linewidth}
      \centering
      \centerline{\includegraphics[width=1.0\linewidth]{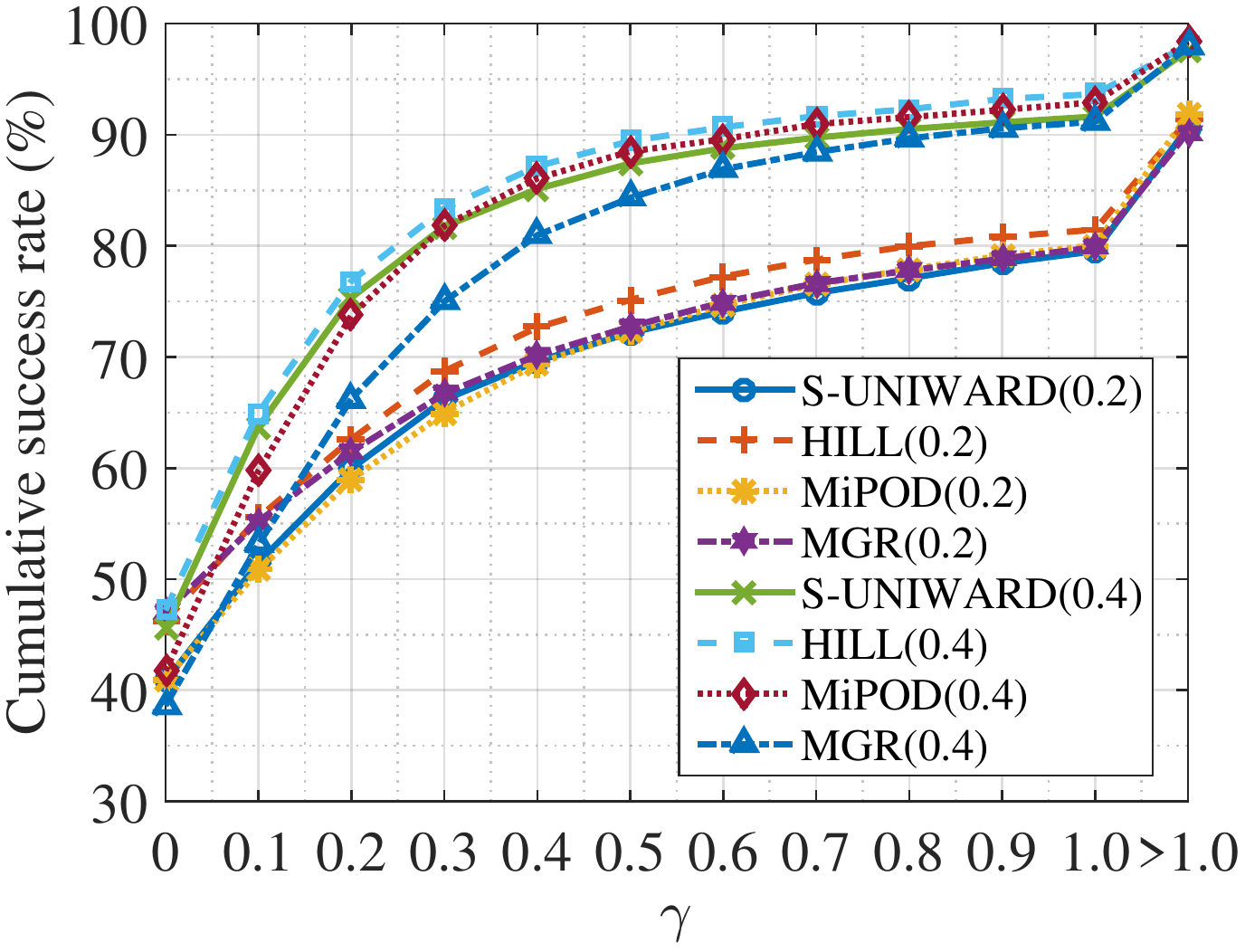}}
      \centerline{(a) The target classifiers are XuNet models.}\medskip
    \end{minipage}
    \vfil
    \begin{minipage}[]{0.8\linewidth}
      \centering
      \centerline{\includegraphics[width=1.0\linewidth]{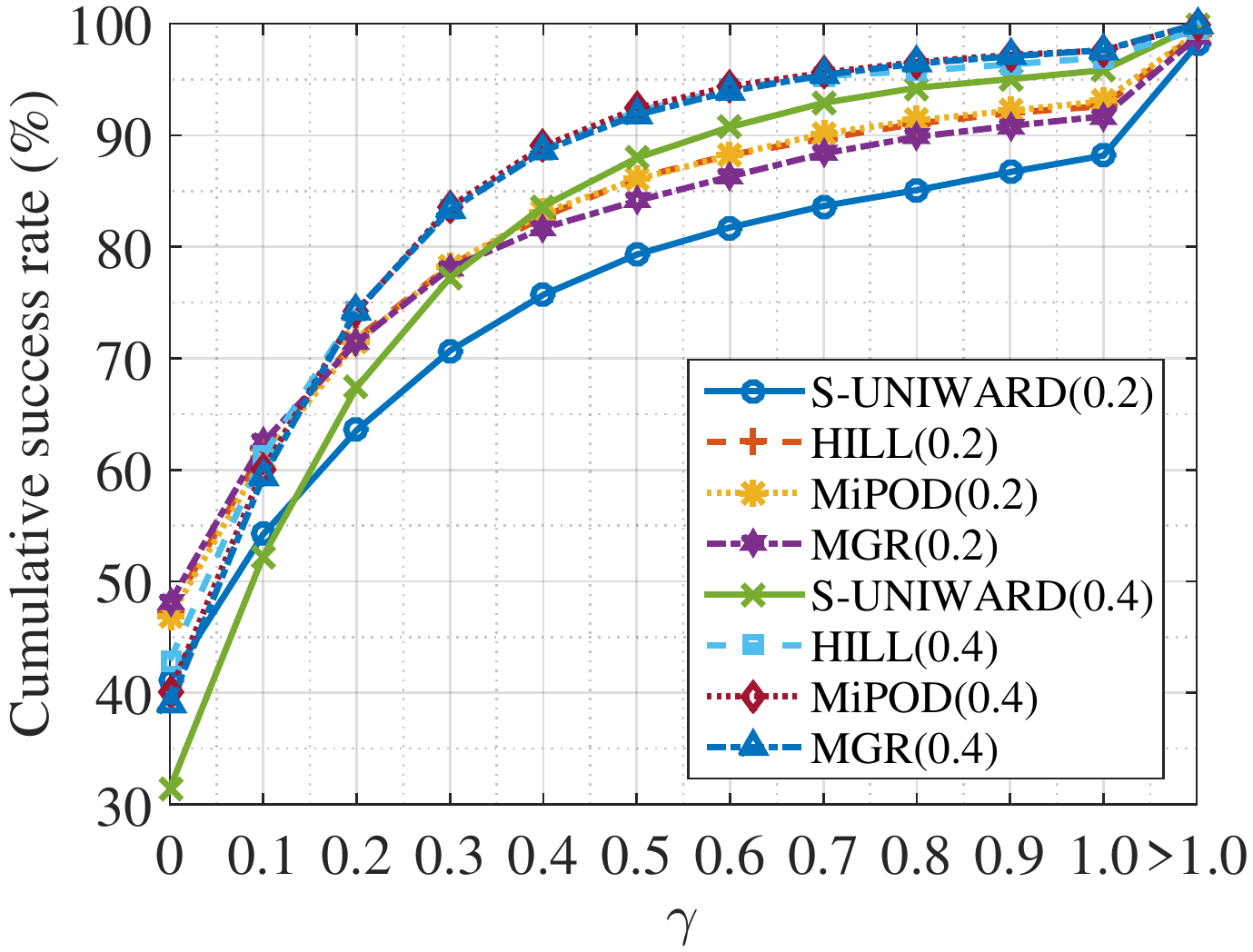}}
      \centerline{(b) The target classifiers are YeNet models.}\medskip
    \end{minipage}
  \caption{Cumulative success rates (in \%) of creating adversarial stego images for BOSS256. Decimal fractions in parenthesis indicate payload rates 0.2 bpp and 0.4 bpp respectively.}
  \label{fig:csr_adv}
\end{figure}

In practice, STCs are used to embed steganographic data. For comparing for computational complexity of producing adversarial stego images, we provided statistical information of computational time of 5000 images in  testing image set of BOSS256 by using STCs. We used the four 
schemes S-UNIWARD, HILL, MiPOD and MGR to compute initial costs. All operations were executed on the workstation equipped with GPU Tesla P100-PCIE-16GB, and the results are shown in TABLE \ref{tab:time_stc}.
Based on the statistics, we can make the following observations.
\begin{itemize}
  \item The maximum computation times of creating a single adversarial stego image are more than 70 seconds and 260 seconds for ADV-EMB and ITE-SYN respectively. For ADV-EMB, adversarial area rate is 0 to 1 with step interval 0.1 and the whole image is re-embedded in each split. Correspondingly, for ITE-SYN, adversarial intensity factor is 0 to 10 with step interval 0.1 and only one sub-image is re-embedded in each iteration. In the worst case, all four sub-images are tried re-embedding and finally the creation is failed by using ITE-SYN. Therefore, the maximum iterations of ADV-EMB and ITE-SYN are 10 and 400 respectively, which indicates that computational time of each iteration are approximately 7 seconds and 0.65 seconds for ADV-EMB and ITE-SYN respectively. Consequently, computational time of a single iteration of ITE-SYN is much less than ADV-EMB's.
  \item Except attacking XuNet under payload rate 0.2 bpp, average computational time of ITE-SYN is less than ADV-EMB's. Specially, for attacking YeNet under payload rate 0.4 bpp, computational time of ITE-SYN is nearly half of ADV-EMB's.
  \item For attacking XuNet under payload rate 0.2 bpp, computational time of ITE-SYN is almost twice of ADV-EMB's. It is noticed that success rate of ITE-SYN is almost $5\%$ inferior to ADV-EMB.
\end{itemize}

For both ADV-EMB and ITE-SYN, failure cases contribute the maximum computational times. For investigating impact of adversarial intensity factor, we collected corresponding information of creating adversarial stego images of testing image set of BOSS256 by using STCs. The cumulative distribution function (CDF) is expressed as:
\begin{equation}\label{eq:fun_cdf}
  \mathscr{P}(x_0)=P_r\{x \leq x_0\}=\int_{-\infty }^{x_0}f(t)dt,
\end{equation}
where $f$ is probability distribution function (PDF).
The cumulative success rates 
of creating adversarial stego images respect to adversarial intensity factor $\gamma$ are exhibited in 
Fig. \ref{fig:csr_adv}.
It is observed that when $\gamma=1$ cumulative success rates of creating adversarial stego images is over 80\% except SUN-SYN-XU under payload rate 0.2 bpp, which is 79.52\%. Therefore, it is practicable selecting an appropriate $\gamma_{max}$ to reduce the maximum computational time of ITE-SYN, such as $\gamma_{max}=1$, in the case the maximum amount of iterations of ITE-SYN is only 40.

For Min-Max method with nine rounds of optimal selections, there are $10$ different stego images (including the original stego image and nine adversarial stego images for each round best classifier) produced for each cover image, which indicates that the minimum computational time for a cover image by using Min-Max method is at least 19 seconds with ignoring expense of optimal selection operations. Using the minimum mean computational time of ADV-EMB, it spends approximate 70 seconds for a most difficult stego image for a single cover image. Which is more than average computational time of ITE-SYN. 

\section{Conclusion}
\label{sec:conclusion}

This paper proposes a novel steganagraphy named ITE-SYN for spatial images. The method incorporates synchronizing modification directions profile and adversarial perturbations together to enhance steganographic security countering both conventional steganalysis and convolutional neural network steganalysis. Based on clustering modification directions, adversarial perturbations are limited in a sub-image with minimum intensity by applying iteratively increasing adversarial intensity. Experiments have evaluated the proposed method and we can carry out conclusions as 
\begin{itemize}
\item [1)] For original classifiers, steganographic performances are improved from baselines, which indicates ITE-SYN effectively resisting on detection of original classifiers, including target and non-target CNN classifiers and original conventional classifiers.
\item [2)] ITE-SYN has significant capability to resist on re-trained classifiers including both CNN models and conventional models.
\item [3)] Only re-embedding one sub-image to generate adversarial noise for an image, ITE-SYN has low computational expense.
\end{itemize}

However, synchronizing modification directions profile is only appropriate for spatial images. Future works will be devoted to investigate effective cost profile of images for other domain such as JPEG to corporate iterative adversarial perturbations.

Parameters both CMD factor $\beta$ and adversarial intensity factor $\gamma$ are arbitrarily set in this paper. In practice, optimal parameters can be pursued.

\bibliographystyle{IEEEtran}
\bibliography{IEEEabrv,stegan}

\end{document}